\documentclass{amsart}
\usepackage{microtype}
\usepackage{amsmath, amssymb, amsthm}
\usepackage{mathtools}
\usepackage{tikz-cd}
\usepackage{mathbbol} % changes \mathbb{} and adds more support

\usepackage[
backend=biber,
style=alphabetic,
backref=true,
url=false,
doi=false,
isbn=false,
eprint=false]{biblatex}

\renewbibmacro{in:}{} % don't display "in:" before the journal name
\AtEveryBibitem{\clearfield{pages}} % don't show page numbers

\DeclareFieldFormat{title}{\myhref{\mkbibemph{#1}}}
\DeclareFieldFormat
[article,inbook,incollection,inproceedings,patent,thesis,unpublished]
{title}{\myhref{\mkbibquote{#1\isdot}}}

\newcommand{\doiorurl}{%
	\iffieldundef{url}
	{\iffieldundef{eprint}
		{}
		{http://arxiv.org/abs/\strfield{eprint}}}
	{\strfield{url}}%
}

\newcommand{\myhref}[1]{%
	\ifboolexpr{%
		test {\ifhyperref}
		and
		not test {\iftoggle{bbx:eprint}}
		and
		not test {\iftoggle{bbx:url}}
	}
	{\href{\doiorurl}{#1}}
	{#1}%
}
\usepackage[
bookmarks=true,
linktocpage=true,
bookmarksnumbered=true,
breaklinks=true,
pdfstartview=FitH,
hyperfigures=false,
plainpages=false,
naturalnames=true,
colorlinks=true,
pagebackref=false,
pdfpagelabels]{hyperref}

\hypersetup{
	colorlinks,
	citecolor=blue,
	filecolor=blue,
	linkcolor=blue,
	urlcolor=blue
}

\usepackage[capitalize, noabbrev]{cleveref} % hyper- and cross-referencing 

\makeatletter
\@namedef{subjclassname@2020}{%
	\textup{2020} Mathematics Subject Classification}
\makeatother

\newtheorem{theorem}{Theorem}
\newtheorem*{theorem*}{Theorem}
\newtheorem{proposition}[theorem]{Proposition}
\newtheorem*{proposition*}{Proposition}
\newtheorem{conjecture}[theorem]{Conjecture}
\newtheorem*{conjecture*}{Conjecture}
\newtheorem{lemma}[theorem]{Lemma}
\newtheorem*{lemma*}{Lemma}
\newtheorem{corollary}[theorem]{Corollary}
\newtheorem*{corollary*}{Corollary}
\theoremstyle{definition}
\newtheorem{definition}[theorem]{Definition}
\newtheorem*{definition*}{Definition}
\newtheorem{remark}[theorem]{Remark}
\newtheorem*{remark*}{Remark}

\newtheorem*{example*}{Example}

\newtheorem*{construction*}{Construction}

\newtheorem*{convention*}{Convention}

\newtheorem*{terminology*}{Terminology}

\newtheorem*{notation*}{Notation}

\newtheorem*{question*}{Question}

\newcommand{\R}{\mathbb{R}}

\DeclarePairedDelimiter\angles{\langle}{\rangle}

\usepackage{todo}

\usepackage{amsfonts}
\usepackage{graphicx}
\usepackage{xcolor}
\usepackage{enumitem}
\usepackage{caption}
\usepackage{subcaption}
\usepackage{algorithm}
\usepackage{algpseudocode}
\usepackage{booktabs}
\usepackage{enumitem}
\usepackage{hyperref} % new packages here
\DeclareMathOperator{\length}{length}

\newcommand{\AK}{\text{AK}}
\newcommand{\MS}{\text{MS}}

\newcommand{\dm}{d_{\text{model}}}

\newcommand{\softmax}{\text{Softmax}}

\newcommand{\LN}{\text{LN}}

\theoremstyle{theorem}
\newtheorem{introtheorem}{Theorem}  
\title[What makes math problems hard for RL]{What makes math problems hard for reinforcement learning: a case study}
\author[A.~Shehper et al.]{A.~Shehper}
\address{A.S. | NHETC, Department of Physics and Astronomy, Rutgers University, Piscataway, New Jersey 08854, USA}

\author[]{A.~Medina-Mardones}
\address{A.M. | Department of Mathematics, Western University, ON, Canada}

\author[]{L.~Fagan}
\address{L.F. | Department of Mathematics, University of California, Santa Barbara, CA 93106, USA}

\author[]{B.~Lewandowski}
\address{B.L. | Institute of Mathematics, University of Warsaw, ul. Banacha 2, 02-097 Warsaw, Poland}

\author[]{A.~Gruen}
\address{A.G. | Polygon Zero}

\author[]{Y.~Qiu}
\address{Y.Q. | Chern Institute of Mathematics and LPMC, Nankai University, Tianjin 300071, China}

\author[]{P.~Kucharski}
\address{P.K. | Institute of Mathematics, University of Warsaw, ul. Banacha 2, 02-097 Warsaw, Poland}

\author[]{Z.~Wang}
\address{Z.W. | Department of Mathematics, University of California, Santa Barbara, CA 93106, USA}

\author[]{S.~Gukov}
\address{S.G. | Richard N. Merkin Center for Pure and Applied Mathematics, California Institute of Technology, Pasadena, CA 91125, USA}

\date{\today}

\begin{document}
	\vspace*{1cm}
	\begin{center}
		\textsc{---PREPRINT---}
	\end{center}
	\vspace*{1cm}

	% !TEX root = ../ac_paper.tex

\begin{abstract}
    Using a long-standing conjecture from combinatorial group theory, we explore, from multiple perspectives, the challenges of finding rare instances carrying disproportionately high rewards.
    Based on lessons learned in the context defined by the Andrews--Curtis conjecture, we propose algorithmic enhancements and a topological hardness measure with implications for a broad class of search problems.
    As part of our study, we also address several open mathematical questions.
    Notably, we demonstrate the length reducibility of all but two presentations in the Akbulut--Kirby series (1981), and resolve various potential counterexamples in the Miller--Schupp series (1991), including three infinite subfamilies.
\end{abstract}
	\maketitle
	\tableofcontents
	% !TEX root = ../ac_paper.tex

\section{Introduction\label{sec:intro}}

We live in an extraordinary era where artificial intelligence (AI) is transforming numerous sectors and professions. Recent advancements in Large Language Models (LLMs) have empowered AI to read, write, and converse with a proficiency comparable to that of human experts. In the realm of board games, AI has outperformed even the most skilled human players, and it has tackled complex scientific challenges like protein folding, where steady progress was suddenly overtaken by a near-complete solution. As AI continues to evolve, one critical question remains: How wide is the range of domains in which AI systems can reason as effectively as humans?

Mathematics appears to be a natural progression on the path toward Artificial General Intelligence (AGI) due to its universal syntactic and logical structure, similar to that of natural language. Additionally, mathematics provides a framework for the quantitative evaluation of logical and analytical reasoning, making it an ideal domain for self-improving AI systems on the path to AGI. In a moment, we will explain another reason why mathematics could play a crucial role in AGI development, but first, we need to introduce one more key element: reinforcement learning (RL).

Machine learning, a subfield of AI, involves developing algorithms and statistical models that enable computers to learn from data and make predictions. Among the three primary areas of machine learning --- supervised learning, unsupervised learning, and reinforcement learning --- RL emphasizes learning through interaction with an environment and receiving feedback in the form of rewards or penalties. This aspect of machine learning, often characterized by its focus on AI models `playing games,’ will be central to our discussion.

A typical chess game lasts about 30 to 40 moves, with the longest recorded professional game reaching 269 moves, ending in a draw between Ivan Nikolic and Goran Arsovic in 1989. Notably, the number of moves in a typical chess game is relatively consistent, with the longest professional game having only about an order of magnitude more moves than the average. Similarly, a typical game of Go involves a few hundred moves, with the longest recorded professional game, played by Go Seigen and Kitani Minoru in 1933, lasting 411 moves.

At first glance, proving or disproving mathematical conjectures can be formulated as games. For example, proving a theorem involves finding a path from the hypothesis to the conclusion, consisting of basic logical steps, such as Lean steps. From the RL perspective, this type of game can be quite complex due to its large action space. Conversely, finding examples or counterexamples to settle important conjectures may require only a few basic moves (actions); the case study in this paper serves as a good illustration of such a problem. In all cases, the problem is fundamentally a search process, with complexity largely determined by the size of the action space and the search space.

In addition, with hard research-level mathematics problems, one faces yet another challenge: the sought-after instance can be so rare and difficult to find that the problem effectively becomes a search for a needle in a haystack, i.e., a problem with ultra-sparse rewards. For example, in the context of theorem proving, one might consider an extremely hard theorem\footnote{A proxy for such a problem could be the Riemann Hypothesis or any other unsolved Millennium Prize Problem.} that may require a very large number of steps. If there aren't many alternative proofs, finding even a small number of very long ones then becomes akin to searching for a needle in a haystack or, depending on one's preference, a search for a unicorn.
\footnote{Similar challenges, though not as critical, also arise in non-research-level math problems; see, e.g., \cite{peano,dabelow2024symbolicequationsolvingreinforcement,trinh2024solving} for recent discussion. Meanwhile, in a parallel line of development, new benchmarks have been proposed in the past couple of years \cite{procgen,NeedleInAHaystack}, which can be useful in such contexts.}

Fortunately, mathematics offers a robust framework for developing and testing new algorithms with adaptive capabilities that dynamically `learn how to learn.' Testing these algorithms on mathematical problems, rather than directly on societal issues like market crash predictions or extreme weather events, provides a risk-free and cost-effective approach. Additionally, this method offers the dual benefit of potentially solving hard mathematical problems and resolving long-standing conjectures in the process.

Our approach to problems of this type involves equipping the RL model with the ability to assess the hardness of problems during the training process. First and foremost, this requires a practically useful notion of hardness, a concept we thoroughly explore. By learning the distribution of problems based on their difficulty, one can enhance existing off-the-shelf algorithms with new self-improvement strategies, identifying key features that facilitate solving the most challenging problems.

We begin a systematic implementation of this approach by carefully analyzing the distribution of hardness in potential counterexamples to a long-standing conjecture, the Andrews--Curtis conjecture.
This setting offers distinct advantages for algorithm development compared to generic mathematical problems.
First, the action space in this setting is finite and small, consisting of only a few basic moves.
Second, each instance has an intrinsic measure of complexity---its \textit{length}---
relative to which
the number of steps required by an RL agent ranges from linear to superexponential
\cite{Bridson, Lishak}.

This work formulates the Andrews–Curtis trivialization problem as a reinforcement learning environment with long horizons, sparse rewards, and an intrinsic hardness distribution. It systematically explores this hardness distribution across a broad class of examples using classical search, reinforcement learning, language modeling, and topological data analysis. Additionally, it introduces a principled topological measure applicable to similar \textit{path-to-base} search problems. Finally, it examines how reinforcement learning agents navigate problems of varying difficulty, leveraging insights into problem hardness to propose novel algorithms that dynamically adapt to challenging tasks.

While our primary focus is on exploring the distribution of hardness to inform algorithm development, we also discover new mathematical results about long-standing potential counterexamples of this conjecture. 

A notable family of potential counterexamples, denoted $\AK(n)$, due to Akbulut and Kirby, 
\[
\AK(n) = \angles{x, y \mid x^n = y^{n+1}, xyx = yxy}, \quad n \geq 3
\]
has been open in mathematics for more than four decades \cite{Akbulut--Kirby}.\footnote{The original work included the case $n=2$, but this case was solved approximately two decades later \cite{genetic}.The case $n=1$ also gives a balanced presentation of the trivial group, which is easy to solve in a few steps.}
The length of the presentation $\AK(n)$ is $2n + 7$, and until now, it was unknown whether this length can even be reduced using AC moves. We find that for every fixed $n$, $\AK(n)$ is AC-equivalent to each element of the following 1-parameter family of presentations parameterized by $k \in \mathbb{Z}$,
\[
P(n, k) = \angles{x, y \mid  y^{n-k-1}  x^{-1} y x = x y x^{-1} y^{n-k} \ , \ x = y^{-k} x^{-1} y x y }.
\]
When $n \geq 5$, this family contains presentations of length less than $2n + 7$. The shortest presentation occurs in the case $k=n-1$, which gives us the following result, proven as \cref{t:n+11} in the main text.

\begin{introtheorem}
	For every $n\geq 2$, $\AK(n)$ is AC-equivalent to the presentation
	\[
	\angles{ x,y \mid x^{-1} y x = x y x^{-1} y \ ,\  xy^{n-1}x=yxy },
	\]
	of length $n+11$. This gives a reduction in length of $AK(n)$ for all $n \geq 5$.
\end{introtheorem}

Another important family of potential counterexamples, due to Miller and Schupp, is
\[
\MS(n, w) = \angles{x, y \mid x^{-1} y^n x = y^{n+1}, x = w},
\]
where $n > 0$ and $w$ is a word with exponent sum zero on $x$. 
It has been open for more than 25 years \cite{Miller--Schupp}.

\begin{introtheorem}
        The following infinite subfamilies of Miller--Schupp presentations are AC-trivial:
        \begin{enumerate}[label=(\roman*)]
            \item $\MS(1, w)$ for all $w$,
            \item $\MS(n, w_\star)$ for all $n > 0$ and $w_\star = y^{-1} x y x^{-1}$, 
            \item $\MS(2, w_k)$ for $w_k = y^{-k} x^{-1} y x y$ where $k \in \mathbb{Z}$.
        \end{enumerate}
        Furthermore, for each fixed $n > 0$, all members of the 1-parameter family, $\MS(n, w_k)$, parameterized by $k \in \mathbb{Z}$, are AC-equivalent to each other and to $\AK(n)$.
\end{introtheorem}

In addition to these results involving infinite families, respectively proven as \cref{t:ms1wt,t:msnw,c:ms2wk}, we also found AC-trivializations of hundreds of Miller--Schupp presentations. Of special note are two hard presentations, which our classical search algorithms could not solve at first, but were solved by a reinforcement learning agent (\cref{thm:MS}):
	\begin{gather*}
		\angles{x, y \mid x^{-1} y^3 x y^{-4} , \ x^{-1} y^{-1} x y^{-3} x^{-1} y^{-1}}, \\
		\angles{x, y \mid x^{-1} y^4 x y^{-5} , \ x^{-1} y x y^{-1} x^{-1} y^{-3}}.
	\end{gather*}

Note that the first of these presentations is extremely similar to the following rewriting of $\MS(3, w_{-3})$,
\[
\MS(3, w_{-3}) = \angles{x, y \mid x^{-1} y^3 x y^{-4}, x^{-1} y^{-1} x y^{-3} x y^{-1}}.
\]
with all but one letter being different. The latter presentations is AC-equivalent to $\AK(3)$---perhaps the most important potential counterexample to the Andrews--Curtis conjecture.
This shows how very similar-looking presentations can have vastly different hardness properties, emphasizing the need for a systematic understanding of hardness in math problems.

\medskip
A weaker version of the AC conjecture can be obtained by incorporating additional \textit{stable moves}. This establishes a direct connection between the conjecture and knot theory, a relationship we explore in \cref{sec:stable}. Motivated by this connection and by the claim in \cite{MMS} that the presentation
\begin{equation}\label{eq:MMS3}
	\langle x, y \mid
	x^{-1}y^{-1}xy^{-1}x^{-1}yxy^{-2}xyx^{-1}y, \
	y^{-1}x^{-1}y^2x^{-1}y^{-1}xyxy^{-2}x \rangle
\end{equation}
is stably AC-trivial, we apply reinforcement learning to demonstrate its AC equivalence to $\AK(3)$.
However, after closer inspection, we discovered a subtle misprint in \cite[p.10]{MMS} that had gone unnoticed for more than 20 years. This misprint undermines the claim that \eqref{eq:MMS3} is stably AC-trivial.

\medskip
We should emphasize that this entire project was carried out using relatively modest computational resources that a small academic research group can easily afford. Consequently, we placed greater emphasis on theoretical tools and techniques that, when combined with scaled-up computational resources, can lead to further advancements.

\subsection*{Outline}

In \cref{sec:AC}, we provide an overview of the Andrews--Curtis conjecture.

We then apply classical search algorithms to examples in the Miller--Schupp series in \cref{sec:search}, where we devise a greedy search algorithm that significantly outperforms the widely used breadth-first search algorithm.

In \cref{sec:rl}, we employ reinforcement learning, specifically the Proximal Policy Optimization (PPO) algorithm \cite{schulman2017proximal}, and find that while it performs better than breadth-first search and solves some presentations that are hard for greedy search, it does not strictly surpass the greedy search algorithm.

Building on these insights, \cref{sec:algo} presents several ideas for algorithm development, proposing strategies to mitigate the overwhelming complexity faced by an RL agent in challenging problems.

In \cref{sec:lm}, we examine the linguistic structure of balanced presentations using a decoder-only Transformer model, observing distinct clusters within the embedding space corresponding to presentations solved and unsolved by the greedy search algorithm.

In \cref{sec:global}, we introduce a principled measure of the hardness of general \textit{path-to-base} search problems.
We approximate this invariant for the AC and AC$'$ trivialization problems showing the consistency of the measure up to an overall scaling.

In \cref{sec:local}, we analyze the hardness of a presentation by examining how much of it can be predicted from the local structure of the problem. 
Specifically, we study the subgraph of the AC graph consisting of all presentations connected to a given presentation through a small number of AC moves.  

Finally, in \cref{sec:stable}, we study the stable version of the Andrews--Curtis conjecture, and its relationship with knot theory.
This relationship has previously been used to give potential counterexamples of the standard Andrews--Curtis conjecture.
We prove AC-triviality of a large class of these presentations. 

\medskip\noindent
We encourage the reader to explore the accompanying GitHub repository:
\begin{center}
	\href{https://github.com/shehper/AC-Solver}{https://github.com/shehper/AC-Solver}
\end{center}
    %\input{sec/related}
	% !TEX root = ../ac_paper.tex

\subsection*{Acknowledgment}

We would like to thank Danil Akhtiamov, Anna Beliakova, Jessica Craven, Michael Douglas, Konstantin Korovin, Alexei Lisitsa, Maksymilian Manko, Ciprian Manolescu, Fabian Ruehle, Josef Urban, Richard Wedeen, Tony Yue Yu, and Miriam Lipniacka for insightful discussions and comments. We especially want to thank Anna Beliakova for igniting our interest in the Andrews--Curtis conjecture as a framework for exploring problems with long and rare sequences of moves that an RL agent must discover.
	% !TEX root = ../ac_paper.tex

\section{Andrews--Curtis conjecture\label{sec:AC}}

The Andrews--Curtis Conjecture concerns the study of \textit{balanced presentations} of the trivial group, i.e.
presentations of the trivial group with an equal number of generators and relators.
The conjecture proposes that any balanced presentation of the trivial group
\[
\angles{x_1, \dots, x_n \mid r_1, \dots, r_n}
\]
can be converted to the trivial presentation
\[
\angles{x_1, \dots, x_n \mid x_1, \dots, x_n}
\]
through a series of the following operations known as \textit{AC-moves} \cite{Andrews--Curtis}:
\begin{enumerate}[label=(AC\arabic*)]
	\item Substitute some $r_i$ by $r_i r_j$ for $i \neq j$.
	\item Replace some $r_i$ by $r_i^{-1}$.
	\item Change some $r_i$ to $g r_i g^{-1}$ where $g$ is a generator or its inverse.
\end{enumerate}

We will refer to the sum of the word lengths of all relators as the \textit{length} of a presentation.
Two presentations that can be transformed into each other by a sequence of AC-moves are said to be \textit{AC-equivalent}.
A presentation that is AC-equivalent to the trivial presentation is referred to as \textit{AC-trivial}.
Despite considerable efforts, little progress has been made in establishing a proof of the conjecture.
At the same time, several families of potential counterexamples have been proposed in the literature.

To investigate a given presentation, one may systematically explore the entire space of possible sequences of AC-moves in search of a sequence that renders the presentation trivial.
This space grows exponentially with the length of the sequence.
For a presentation with $n$ generators, there are $3n^2$ AC-moves, and the total number of sequences of AC-moves of length $k$ is $(3n^2)^k$.
Even for a modest case like $n=2$ and $k=20$, the number of possible sequences is on the order of $10^{21}$, making a brute-force approach impractical.

Traditional search algorithms such as genetic algorithms \cite{genetic}, and breadth-first search \cite{bfs-ac} have been employed to search through this space and achieved success in trivializing balanced presentations with two generators and lengths less than 13.
The following presentation of length 13,
\[
\angles{x, y \mid x^3 = y^4, xyx = yxy}
\]
is the shortest presentation, up to AC-equivalence, that eludes all attempts at length reduction.
This presentation is a part of an infinite series of potential counterexamples by Akbulut and Kirby \cite{Akbulut--Kirby}:
\[
\AK(n) = \angles{x, y \mid x^n = y^{n+1}, xyx = yxy}, \quad n \geq 2.
\]
$\AK(2)$ has length 11 and has been established to be AC-trivial \cite{genetic}, whereas $\AK(3)$ is the aforementioned presentation with length 13.

In over two decades since the first utilization of search algorithms \cite{genetic, bfs-ac}, all attempts to trivialize $\AK(3)$ using different variants of breadth-first search algorithms, even with increasing amount of computational resources, have failed \cite{Bowman-McCaul, krawiec2016distance, Panteleev-Ushakov}.
Notably, \cite{Panteleev-Ushakov} found that no sequence of AC-moves that allows relator lengths to increase up to 20 trivializes $\AK(3)$.
This persistent lack of success could be interpreted as suggestive evidence that $\AK(3)$ might be a counterexample to the Andrews--Curtis conjecture.

However, recent works by Bridson and Lishak have shown that there exist AC-trivializable balanced presentations, for which the number of AC-moves in a trivializing sequence is bounded below by a superexponential function of the length of the presentation \cite{Bridson, Lishak}.
Roughly speaking, for these presentations, if the sum of word lengths is $k$, the number of AC-moves required to trivialize the presentation is at least $\Delta (\lfloor \log_2 k \rfloor)$ where $\Delta \colon \mathbb{N} \to \mathbb{N}$ is defined recursively as $\Delta(0) = 2$ and $\Delta (j) = 2^{\Delta(j-1)}$ for $j \geq 1$.
In particular, $\Delta (\lfloor \log_2 (13) \rfloor) = 65536$, whereas presentations trivialized by the aforementioned search algorithms have AC sequences of length less than $1000$.
Although $\AK(3)$ is itself not amongst the presentations studied by Bridson and Lishak, their findings challenge the inclination to view $\AK(3)$ as a counterexample. Their work further highlights the need for novel approaches to study the Andrews-Curtis conjecture.

One approach, which we explore in this paper, is to use compositions of AC-moves instead of AC-moves themselves to solve potential counterexamples of the conjecture. The main motivation here being that while a presentation may require a superexponential number of AC-moves to trivialize, the number of composed moves, which we call ``supermoves", may be much smaller. As an example, consider the following \emph{substitution} move which comes from \cite{BurnsI},
\footnote{A much larger class of ``supermoves" called ``elementary M-transformations" was studied in \cite{BurnsI}. However, these moves are infinite in number and are hence not suitable for classical search techniques. In \cref{sec:algo}, we propose an alternative method to discover new supermoves using deep learning methods.}

\begin{definition}[Substitution]\label{def:ac-sub}
	Let $\angles{x_1, \dots, x_n \mid r_1, \dots, r_{i-1}, w^{-1}w', r_{i+1}, \dots  r_n}$ be a balanced presentation of the trivial group with some words $w$, $w'$ in generators. By a sequence of AC-moves, we may replace any occurrence of $w$ in a relator $r_j$ (where $j \neq i$) with $w'$.
\end{definition}

To see how to realize substitutions with AC-moves, consider by means of an example substituting $w'$ for $w$ in the relator $w_1ww_2$: first, conjugate to get $w_2w_1w$; then multiply by $w^{-1}w'$ to write $w_2w_1w'$; and finally, conjugate again to get the required result $w_1w'w_2$. For a relator of the form $w_1 w w_2 \dots w_{l-1} w w_l$, we may similarly substitute any number of occurences of $w$ with $w'$. The number of AC moves packaged into the substitution move is a linear function of the lengths of $w'$ and $w_i$.
We will frequently use this move in \cref{sec:ms1w_} and \cref{sec:len-reduction} to obtain new results pertaining to potential counterexamples of the Andrews-Curtis conjecture.

We also aim to uncover additional insights related to the Andrews-Curtis conjecture. To achieve this, we employ various computational tools in this paper to deepen our understanding of the properties of balanced presentations of the trivial group.
We will test the efficacy of our approaches on a subset of presentations from the Miller--Schupp series of potential counterexamples \cite{Miller--Schupp}:
\[
\MS(n, w) = \angles{x, y \mid x^{-1} y^n x = y^{n+1}, x = w}.
\]
Here, $n \ge 1$ is a positive integer and $w$ is a word in $x$ and $y$ with zero exponent sum on $x$.
For $w_1 = y^{-1} x^{-1} y x y$, the presentations $\MS(n, w_1)$ are AC-equivalent to the presentations from Akbulut--Kirby series \cite{MMS}.
In particular, the presentation
\[
\MS(3, w_1) = \angles{x, y \mid x^{-1} y^3 x = y^{4}, x = y^{-1} x^{-1} y x y}.
\]
of length 15 is AC-equivalent to $\AK(3)$.
\footnote{We will find, in \cref{sec:len-reduction} of this paper, that $\AK(n)$ is, in fact, AC-equivalent to all members of a 1-parameter family of presentations, $MS(n, w_k)$, where $w_k = y^{-k} x^{-1} y x y$ and $k \in \mathbb{Z}$.}

We will only consider presentations with $n \leq 7$ and $\length(w) \leq 7$.
Our selection criteria aims to strike a balance: we seek a dataset of presentations large enough to allow for meaningful analysis, yet small enough to ensure all computations are feasible within a practical timeframe.
We reduce $x^{-1}w$ freely and cyclically, and if two presentations have the same fixed $n$, but different words $w$ and $w'$ such that letters of $x^{-1} w$ can be cyclically permuted to obtain $x^{-1} w'$, we keep only one of these presentations.
After these simplifications, we find $170$ choices of presentations for each fixed $n$, resulting in a dataset of $7 \times 170 = 1190$ presentations for our analysis.

Our implementation of AC-transformations differed from the AC-transformations mentioned above in two ways.
First, we considered the following set of operations.
\begin{enumerate}[label=(AC$'$\arabic*)]
	\item Replace some $r_i$ by $r_i r_j^{\pm 1}$ for $i \neq j$.
	\item Change some $r_i$ to $g r_i g^{-1}$ where $g$ is a generator or its inverse.
\end{enumerate}
The group generated by these AC-transformations is isomorphic to the group generated by the original AC-transformations.\footnote{
	The difference lies in how the inversion of a relator is handled: we always follow an inversion by a concatenation, while the original AC-moves allow for standalone inversion moves.
	The original inversion moves may be retrieved from the new generators through a sequence of transformations.
	Consider, for example, a two-generator presentation $\angles{x_1, x_2 \mid r_1, r_2}$. The sequence of moves $r_2 \to r_2 r_1$, $r_1 \to r_1 r_2^{-1}$, $r_2 \to r_2 r_1$, and $r_2 \to r_1 r_2 r_1^{-1}$ results in the presentation $\angles{x_1, x_2 \mid r_2^{-1}, r_1}$, which is the same as $r_2 \to r_2^{-1}$ up to swapping the two relators.
	We also enhanced the notion of trivial presentation(s) with $n$ generators to include all presentations of length $n$. For $n=2$, these are $\{\angles{x_1, x_2 \mid x_i^{a}, x_j^{b}} \mid i, j = 1, 2; a, b = \pm 1; i \neq j \}$.
}
The reason for this change is due to its effect on performance in greedy search and reinforcement learning algorithms studied in \cref{sec:search} and \cref{sec:rl}.
Specifically, the length of a presentation provides a useful signal when searching through the space of presentations with these algorithms.
An inversion transformation leaves the length invariant providing no signal to the search process and slowing down the performance of the algorithm significantly.
For the rest of the paper we will refer to the new transformations (instead of the original AC-transformations) as ``AC-transformations" or ``AC-moves".

Second, in order to make the search space finite in size, we set a maximum length that each relator is allowed to take. In other words, we mask all AC-moves that lead to presentations with relators exceeding this maximum length.
In the search of a sequence of AC-moves that trivialize a presentations of the Miller--Schupp series $\MS(n, w)$, we set this maximum length to be $2 \times \text{max}(2 n+3, \length(w)+1) + 2$.
This specific choice was made to allow for at least one concatenation move followed by a conjugation move in the search process.
	% !TEX root = ../ac_paper.tex

\section{Classical search algorithms}\label{sec:search}

In this section, we compare the effectiveness of breadth-first and greedy search algorithms to AC-trivialize presentations in the Miller--Schupp series.
We find that the latter significantly outperforms the former. Based on our empirical observations, we find AC-triviality of infinite subfamilies of the Miller--Schupp series, as well as length reductions of $\AK(n)$. 

\subsection{Breadth-first search}

In breadth-first search (See \cref{alg:bfs} for pseudocode), we start with an initial state, which is a balanced presentation we aim to AC-trivialize, and place it in a queue. At each iteration, a state is removed from the queue, and its neighbors are added if they have not already been visited. This process continues until the sought-after state, i.e., a trivial balanced presentation, is found or a maximum number of states $N$ is visited. In our experiments, we set $N = 10^6$.

\begin{algorithm}
	\caption{Breadth-First Search}\label{alg:bfs}
	\begin{algorithmic}[1] % The number [1] ensures lines are numbered
		\State \textbf{Input:} A balanced presentation $\pi$, maximum number of states to visit $N$
		\State \textbf{Output:} Boolean for whether an AC trivialization is found
		\State Initialize a queue $Q$ and enqueue the starting node $\pi$
		\State Mark $\pi$ as visited
		\While{Number of visited states is less than $N$}
		\State $u \gets Q$.dequeue() \Comment{Remove the front node of $Q$}
		\For{each neighbor $v$ of $u$}
		\If{$v$ is a trivial state}
		\State \Return True \Comment{Return True if $v$ is a trivial state}
		\EndIf
		\If{$v$ has not been visited}
		\State Mark $v$ as visited
		\State $Q$.enqueue($v$) \Comment{Add $v$ to the queue}
		\EndIf
		\EndFor
		\EndWhile
		\State \Return False \Comment{Return False if no trivial state is found}
	\end{algorithmic}
\end{algorithm}

\subsection{Greedy search}\label{ss:greedy_search}

Greedy search, described in \cref{alg:gs}, differs only slightly from breadth-first search. We replace the queue with a priority queue, which stores the states in order determined by a tuple of values $(k, l)$, where $k$ is the length of the presentation and $l$ is the path length between the state and the initial state.

Instead of dequeuing the earliest state, the algorithm dequeues the state with the smallest value of $k$. If there is more than one state in the priority queue with the same value of $k$, the state with the smallest value of $l$ is chosen.

\begin{algorithm}
	\caption{Greedy Search}\label{alg:gs}
	\begin{algorithmic}[1] % The number [1] ensures lines are numbered
		\State \textbf{Input:} A balanced presentation $\pi$ of length $k$, maximum number of states to visit $N$
		\State \textbf{Output:} Boolean for whether an AC trivialization is found
		\State Initialize a \textit{priority} queue $Q$ ordered by $(k, l)$ and enqueue the starting node $\pi$.
		$l$ is the length of the path connecting $\pi$ to the current node.
		\State Mark $\pi$ as visited
		\While{Number of visited states is less than $N$}
		\State $u \gets Q$.dequeue() \Comment{Remove the front node of $Q$}
		\For{each neighbor $v$ of $u$}
		\If{$v$ is a trivial state}
		\State \Return True \Comment{Return True if $v$ is a trivial state}
		\EndIf
		\If{$v$ has not been visited}
		\State Mark $v$ as visited
		\State $Q$.enqueue($v$) \Comment{Add $v$ to the queue}
		\EndIf
		\EndFor
		\EndWhile
		\State \Return False \Comment{Return False if no trivial state is found}
	\end{algorithmic}
\end{algorithm}

\subsection{Comparison of performance on Miller--Schupp series}\label{sec:search-ms}

\begin{figure}
	\centering
    \includegraphics[width=0.6\textwidth]{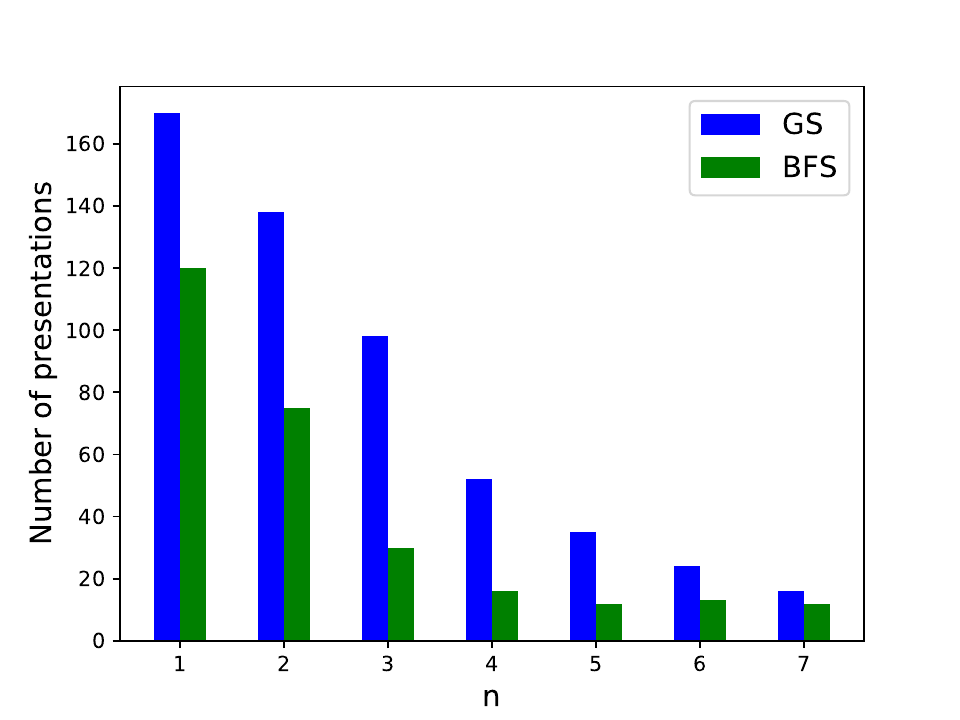}
    \caption{Comparison of greedy and breadth-first search algorithms as a function of $n$.
    The number of presentations of the Miller--Schupp series, $\MS(n, w)$, solved by an algorithm is given on the vertical axis.}
	\label{fig:performance_of_search_vs_n}
\end{figure}

We find that greedy-search outperforms breadth-first search in the task of AC-trivializing Miller--Schupp presentations. %\cref{fig:performance}
Out of the 1190 presentations in the Miller--Schupp series with $n \leq 7$ and $\length(w) \leq 7$, greedy search solved 533 while BFS solved only 278.
Each algorithm was constrained to visit a maximum of 1 million nodes.
The percentage of presentations solved by these algorithms decreases monotonically as a function of $n$, with greedy search being able to solve all presentations with $n=1$ (see \cref{fig:performance_of_search_vs_n}).
This empirical result encouraged us to prove AC-triviality of the infinite family $\MS(1, w)$, which we give in \cref{sec:ms1w_}.

Greedy search was also able to solve all presentations of length less than 14.
At length 14, there are six presentations that greedy search could not solve. We verified that four of these,
\[
\angles{x, y \mid x^{-1} y^2 x = y^{3} , x = x^{-2} y^{-1} x^2 y^{\pm 1}}
\]
\[
\angles{x, y \mid x^{-1} y^3 x = y^{4} , x = y^{\pm 1} x^2 y^{\pm 1}}
\]
are AC-equivalent to $\AK(3)$, while the other two
\[
\angles{x, y \mid x^{-1} y^2 x = y^{3} , x = y x^2 y^{\pm 1} x^{-2}}
\]
could be related neither to $\AK(3)$ nor to the trivial presentation through any path involving only presentations with both relators having length less than 20.

\begin{figure}
	\centering
	\begin{subfigure}[b]{0.5\textwidth}
		\includegraphics[width=\textwidth]{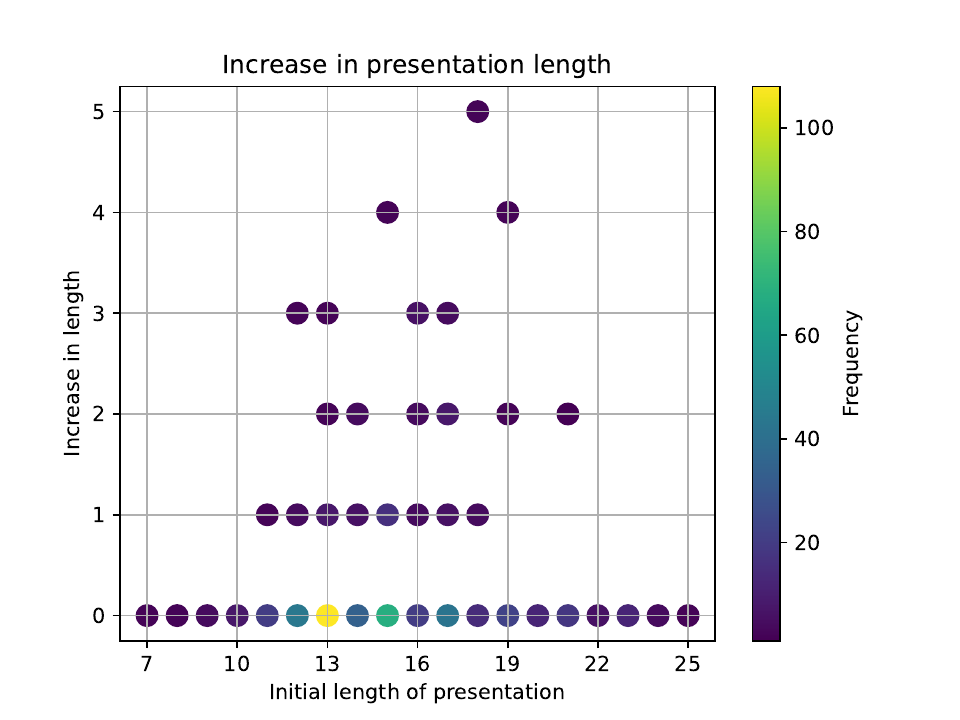}
		\caption{Distribution versus initial presentation length.}
		\label{fig:gs_length_increase_vs_length}
	\end{subfigure}%
	%add desired spacing between images, e. g. ~, \quad, \qquad etc.
	%(or a blank line to force the subfigure onto a new line)
	\begin{subfigure}[b]{0.5\textwidth}
		\centering
		\includegraphics[width=1.1\textwidth]{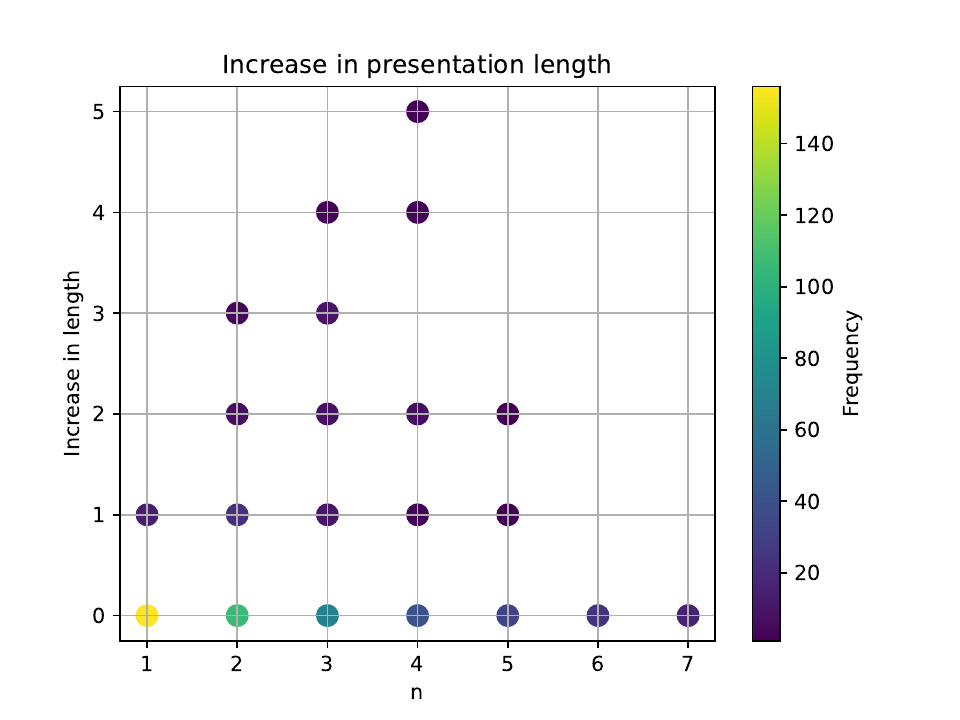}
		\caption{Distribution versus $n$.}
		\label{fig:gs_length_increase_vs_n}
	\end{subfigure}
	\caption{The maximum increase in the length of a presentation relative to its initial length along the AC trivialization path. The increase is plotted as a function of the initial length of the presentation on the left and as a function of $n$ on the right.} \label{fig:gs_length_increase}
\end{figure}

For presentations solved by greedy search, we plot the maximum amount by which the length of a presentation increased in an AC trivialization path in \cref{fig:gs_length_increase}.
In most cases, there was no increase in length; and the maximum increase was only~5.
At first glance, this seemed surprising to us, given that we allowed the relator lengths to increase by a much larger amount in our search process.\footnote{The length of each relator was allowed to increase up to \(2 \times \text{max}(2n+3, \length(w)+1) + 2\), which is twice the maximum of the initial lengths of the two relators in a presentation, plus an additional 2.
The maximum possible increase in presentation length is twice this number minus the original length.
For $n \leq 7$ and $\length(w) \leq 7$, this value lies in the range $[17, 53]$.}
However, the hard cutoff set by visiting a maximum of only 1 million nodes ensures that any presentation that needs to be mapped to a much longer presentation before it is trivialized would remain unsolved by the greedy search algorithm.
This limitation could be cured either by increasing the number of maximum nodes (at the cost of higher memory use) or by using a different criterion to order nodes in the priority queue.
It will be useful to explore the latter approach perhaps by looking for a criterion itself using deep learning algorithms.

\begin{figure}
	\centering
	\begin{subfigure}[b]{0.4\textwidth}
		\includegraphics[width=\textwidth]{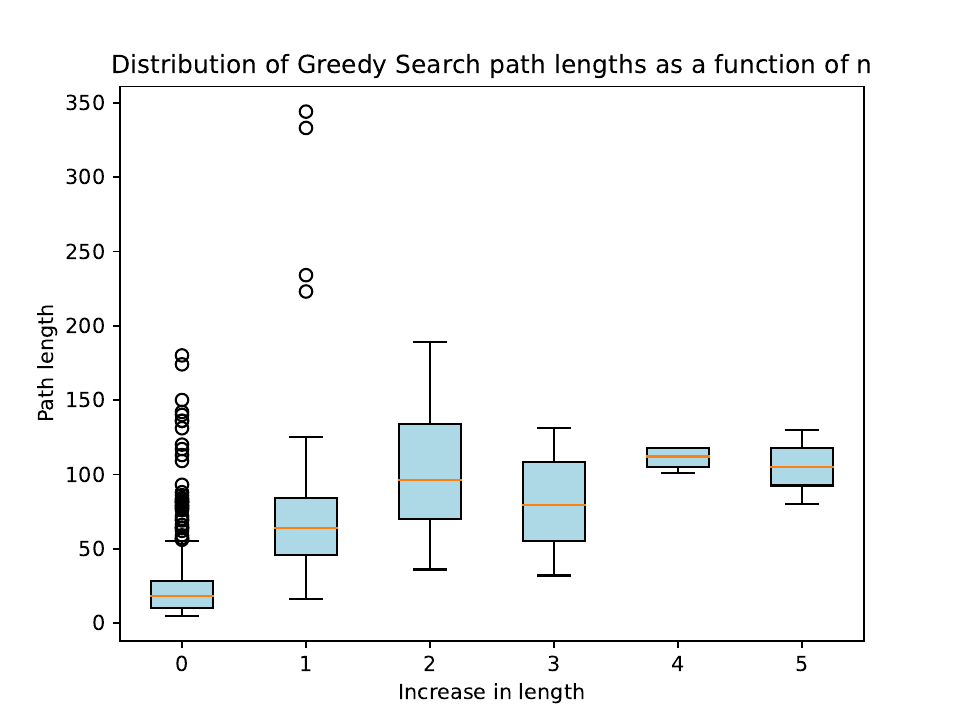}
		\caption{Distribution versus maximum increase in presentation length.}
		\label{fig:path_lengths_vs_length_increase}
	\end{subfigure}
	\begin{subfigure}[b]{0.4\textwidth}
		\centering
		\includegraphics[width=1.1\textwidth]{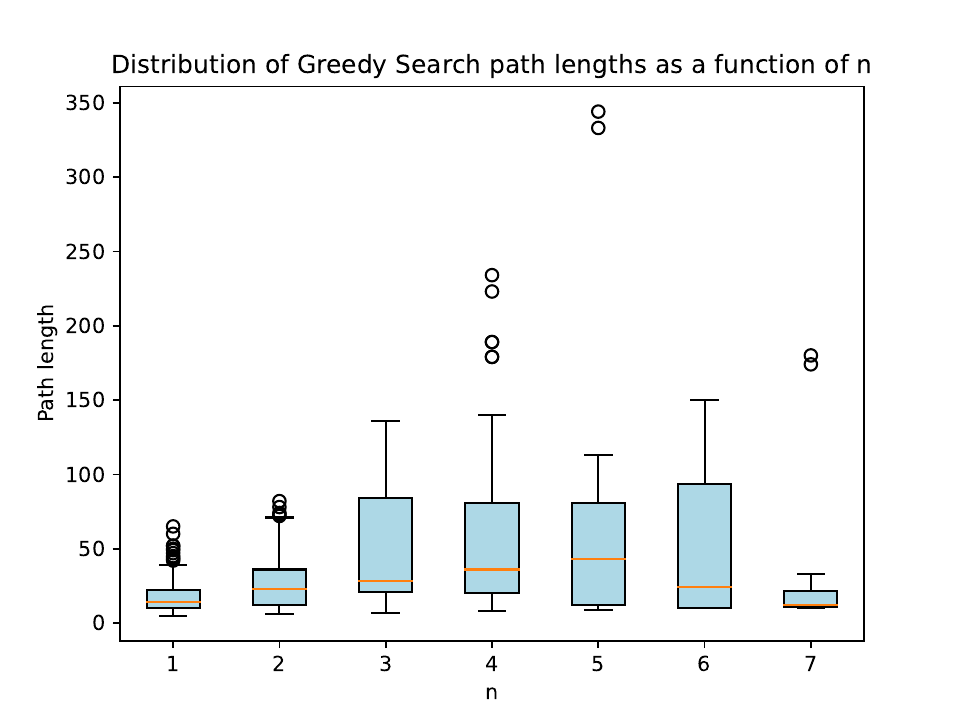}
		\caption{Distribution versus $n$.}
		\label{fig:gs_path_lengths}
	\end{subfigure}%
	\caption{Distribution of lengths of AC-trivialization paths learned by greedy search as a function of maximum increase in presentation length (left) and $n$ (right).} \label{fig:gs_path_length}
\end{figure}

We also plot the lengths of AC sequences discovered by greedy search as functions of $n$ and the maximum increase in the presentation length (\cref{fig:gs_path_length}).
Unsurprisingly, path lengths increase proportionally with the increase in the length of the presentation (\cref{fig:path_lengths_vs_length_increase}).
The following presentation with $n=5$ had the longest AC trivialization path,
\[
\angles{x, y \mid x^{-1} y^5 x = y^6,  x = y x^2 y^{-1}}
\] %correcting : \[\angles{x^{-1} y^5 x = y^6 \mid x = y x^2 y^{-1}}\]
requiring a sequence of 344 AC-moves.
Note that greedy search does not necessarily find the shortest paths of trivialization.
We will see in \cref{sec:application} that a Reinforcement Learning algorithm finds shorter trivializing sequences for many examples of the Miller--Schupp series.
This again hints at the potential utility of exploring more efficient criteria for ordering nodes in the priority queue.

In the remainder of this paper, we will refer to the presentations from the Miller--Schupp series that were solved and unsolved by the greedy search as ``GS-solved" and ``GS-unsolved" presentations, respectively. In other words, many of our experiments will be tested on two datasets that consists of Miller--Schupp presentations with $n \leq 7$ and $\length(w) \leq 7$: the GS-solved dataset has 533 presentations, whereas GS-unsolved dataset has 657 presentations. The combined dataset that contains all presentations with $n \leq 7$ and $\length(w) \leq 7$ has size 1190.

\subsection{AC-triviality of $\MS(1, w)$} \label{sec:ms1w_}

As mentioned in the previous subsection, the greedy search could solve all $n=1$ presentations of the Miller--Schupp series in our dataset. This motivated us to prove AC-triviality for all $\MS(1, w)$. 

In this and the following subsections, we frequently use the \emph{substitution} transformation mentioned in Section~\ref{sec:AC}.

\begin{theorem}\label{t:ms1wt}
    $\MS(1,w)$ is AC-trivial for all $w$.
\end{theorem}

\begin{proof}
We have 
\[
\MS(1, w) = \angles{x, y \mid x^{-1} y x = y^{2}, x = w}
\]
where $w$ has exponent sum 0 in $x$. We can re-write the first relator in the following four ways: 
\begin{enumerate}[label=(\roman*)]
    \item $yx=xy^2$
    \item $x^{-1}y = y^2 x^{-1}$
    \item $x^{-1}y^{-1}=y^{-2}x^{-1}$
    \item $y^{-1}x = xy^{-2}$
\end{enumerate}

We can substitute these equations in the second relator to move around all occurrences of $x$ and $x^{-1}$: relations (i) and (iv) move $x$ to the left, relations (ii) and (iii) move $x^{-1}$ to the right. We continue until the second relator becomes $x^{a-1} y^b x^{-a}$ for some $a, b \in \mathbb{Z}$. Through conjugation, this simplifies to $x = y^b$, which when substituted into the first relator leads to a trivialization of the presentation.
\end{proof}

The following result follows from this theorem.

\begin{theorem}\label{t:msnw}
    $\MS(n, w_\star)$ for $w_\star = y^{-1} x y x^{-1}$ is AC-trivial for all $n$.
\end{theorem}

\begin{proof}
In the presentation,
\[
\MS(n, w_\star) = \angles{x, y \mid x^{-1} y^n x = y^{n+1}, x = y^{-1} x y x^{-1} }
\]
we can rewrite the second relation as $y^{-1} x y = x^2$, and apply the automorphism $x \leftrightarrow y$ to get
\[
\MS(n, w_\star) = \angles{x, y \mid
x^{-1} y x = y^2 , y^{-1} x^n y = x^{n+1}}.
\]
This presentation is the same as $\MS(1, y^{-1} x^n y x^{-n})$, which is AC-trivial by \cref{t:ms1wt}. 
\end{proof}

AC-trivializations of $\MS(n, w_\star)$ for $n = 3, 4, 5, 6, 7, 8$ were recently obtained using automated theorem proving in \cite{new-ac-for-ms}. Here, we have obtained AC-trivializations of this family for all $n$.

% Indeed, to see how to realize this with AC moves, consider by means of example the application of (1) to $w_1yxw_2$. First, we cycle the relator to get $yxw_2w_1$, then invert it to get $w_1^{-1}w_2^{-1}x^{-1}y^{-1}$, and then multiply by (1) written as $yxy^{-2}x^{-1}$ onto the end of the relator, giving $w_1^{-1}w_2^{-2}y^{-2}x^{-1}$. Inverting again and cycling gives us $w_1xy^2w_2$. 

% After moving all positive powers of $x$ to the left and negative powers of $x$ to the right, we can cancel them with (AC3) by conjugating by $x^{-1}$. We are then left with $y^kx^{-1}$ for some $k$, since the exponent sum on $x$ in the second relator was $-1$ to begin with and is unchanged throughout these moves. Substituting this into the first relation, we get $y=y^2$ or $y=1$, from which we can easily trivialize the second relator as well. 

\subsection{Length reduction for $\AK(n)$}
\label{sec:len-reduction}

In \cref{sec:search-ms}, we discussed the performance of classical search algorithms for presentations of the Miller--Schupp series. But what if we apply these algorithms for presentations of the Akbulut--Kirby series instead? We find that starting from $\AK(5)$---a presentation of length $17$---and performing greedy search for $20$ million nodes, reduces its length to 16. Examining the sequence of AC moves connecting the two presentations, we found the following result.

\begin{theorem}\label{t:n+11}
	For every $n\geq 2$, $\AK(n)$ is AC-equivalent to the presentation
	\[
	\angles{ x,y \mid x^{-1} y x = x y x^{-1} y \ ,\  xy^{n-1}x = yxy },
	\]
	of length $n+11$. This gives reduction in length of $AK(n)$ for all $n \geq 5$.
\end{theorem}

To prove this theorem, we will  use the following result due to \cite{MMS}, and prove two additional theorems (\cref{t:mswk_wk1} and \cref{t:ms-short}). From these results, \cref{t:n+11} follows immediately. 

\begin{proposition}
    [Myasnikov, Myasnikov, and Shpilrain, \cite{MMS}]\label{t:MS-AK-MS}
    For all $n \geq 2$, $AK(n)$ is AC-equivalent to $\MS(n, w_1)$ where $w_1 = y^{-1} x^{-1} y x y$.
\end{proposition}

\begin{theorem}\label{t:mswk_wk1}
    For each fixed $n > 0$, the 1-parameter family of presentations $\MS(n, w_k)$ are all AC-equivalent. Here, $w_k = y^{-k} x^{-1} y x y$ is parameterized by $k \in \mathbb{Z}$.
\end{theorem}

\begin{proof}
    Starting with the presentation,
        \[
\MS(n, w_k) = \angles{x, y \mid x^{-1} y^n x = y^{n+1}, x = y^{-k} x^{-1} y x y},
\]
    we will show that for any fixed $n$ and $k$, $\MS(n, w_k)$ is AC-equivalent to $\MS(n, w_{k+1})$.

    First, note that the first relation in $\MS(n, w_k)$ can be rearranged in the following three ways:
    \begin{enumerate}[label=(\roman*)]
        \item Multiplying by $y^{k-n}x$ from the left and by $y^{-1}$ from the right gives 
        \[
            y^kxy^{-1}=y^{k-n}xy^n.
        \]
        \item Multiplying by $y^{-1}$ from the left and by $x^{-1}y^{-1}$ from the right, and inverting the relation gives 
        \[
            y^{-(n-1)}xy = yxy^{-n}.
        \]
        \item Multiplying by $y^{-1}x$ from the left and by $y^{k-n}$ from the right, and inverting the relation gives
        \[
            y^{n-k}x^{-1}y^{-(n-1)}=y^{-(k+1)}x^{-1}y.
        \]
    \end{enumerate}

    Now, we can rearrange the second relation  to get $y^k x y^{-1}=x^{-1}yx$. Substituting (i) gives $y^{k-n} x y^n = x^{-1}yx$, which  we can rewrite as $yxy^{-n}=xy^{k-n}x$. Now substituting (ii) gives $ y^{-(n-1)}xy= xy^{k-n}x$, which we may rewrite as $y^{n-k}x^{-1}y^{-(n-1)}=xy^{-1}x^{-1}$. Finally, substituting (iii) gives $y^{-(k+1)}x^{-1}y=xy^{-1}x^{-1}$, which is equivalent to $x=y^{-(k+1)}x^{-1}yxy=w_{k+1}$. 
\end{proof}

\begin{theorem}\label{t:ms-short}
    For each $n > 0$ and $k \in \mathbb{Z}$, $\MS(n, w_k)$ is AC-equivalent to the presentation
    \[P(n, k) = \angles{x, y \mid  y^{n-k-1}  x^{-1} y x = x y x^{-1} y^{n-k} \ , \ x = y^{-k} x^{-1} y x y },
    \]
    of total length $|k| + |n - k| + |n - k - 1| + 11$. For each fixed $n$, this length is minimum when $k=n-1$, which gives the presentation
    \[P(n, n-1) = \angles{x, y \mid   x^{-1} y x = x y x^{-1} y \ , \ x = y^{-(n-1)} x^{-1} y x y },
    \]
    of length $n+11$.
\end{theorem}

\begin{proof}
    Starting with the presentation,
        \[
        \MS(n, w_k) = \angles{x, y \mid x^{-1} y^n x = y^{n+1}, x = y^{-k} x^{-1} y x y},
        \]
    the second relation may be equivalently expressed as 
    \begin{enumerate}[label=(\roman*)]
        \item $y^{k}xy^{-1}=x^{-1}yx$
        \item $y^{-1}xy^{k}=xyx^{-1}$
    \end{enumerate}
    Multiplying the first relation by $y^{-1} x$ from the left and by $y^{-1}$ from the right, we get $y^{n-1} x y^{-1} = y^{-1} x y^{n}$, which we may rewrite as $y^{n-k-1} \left(y^k x y^{-1}\right) = \left(y^{-1} x y^{k}\right) y^{n-k}$. Substituting from (i) on the LHS and from (ii) on the RHS gives the required result.
\end{proof}

Combining the previous three results shows that, for each fixed $n$, $AK(n)$ is AC-equivalent to all members of the infinite families $\MS(n, w_k)$ and $P(n, k)$. From this, \cref{t:n+11} follows immediately from setting $k = n - 1$.

\begin{remark}
While $P(n, n-1)$ is always the shortest presentation from the $P$ family, other $P(n, k)$ may also give reductions in length of $\AK(n)$. To find which presentations of $P$-family are shorter than $\AK(n)$, we compare the length of $\AK(n)$, i.e. $2n + 7$, with the length of $P(n, k)$ given by:
\[
    \begin{cases} 
        2n + 10 - k, & \text{if } k \leq n-1, \\
        3k + 12 - 2n, & \text{if } k \geq n.
    \end{cases}
\]
This is less than $2n+7$ when $4 \leq k \leq n-1$ or $n \leq k < \frac{1}{3} \left(4n-5 \right)$.
\end{remark}

We also note that the AC-triviality of $\AK(2)$, when combined with \cref{t:mswk_wk1}, gives the following result. 
\begin{corollary}\label{c:ms2wk}
Each member of the infinite families $\MS(2, w_k)$ and $P(2, k)$ is AC-trivial.
\end{corollary}

\subsection{Limitations and extensions}

While the greedy search algorithm performs better than the breadth-first search, it has some of the same limitations.
Namely, it is memory inefficient, and we cannot leverage the parallelizability of modern hardware architectures.
It also does not learn a general algorithm that would find an AC trivialization for any given balanced presentation.

Reinforcement learning algorithms, particularly policy gradient algorithms, present a promising alternative that avoids these downsides.
These algorithms are memory efficient and can be trained in a highly distributed manner, which we will focus on in the next section.

Another interesting direction to pursue involves changing the sorting function used by greedy search. Consider the family of greedy search algorithms where the priority queue is sorted by a fixed function $f: (r_1, r_2) \to \mathbb{R}$ which takes a presentation and outputs a real number\footnote{We can also reinterpret breadth-first search in this manner if, given as input to $f$ the path length $l$, we have the function return $l$. From this perspective it becomes clear that, whenever breadth-first search succeeds, it finds a path of minimal length.}. The greedy search algorithm used here is based on the function $f(r_1, r_2) \to |r_1| + |r_2|$. An interesting option would be to use machine learning to try and learn a good function $f$. This idea is inspired by search-based reinforcement learning systems, such as AlphaGo and AlphaZero, which combine Monte Carlo Tree Search with learned heuristics to evaluate and prioritize states dynamically. While these systems go beyond classical greedy search by incorporating exploration and planning through tree-based rollouts, the core principle of leveraging a learned heuristic function to guide search is analogous.
	% !TEX root = ../ac_paper.tex

\section{Reinforcement learning}\label{sec:rl}

This section is organized as follows: in \cref{sec:mdp}, we discuss how the problem underlying the Andrews--Curtis conjecture can be formulated as a Markov Decision Process.
In \cref{sec:ppo}, we discuss details of a specific reinforcement learning algorithm, called the Proximal Policy Optimization algorithm, which we used to find AC trivializations of balanced presentations.
Finally, in \cref{sec:application}, we discuss the results of our work, comparing the performance of PPO with that of the classical search algorithms studied in the previous section.

\subsection{Markov decision process} \label{sec:mdp}

A \textit{Markov Decision Process (MDP)} is defined as a 5-tuple $(S, A, R, P, \rho)$. Here, $S$ represents the space of states, while $A$ denotes the set of actions, where each action $a \in A$ is a function mapping from one state to another, i.e., $a \colon S \to S$. The function $R \colon S \times A \times S \to \R$ is the reward function, which assigns a real-valued reward based on the transition from one state to another via a specific action. The transition probability function, denoted by $P \colon S \times A \to \mathcal{P}(S)$, provides the probability distribution over the possible next states given a current state and action. Lastly, $\rho$ represents the initial probability distribution of states, describing the likelihood of the system starting in each state.

\begin{figure}[ht]
	\centering
	\includegraphics[trim={0.0in 3.0in 0.0in 2.5in},clip,width=4.0in]{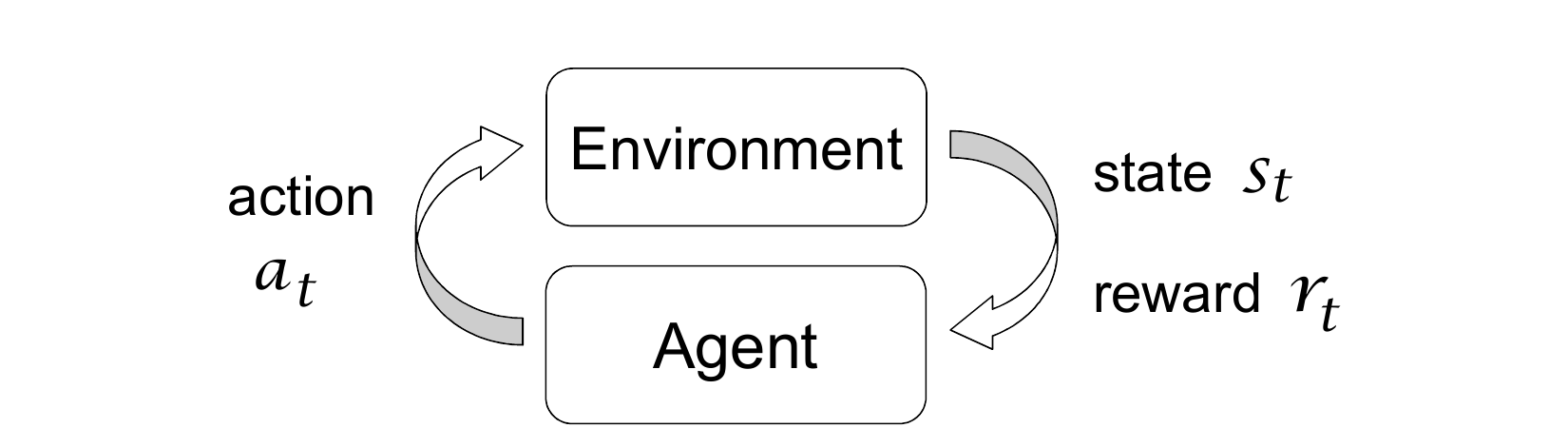}
	\caption{The basic RL cycle.}
	\label{fig:RLbasic}
\end{figure}

A schematic picture of how these objects interact with each other is as follows. We start with a state $s_0$ sampled from the distribution $\rho$ and take an action $a_0$. This results in a state $s_1$ with probability $P(s_1 \mid s_0, a_0) $. The transition gets a ``reward" $r_0 = R(s_0, a_0, s_1)$ which quantifies the effectiveness of the action in contributing toward achieving an ultimate goal. From state $s_1$, we repeat this process, obtaining a trajectory of states
\[
\tau = \left( s_0, a_0, s_1, a_1, \dots \right).
\]
The goal of this process is to maximize the cumulative return,
\[
R(\tau) = \sum\limits_{t=0}^{T} \gamma^t R(s_t, a_t, s_{t+1}).
\]
Here, $T$ is the length of the trajectory, known as the ``horizon length'' and $\gamma \in \left(0, 1 \right)$ is the ``discount factor'' that assigns smaller weights to the reward values obtained farther in the future.

For a given problem at hand, we may not \textit{a priori} know the actions $\{a_t\}$ and states $\{s_{t+1}\}$ that maximize the return. Deep reinforcement learning presents a solution to this problem: we train a neural network that learns a map from states to actions with the objective of maximizing the cumulative return. More precisely, we learn a map called the ``policy" function $\pi \colon S \to \mathcal{P}(A)$ that assigns to each state a probability distribution over actions. At time step $t$, an action $a_t \sim \pi(\cdot \mid s_t)$ is sampled that gives the next state $s_{t+1}$.\footnote{
	In general, we need to specify the probability transition function $P(\cdot \mid s_t, a_t)$ from which the next state would be sampled. In this paper, we take this probability distribution to be a delta function centered at a single state, which we write as $a_t(s_t)$.}
In the next subsection we discuss the specific algorithm and the objective function that we used in our study.

\subsection{Proximal policy optimization} \label{sec:ppo}

The goal of our optimization process is to find a policy that maximizes the cumulative return. The most naive way to achieve this goal is through an algorithm known as ``vanilla policy gradient.''
We perform gradient updates guided by the expected return $J(\pi_\theta) \equiv \mathbb{E}_{\tau \sim \pi_\theta} R(\tau)$, where the expectation is over a set of trajectories consisting of states and actions sampled according to our current policy,
\[
\theta_{k+1} = \theta_k + \nabla_\theta J(\pi_\theta).
\]

It turns out that this update depends on the gradient of the logarithm of the policy function itself and an ``advantage function'' \( A^\pi (s, a) \) which quantifies the relative benefit of taking action \( a \) in state \( s \) under policy \( \pi \) \cite{SpinningUp2018}.\footnote{
	Mathematically, the advantage function is the difference between ``on-policy action-value function" $Q^{\pi}(s, a) = \mathbb{E}_{\tau \sim \pi_\theta} [R(\tau) \mid s_0 = s, a_0 = a]$ and the ``on-policy value function", $V^{\pi}(s) = \mathbb{E}_{\tau \sim \pi_\theta} [R(\tau) \mid s_0 = s]$.
	These functions give an estimate of the cumulative return given that the current state-action pair is $(s, a)$ in the case of $Q^\pi$ and the current state is $s$ in the case of $V^{\pi}$.
	\label{ft:advantage}}
Explicitly,
\[
\nabla_\theta J(\pi_\theta) =
\mathbb{E}_{\tau \sim \pi_\theta} \left[ \sum\limits_{t=0}^T \nabla_\theta \log \pi_\theta (a_t \mid s_t) A^{\pi_\theta} (s_t, a_t) \right].
\]
Thus, vanilla policy gradient algorithm amounts to optimizing the objective function
\[
L^{PG} = \mathbb{E}_{\tau \sim \pi_\theta} \left[ \sum\limits_{t=0}^T \log \pi_\theta (a_t \mid s_t) A^{\pi_\theta} (s_t, a_t) \right].
\]
Although it has the advantage of being simple, this algorithm has the drawback of making excessively large updates, which can cause the updated policy to deviate significantly from the current policy.

Proximal Policy Optimization (PPO) algorithms seek to make these updates more robust by limiting the extent to which the policy \( \pi \) can change in a single update \cite{schulman2017proximal}.
PPO implements this through an objective function that includes a clipped probability ratio between the new policy \( \pi_\theta \) and the old policy \( \pi_{\text{old}} \), thus constraining the updates within a predefined range.
\[
L^{CLIP}(\theta) = \mathbb{E}_{t} \left[ \min(r_t(\theta) A_t, \text{clip}(r_t(\theta), 1 - \epsilon, 1 + \epsilon) A_t) \right]
\]

Here \( r_t(\theta) = \frac{\pi_\theta(a_t | s_t)}{\pi_{\text{old}}(a_t | s_t)} \) represents the probability ratio, and \( \epsilon \) is a small positive constant (usually set around 0.1 or 0.2) that controls the clipping value.
The raw and the clipped ratio ensure that excessively large policy updates are curtailed, making the optimization process more stable.

The advantage function is estimated during the optimization process using the Generalized Advantage Estimation method of \cite{schulman2018highdimensional}.
This method requires an estimation of the on-policy value function (c.f. \cref{ft:advantage}).
In PPO, the actor-critic framework is used for this purpose. We train two neural networks in parallel, where the ``actor" learns the policy $\pi_\theta$ and the ``critic" learns the value function.\footnote{
	Sometimes, these neural networks are made to share their parameters, which helps in stability during training.
	We experimented with shared as well as unshared parameters and observed better performance by keeping the parameters independent.}
The value loss $L^V$ is the mean squared error between the values for a state as estimated by the critic network before and after a gradient update.
Lastly, PPO adds entropy $S$ of the action-space distribution to the full loss function to prevent premature convergence to a suboptimal policy. The full loss function is as follows,
\[
L = \mathbb{E}_{t} \left[ L^{CLIP}(\theta) - c_1 L^{V}(\phi) + c_2 S(\pi_\theta)(s_t) \right]
\]
where $c_1$ and $c_2$ are tunable hyperparameters.

In policy gradient algorithms such as PPO, we alternate between collecting data through the current policy in the ``rollout phase" and updating the policy weights through gradient descent in the ``optimization phase". There are various hyperparameters associated to these phases, which we briefly describe now. The rollout phase involves generating multiple trajectories of the underlying Markov Decision Process in parallel using the current policy function.
The hyperparameters of this phase include the number of parallel actors $N$, the rollout length $T'$, i.e. the number of steps each actor takes, the discount factor $\gamma$ (c.f. \cref{sec:mdp}) and a bootstrapping hyperparameter of Generalized Advantage Estimation $\lambda_{\text{GAE}}$. The dataset of $N \times T'$ examples collected in this phase is then used to update the policy. In each epoch, the dataset is shuffled and split into mini-batches before performing gradient descent.
In addition to the optimizer hyperparameters (such as learning rate schedule), the number of epochs and the size of mini-batches are hyperparameters of the optimization process.
Further details of hyperparameters are summarized in \cref{app:hyperparameters}.

\subsection{Application to the MS series}\label{sec:application}

The set of all balanced presentations of the trivial group with a fixed number of generators and the set of AC-transformations play the roles of sets $S$ and $A$, respectively. (Recall notations introduced in \cref{sec:mdp}). Once we choose a reward function $R$ and an initial state distribution $\rho$, we may use the Proximal Policy Optimization algorithm to learn the policy function $\pi$. We tested a few different reward functions in our experiments, observing that the following candidate led to the best performance and stability in training.
\[
R(s_{t}, a_{t}, s_{t+1}) =
\begin{cases}
	-\min(10, \ \length(s_{t+1})) & \text{if length}(s_{t+1}) > 2, \\
	\hfil 1000 & \text{otherwise.}
\end{cases}
\]
Here, $\length(s_{t+1})$ is the length of the presentation at timestep $t+1$. The reward function assigns $-\min(10, \length(s_{t+1}))$ to a non-terminal state and $1000$ to a terminal state. We found that clipping the reward to $-10$ led to less variance in gradients of the loss function with respect to weights.

We define the initial state distribution as a distribution over the presentations of the Miller--Schupp series with $n \leq 7$ and $\length(w) \leq 7$. Initially, each presentation was selected exactly once in ascending order by $n$ and $\length(w)$. Following this initial sequence, we maintained an ongoing record of presentations that were either solved or unsolved at any given time. During each rollout phase, a presentation was randomly chosen from the set of solved or unsolved presentations with probabilities of $\frac{1}{4}$ and $\frac{3}{4}$, respectively. This method was designed to allow the policy network to refine its strategies on presentations that were already solved, potentially discovering shorter sequences of AC-moves, while also tackling presentations that remained unsolved.

We also need to choose a horizon length $T$ over which the cumulative return is calculated (c.f. \cref{sec:mdp}). In reinforcement learning, training on tasks with long horizons is usually harder than on those with short horizons. This is because in long horizon tasks it is often more difficult to figure out which actions are responsible for the eventual outcomes, a problem known as the credit assignment problem. There is also more uncertainty in the returns (high variance), which can slow down convergence and destabilize the learning process. The size of the exploration space also becomes a critical issue. With longer horizons, the agent needs to explore a more diverse set of actions to learn about long-term consequences without getting stuck in suboptimal regions. However, it does not mean that we can not train well-performing agents for long-horizon tasks; rather, it indicates that with longer horizons we may need significantly stronger computational power and extended training periods.\footnote{Another option for improving performance in long-horizon tasks is to provide good intermediate rewards; unfortunately, this is rather hard in the present context of the AC conjecture.}

The horizon length $T$ is an important hyperparameter in the context of our problem as any presentation that requires a sequence of AC-moves of length greater than the horizon length will necessarily remain unsolved by PPO. On the other hand, choosing a horizon length that is too long can significantly slow down the training process. We experimented with two different schedules for horizon length during training:
\begin{enumerate}[label=(\roman*)]
	\item constant horizon length throughout training, and
	\item varying horizon length during training.
\end{enumerate}

The trained agents behave differently in the two cases as we will now describe.

\subsection{PPO with constant horizon length: experimental results} 
\begin{figure}
	\centering
	\begin{subfigure}[b]{0.45\textwidth}
		\includegraphics[width=1.1\textwidth]{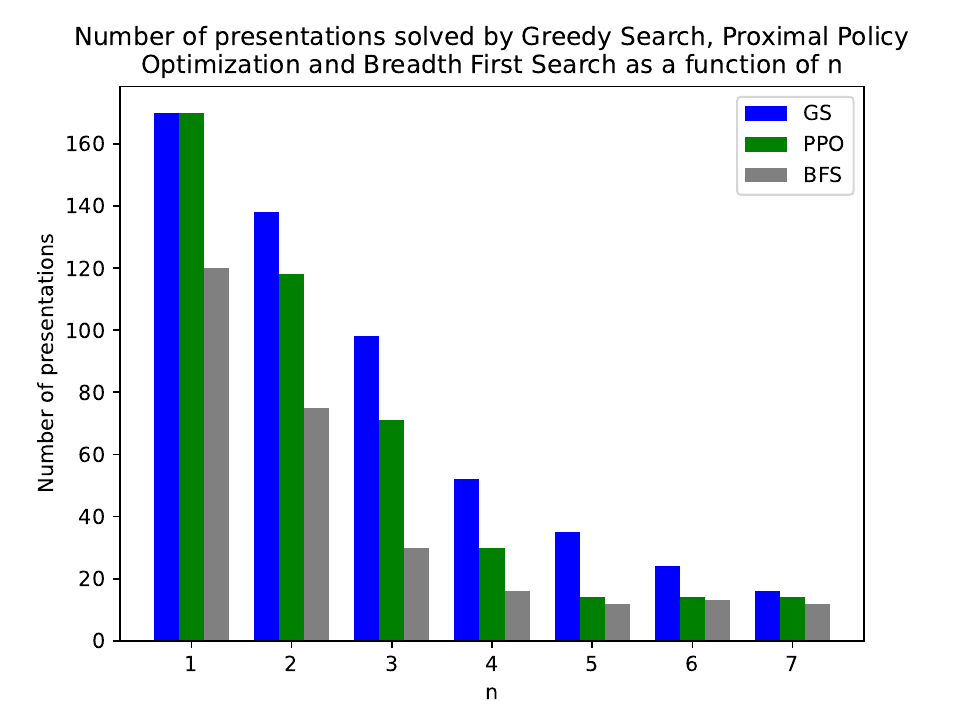}
		\caption{Distribution versus $n$}
		\label{fig:performance_vs_n}
	\end{subfigure}
	\begin{subfigure}[b]{0.45\textwidth}
		\centering        \includegraphics[width=1.1\textwidth]{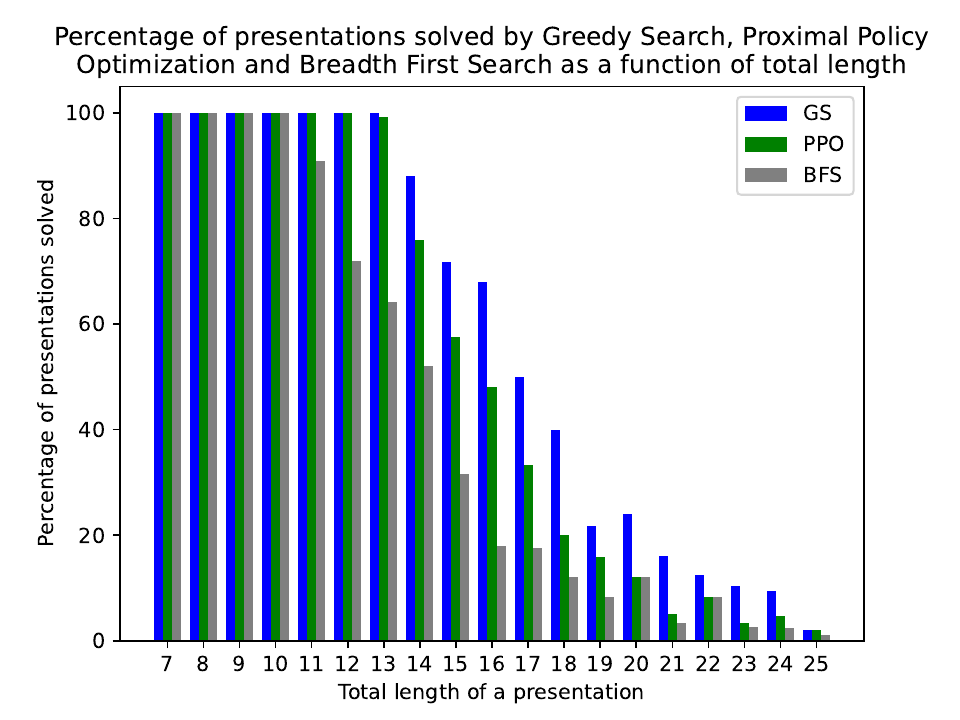}
		\caption{Distribution versus length}
		\label{fig:performance_vs_length}
	\end{subfigure}
	\caption{A comparison of three algorithms ---breadth-first search, greedy search, and Proximal Policy Optimization (PPO) with small and constant horizon length during training --- that we used to search through the space of balanced presentations. The number of presentations of the Miller--Schupp series, $\MS(n, w)$, solved by an algorithm is given on the vertical axis. We compare the performance as a function of $n$ (above) and the length of the presentation (below). Note that varying horizon length during training (not depicted in this figure) helped PPO solve presentations that greedy search could not solve.}
	\label{fig:performance}
\end{figure}

When keeping the horizon length fixed during training, we experimented with considerably smaller values, i.e. $T=200$ and $T=400$. Out of the $1190$ Miller--Schupp presentations of the initial state distribution, the best performing PPO agents could solve $431$ and $402$ presentations respectively. \footnote{We also explored the value $T=2000$. However, we found it much slower to train due to the reasons described above. We could only solve $219$ presentations of the initial state distribution in this case. This training run had not converged, and we expect that with more computational power and extended training periods, it will be worthwhile to experiment with larger values of $T$, perhaps helping us solve even more presentations than greedy search.}
In each case, these presentations formed a subset of the presentations solved by the greedy search (\cref{fig:performance}).

\begin{figure}
	\centering
	\begin{subfigure}[b]{0.5\textwidth}
		\includegraphics[width=\textwidth]{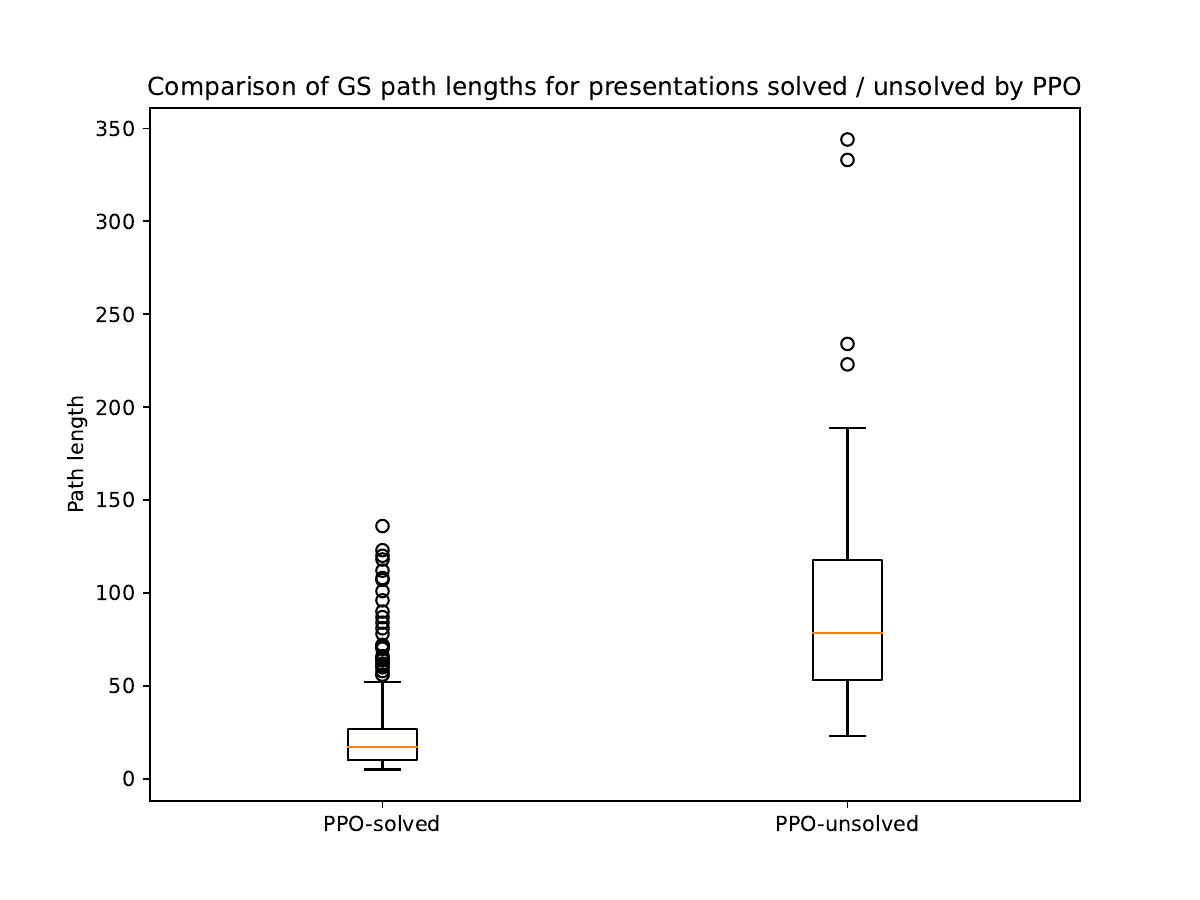}
		\caption{}
		\label{fig:path_lengths_ppo_solved_vs_unsolved}
	\end{subfigure}
	\begin{subfigure}[b]{0.5\textwidth}
		\centering
		\includegraphics[width=\textwidth]{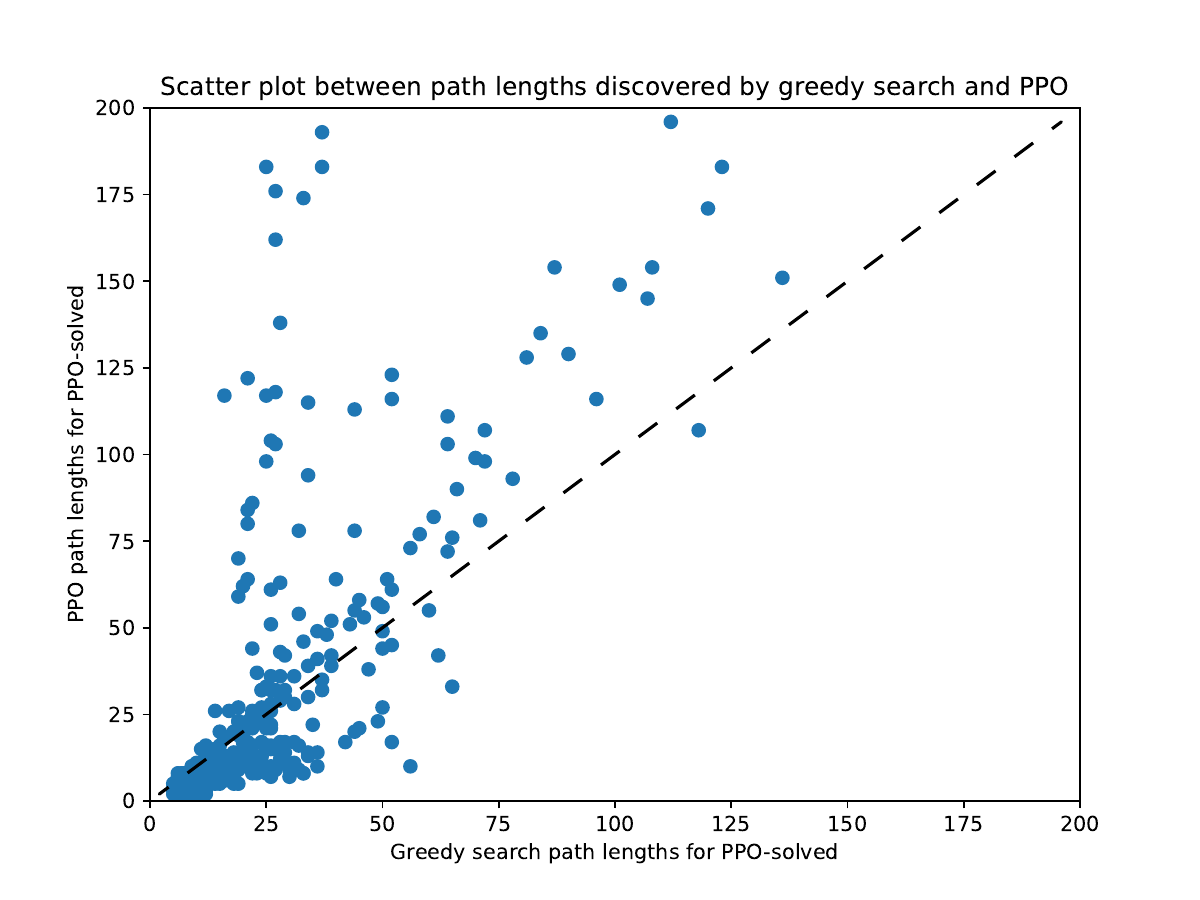}
		\caption{}
		\label{fig:path_lengths_gs_vs_ppo}
	\end{subfigure}
	\caption{A comparison of path lengths discovered by the greedy search and a PPO agent. The left panel shows the distribution of lengths of AC trivialization paths discovered by the greedy search in the cases solved / unsolved by PPO. The right panel shows the scatter plot of path lengths discovered by PPO vs path lengths discovered by the greedy search.}
	\label{fig:path_lengths_gs_vs_ppo_full}
\end{figure}

In the rest of this section, we will describe some observations made for an agent trained with the value $T=200$.\footnote{The complete list of hyperparameters for this experiment is summarized in \cref{app:hyperparameters}.}
While PPO consistently outperformed BFS for all values of $n$ and for all lengths of presentations, it consistently underperformed compared to the greedy search.
\footnote{The results in this section were obtained using relatively small neural networks for both the actor and the critic networks, each consisting of two layers with 512 hidden units. It is natural to assume that increasing the size of these networks could enable a PPO agent to outperform the greedy search. However, such experiments were not performed due to computational limitations. Future work could explore this direction to assess the impact of larger network architectures.}
In \cref{fig:path_lengths_ppo_solved_vs_unsolved}, we plot the distribution of path lengths discovered by greedy search in the two cases of presentations that could / could not be solved by PPO. It is clear that the presentations PPO solved had, in general, smaller path lengths. In particular, all of these had greedy search path lengths less than $200$.

In \cref{fig:path_lengths_gs_vs_ppo}, we give a scatterplot of the path lengths discovered by PPO and greedy search for all of the presentations solved by PPO. We note that in many cases, PPO found shorter paths compared to the greedy search. This is expected as PPO learns to improve its strategy during the training, discovering shorter paths for presentations it may have already solved. The scatterplot shows the shortest paths discovered by PPO for each presentation. We also note that in many cases, PPO found longer paths than greedy search. This shows that our specific run exhibits a suboptimal policy. It could perhaps be improved by performing more hyperparameter tuning on the training process.

\subsection{New Miller--Schupp trivializations through variable horizon length}\label{sec:variable_horizon}

Keeping the horizon length fixed either fails to solve hard presentations if the horizon length is too small, or requires significant computational resources for training to converge if the horizon length is too large. Therefore, we experimented with varying horizon length during training.

We trained a PPO agent for 100 million environment interactions, with the same hyperparameters as in the previous section; only varying the horizon length over the course of training. In the beginning, the horizon length was set to $200$; it was increased to $400$ after 10 million environment interactions, then to $800$ after 25 million environment interactions, and finally to $1200$ after 50 million environment interactions.
This schedule allowed the PPO agent to first solve easier presentations, avoiding the challenges of long-horizon tasks. As the agent’s capabilities improved, the gradual increase in horizon length enabled the agent to tackle more complex Miller-Schupp presentations.

The best-performing PPO agent solved two new presentations of the Miller-Schupp series that the greedy search had failed to solve.

\begin{theorem}\label{thm:MS}
	The following potential counterexamples belonging to the Miller--Schupp series are AC-trivial:
	\begin{gather*}
		\angles{x, y \mid x^{-1} y^3 x y^{-4} , \ x^{-1} y^{-1} x y^{-3} x^{-1} y^{-1}}, \\
		\angles{x, y \mid x^{-1} y^4 x y^{-5} , \ x^{-1} y x y^{-1} x^{-1} y^{-3}}.
	\end{gather*}
\end{theorem}

For these presentations, the AC paths discovered by the agent had lengths of $205$ and $1085$, respectively. These paths contained subsequences of moves that acted trivially. We removed these subsequences, obtaining paths of lengths $195$ and $381$ respectively. The initial lengths of the two presentations are $17$ and $19$, which increased to the maximum values of $29$ and $30$ before the presentations were trivialized.

With the following definitions of AC$'$ moves,
\[
\begin{aligned}
	h_1 &= \ r_2 \rightarrow r_2 r_1, & \quad h_5 &= \ r_2 \rightarrow x^{-1} r_2 x, & \quad h_9 &= \ r_2 \rightarrow x r_2 x^{-1}, \\
	h_2 &= \ r_1 \rightarrow r_1 r_2^{-1}, & \quad h_6 &= \ r_1 \rightarrow y^{-1} r_1 y, & \quad h_{10} &= \ r_1 \rightarrow y r_1 y^{-1}, \\
	h_3 &= \ r_2 \rightarrow r_2 r_1^{-1}, & \quad h_7 &= \ r_2 \rightarrow y^{-1} r_2 y, & \quad h_{11} &= \ r_2 \rightarrow y r_2 y^{-1}, \\
	h_4 &= \ r_1 \rightarrow r_1 r_2, & \quad h_8 &= \ r_1 \rightarrow x r_1 x^{-1}, & \quad h_{12} &= \ r_1 \rightarrow x^{-1} r_1 x, \\
\end{aligned}
\]
the length 195 path for the first of the two presentations above is as follows.
\[
\begin{aligned}
	& h_{2} \cdot h_{8} \cdot h_{10} \cdot h_{4} \cdot h_{10} \cdot h_{10} \cdot h_{10} \cdot h_{2} \cdot h_{8} \cdot h_{6} \cdot h_{6} \cdot h_{6} \cdot h_{8} \cdot h_{10} \cdot h_{10} \cdot h_{10} \cdot h_{10} \cdot h_{4} \\ &
	h_{0} \cdot h_{6} \cdot h_{2} \cdot h_{8} \cdot h_{8} \cdot h_{10} \cdot h_{4} \cdot h_{0} \cdot h_{6} \cdot h_{8} \cdot h_{10} \cdot h_{10} \cdot h_{10} \cdot h_{2} \cdot h_{8} \cdot h_{6} \cdot h_{6} \cdot h_{4} \\ &
	h_{10} \cdot h_{10} \cdot h_{10} \cdot h_{2} \cdot h_{8} \cdot h_{6} \cdot h_{8} \cdot h_{10} \cdot h_{4} \cdot h_{0} \cdot h_{6} \cdot h_{8} \cdot h_{10} \cdot h_{10} \cdot h_{2} \cdot h_{8} \cdot h_{6} \cdot h_{6} \\ &
	h_{4} \cdot h_{10} \cdot h_{10} \cdot h_{10} \cdot h_{2} \cdot h_{8} \cdot h_{6} \cdot h_{6} \cdot h_{8} \cdot h_{10} \cdot h_{4} \cdot h_{0} \cdot h_{6} \cdot h_{8} \cdot h_{10} \cdot h_{2} \cdot h_{8} \cdot h_{6} \\ &
	h_{6} \cdot h_{4} \cdot h_{10} \cdot h_{10} \cdot h_{10} \cdot h_{2} \cdot h_{8} \cdot h_{6} \cdot h_{6} \cdot h_{6} \cdot h_{8} \cdot h_{10} \cdot h_{4} \cdot h_{0} \cdot h_{6} \cdot h_{8} \cdot h_{2} \cdot h_{8} \\ &
	h_{6} \cdot h_{6} \cdot h_{4} \cdot h_{10} \cdot h_{10} \cdot h_{10} \cdot h_{1} \cdot h_{7} \cdot h_{5} \cdot h_{5} \cdot h_{5} \cdot h_{11} \cdot h_{9} \cdot h_{2} \cdot h_{8} \cdot h_{6} \cdot h_{4} \cdot h_{2} \\ &
	h_{8} \cdot h_{6} \cdot h_{4} \cdot h_{2} \cdot h_{8} \cdot h_{6} \cdot h_{6} \cdot h_{8} \cdot h_{10} \cdot h_{4} \cdot h_{10} \cdot h_{10} \cdot h_{2} \cdot h_{8} \cdot h_{6} \cdot h_{6} \cdot h_{2} \\ &
	h_{8} \cdot h_{6} \cdot h_{6} \cdot h_{8} \cdot h_{0} \cdot h_{6} \cdot h_{8} \cdot h_{6} \cdot h_{8} \cdot h_{10} \cdot h_{4} \cdot h_{10} \cdot h_{8} \cdot h_{6} \cdot h_{2} \cdot h_{8} \cdot h_{6} \\ &
	h_{6} \cdot h_{8} \cdot h_{0} \cdot h_{6} \cdot h_{8} \cdot h_{6} \cdot h_{8} \cdot h_{10} \cdot h_{2} \cdot h_{8} \cdot h_{6} \cdot h_{6} \cdot h_{8} \cdot h_{0} \cdot h_{6} \cdot h_{8} \cdot h_{6} \\ &
	h_{8} \cdot h_{8} \cdot h_{6} \cdot h_{8} \cdot h_{0} \cdot h_{6} \cdot h_{8} \cdot h_{6} \cdot h_{8} \cdot h_{8} \cdot h_{0} \cdot h_{6} \cdot h_{8} \cdot h_{6} \cdot h_{8} \cdot h_{10} \cdot h_{0} \\ &
	h_{6} \cdot h_{8} \cdot h_{6} \cdot h_{8} \cdot h_{8} \cdot h_{1} \cdot h_{7} \cdot h_{5} \cdot h_{11} \cdot h_{9} \cdot h_{3} \cdot h_{11} \cdot h_{9} \cdot h_{11} \cdot h_{8} \cdot h_{0} \cdot h_{6} \cdot h_{3} \cdot h_{11}.
\end{aligned}
\]

This sequence should be read from left to right; first apply $h_2$, then $h_8$, and so forth. The length 381 path is presented in \cref{app:paths}.
\footnote{An interested reader may verify the correctness of these paths through a Jupyter Notebook available at the following link for the associated GitHub repository,
	\begin{center}
		\href{https://github.com/shehper/AC-Solver}{https://github.com/shehper/AC-Solver}.
	\end{center}
}
	% !TEX root = ../ac_paper.tex

\begin{figure}
	\centering
	\includegraphics[scale=0.35]{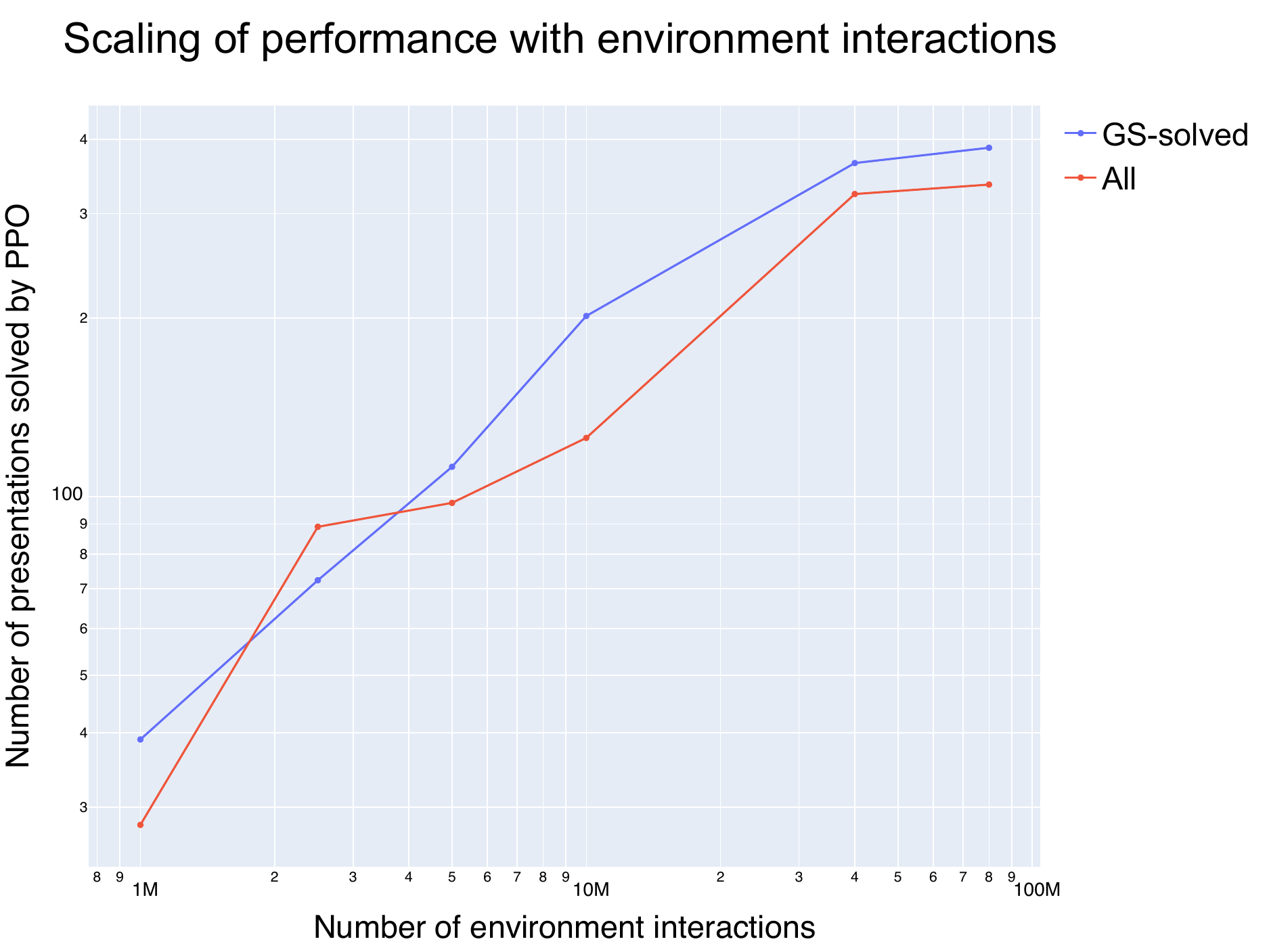}
	\caption{Number of AC presentations solved by our PPO agent as a function of the number of training steps. Here, \textit{GS-solved} refers to a subset of the Miller--Schupp dataset of \cref{sec:search} that was solved by the greedy search algorithm.}
	\label{fig:scaling_env}
\end{figure}

\section{The cure: new algorithms}\label{sec:algo}

In previous sections we explained from a variety of different perspectives that the Andrews--Curtis conjecture is a good example of a mathematical problem where the length of a solution can be much greater than the length of the initial presentation, in some cases with purely analytical lower bounds that are superexponential in the size of the input. In particular, we saw that small increases in presentation length under 20 quickly lead to solution lengths in the range of hundreds and higher, quickly exceeding the number of moves in the longest game of chess.

If solving a mathematical problem required finding a path of length $L$, say with $L=10^6$, an RL agent would be pretty much out of luck under circumstances of a typical hard search problem, where the number of successful paths is exponentially suppressed by $L$. The good news is that in mathematics --- and in many other domains --- such hard search problems never come in isolation. Rather, there is a distribution of problems such that generic instances are ``easy'' and a small fraction is ``hard.'' Learning this distribution for smaller values of $L$ contains crucial information that one can use to solve new cases with the larger value of $L$.

\subsection{Supermoves}

In automated reasoning or search problems where the minimal length solution has a theoretical lower bound that by far exceeds computational capabilities, it is clear that direct approach with fixed size steps is not going to succeed, unless the problem is easy and a large fraction of long paths meets the desired criteria. In order to reach extraordinary path lengths, one must allow progressively longer sequences of elementary moves to be added to the action space.\footnote{In the AI approaches to theorem proving, this can be compared to constructing proofs of Lemmas and subgoals, which then can be used as building blocks in larger proofs \cite{yang2019coqgym,bansal2019holist,polu2020generative,polu2022formal}.} Although this general strategy seems unavoidable in problems like the AC conjecture, it leads to many practical questions. For example, what should be the selection criteria for such ``supermoves''? And, how often should they be added to the action space?

In the context of the AC conjecture, a good example of such supermoves are the ``elementary M-transformations'' \cite{BurnsI, BurnsII}. These transformations trivialize $\AK(2)$ in just two steps, even though this presentation is known to admit the shortest AC trivialization path of length 14. A downside of elementary M-transformations, though, is that they are infinite in number, which complicates their application in classical search techniques; rather, we need a finite action space with a relatively small number of supermoves.

In our study, we explored the simplest version of identifying AC supermoves by selecting some frequently occurring subsequences of AC-moves in the paths discovered by Proximal Policy Optimization (PPO). By extending the action space $A$ of the Markov Decision Process (MDP) with these subsequences and checking whether this enhanced action space helps our agent discover shorter paths of trivialization, we learned a few useful lessons:

\begin{itemize}

	\item First, it helps to augment the action space with subsequences of different kind that include frequently occurring compositions of elementary moves as well as very rare ones.

	\item Also, in the early stage it helps to introduce several supermoves at once.

	\item And, at later stages it helps to allow removing actions from the action space, not only adding them.
\end{itemize}

\noindent
Not following these empirical rules, e.g. introducing too few supermoves initially or too many over the entire length of the training process, leads to considerable reduction in performance of the RL agent. Even in the most optimal regimes that we were able to find, the improvement of the performance due to supermoves was rather modest, leading us to explore other alternatives.

\begin{figure}[ht]
	\centering
	\includegraphics[trim={0.5in 1.1in 0.5in 1.0in},clip,width=3.5in]{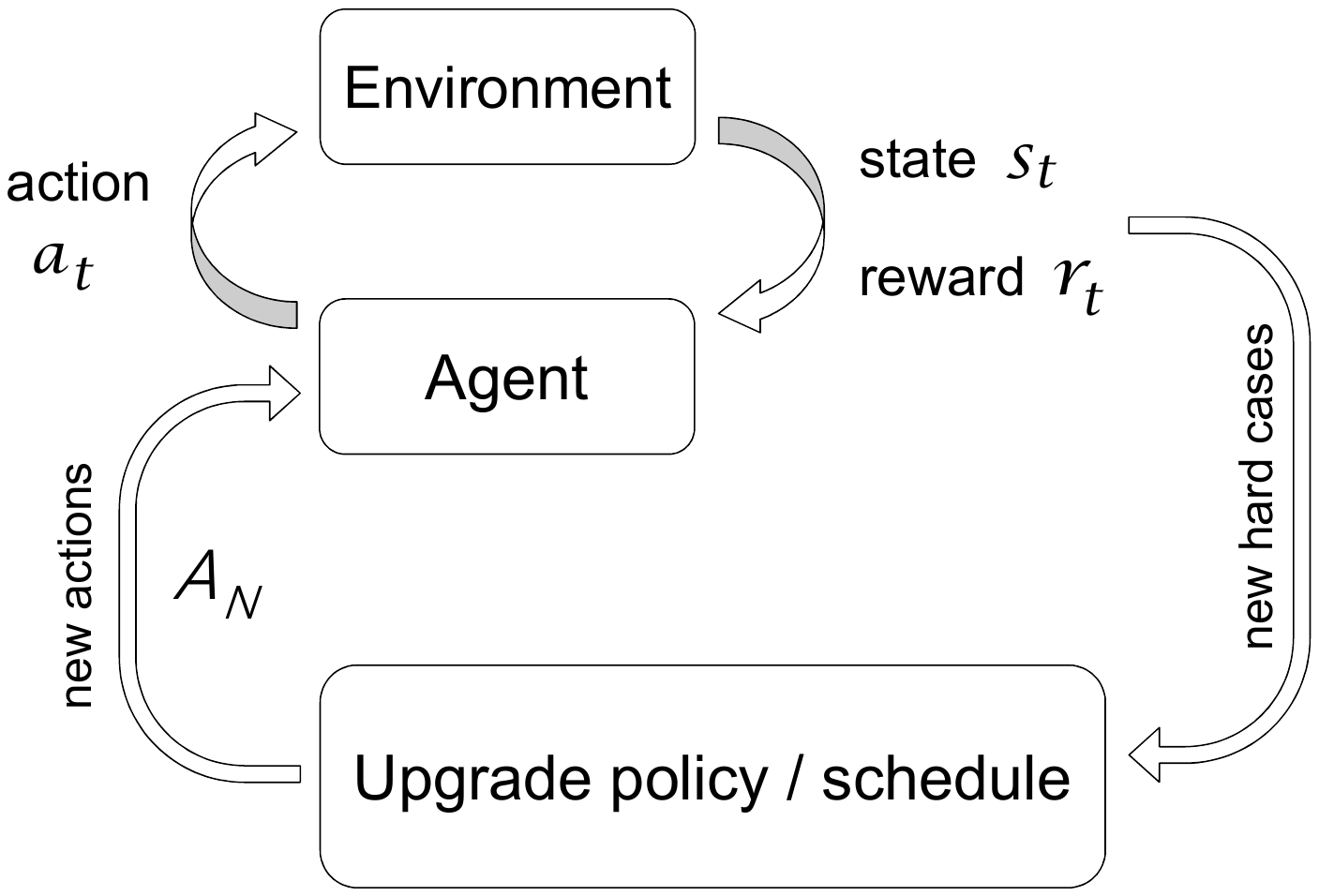}
	\caption{A schematic representation of a model with an adaptive action space.}
	\label{fig:dynamicactions}
\end{figure}

To summarize, the dynamic way of introducing supermoves is crucial for hard search problems like the AC conjecture, and the algorithm (or, policy) for making such adaptive changes to the action space is, in turn, crucial to the performance. Schematically, such adaptive changes to the action space can be illustrated by a diagram in \cref{fig:dynamicactions}, which replaces the traditional RL cycle illustrated in \cref{fig:RLbasic}. In the rest of this section we discuss several ideas on the implementation of adaptive action spaces motivated by the environment of the AC conjecture. They all are based on the notion of ``hardness'' which plays an important role in the following.

\begin{figure}[ht]
	\centering
	\includegraphics[trim={0.0in 2.0in 0.0in 1.5in},clip,width=4.0in]{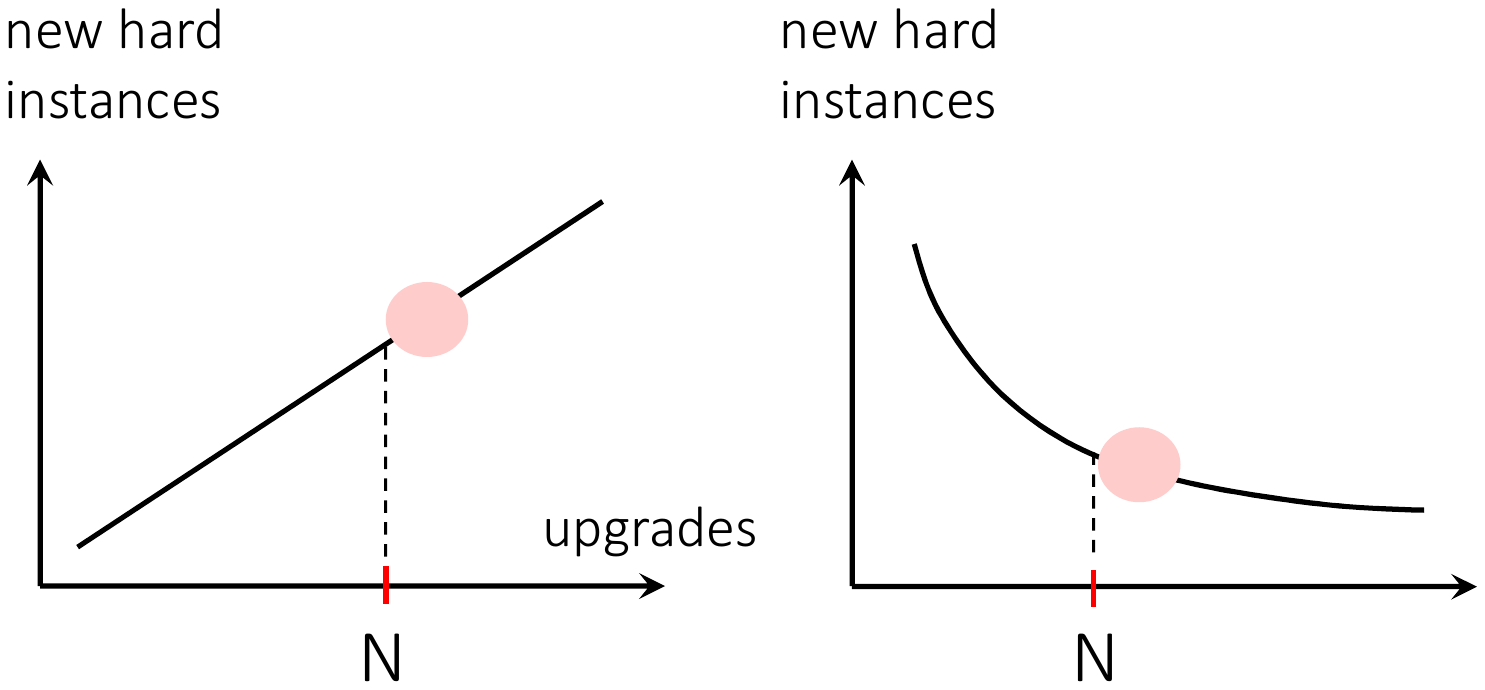}
	\caption{Depending on the problem, the distribution of new hard instances that are found at each iteration of the model upgrades can vary. This distribution, in turn, determines the update schedule of the action space. The action space of the upgrade $N+1$ is determined by the new hard instances found at the stage $N$ and earlier.}
	\label{fig:hardnesscurves}
\end{figure}

\subsection{Adaptive action space and the anatomy of success}

While supermoves clearly need to be a part of the solution in hard problems like the AC conjecture, much of the success depends on the criteria for selecting them. Here, we advocate for an adaptive approach where the model itself learns the criteria for selecting supermoves, in addition to the best ways to implement them. One realization of this approach could be a multi-agent model, where one network is learning to play the game while the other is learning the rules for changing the action space (adding and removing supermoves), cf. \cref{fig:dynamicactions}. We hope that future iterations of this strategy can lead to AI systems that can `learn how to learn' dynamically by making both algorithmic and architectural changes through collecting the information about hard instances.\footnote{Here, by self-improving AI systems we mean algorithms that have the ability to ``interpolate'' between off-the-shelf algorithms such as A2C and TRPO, as well as a myriad of custom algorithms that do not even have a name. Clearly, this level of technology is not presently available (see, however, \cite{lu2022discovered} for an interesting work in this direction), and one of the key points of this section is that developing such systems should be based on the hardest instances the agent encounters.} In other words, we propose AI system to select supermoves based on the notion of ``hardness'' that also can be learned.

\begin{figure}
	\centering
	\includegraphics[scale=0.35]{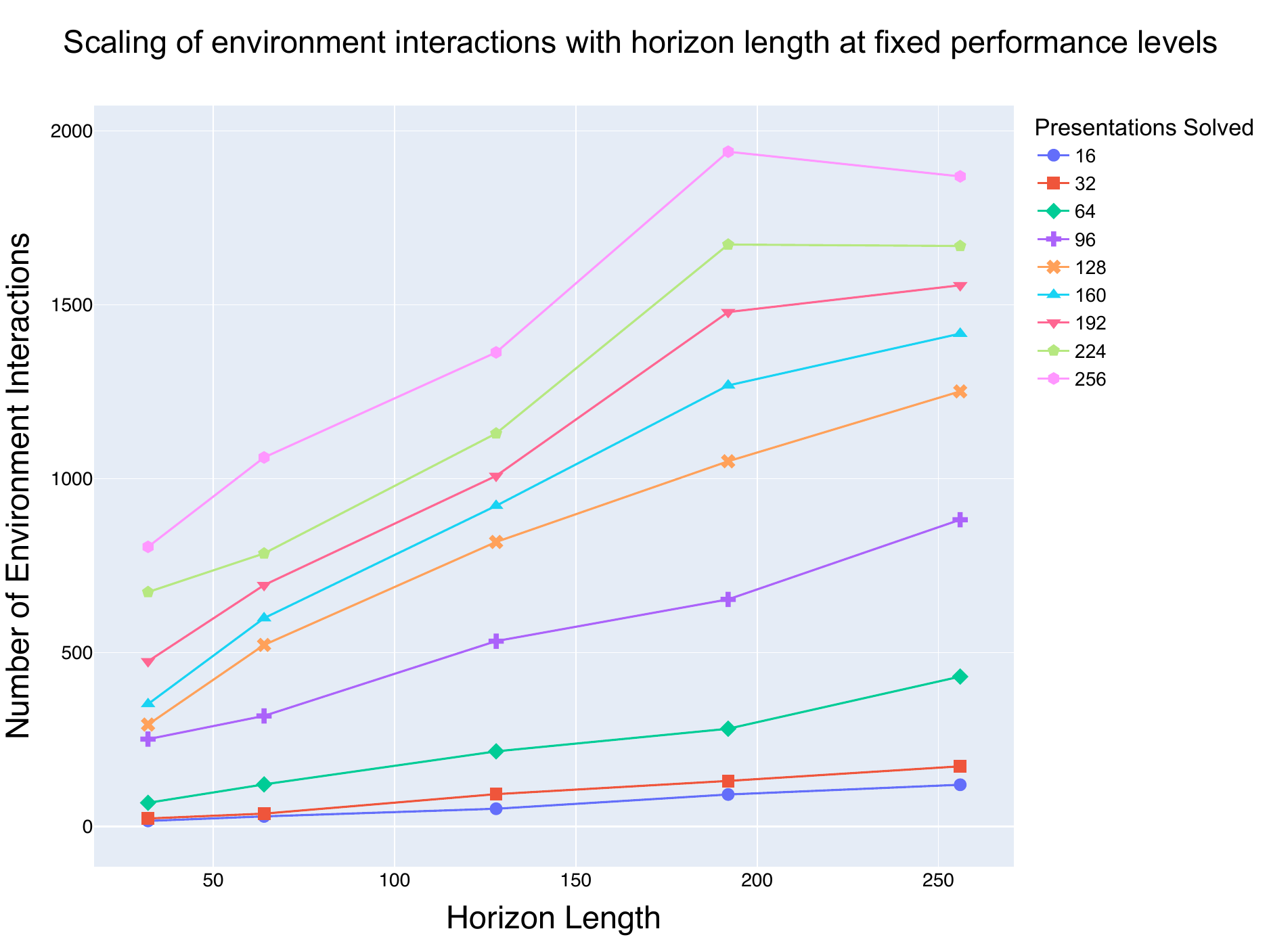}
	\caption{To maintain consistent performance, increasing the horizon length requires a roughly linear increase in the number of training steps (environment interactions).}
	\label{fig:env_vs_horizon}
\end{figure}

For concreteness, suppose $N$ is one of the characteristics of either the algorithm or the architecture that has non-trivial impact on performance. In practice, there can be several such parameters, but for simplicity we explain the idea as if there is only one.\footnote{Analyzing a multi-dimensional landscape is generally a good idea, as it can yield better performance improvements, but the price to pay is that changes of different characteristics become correlated, as illustrated in \cref{fig:env_vs_horizon}.} Thus, in studying scaling laws one typically looks at performance as a function of $a)$ the number of model parameters (depth and width), $b)$ training time, and $c)$ the size of the data. We can pick $N$ to be any of these, and in the context of the AC conjecture the number of environment interactions (training time) is a natural candidate that we consider below when we need a concrete example. As \cref{fig:env_vs_horizon} illustrates, if $N$ represents incremental upgrades in the number of environment interactions, the model of \cref{sec:rl} is able to expand to larger horizon lengths.

Now, we can define \textit{hard instances} as those which the model is able solve at the current setting of $N$ and not with the lower value of the resource $N$. In addition, in search problems we include the length of the path in the notion of hardness, i.e. select a subset of the instances that the model could solve through especially long paths. Note, by the very nature of the search problem we are interested in, there can not be too many such hard instances at each step of increasing $N$, for otherwise the problem would be easy, not hard. Collecting the information about the hardest instances at each increment in $N$ can be used to select supermoves, e.g. one can choose the latter to be subsequences of the sequences of moves that solve the hard instances. \cref{alg:adaptive_ai_model} provides one particular realization of this idea.

\begin{algorithm}
	\caption{Adaptive AI Model Training and Path Discovery}
	\label{alg:adaptive_ai_model}
	\begin{algorithmic}[1]
		\State \textbf{Input:}
		\begin{itemize}
			\item[] Family of AI models $\pi(N)$ with common state space $S$ and action space $A_0$
			\item[] Initial setting $N_0$ and ordered range $\{N_1, N_2, \ldots, N_{\text{max}}\}$
			\item[] Number of epochs for training
			\item[] Validation set $V \subset S$
			\item[] Distinguished state $s_0 \in S$
			\item[] Positive integer $n$
		\end{itemize}

		\State \textbf{Output:}
		\begin{itemize}
			\item[] For each setting $N_i$: Set of pairs $\{v, P\}$ where $v \in V$ and $P$ connects $v$ to $s_0$
		\end{itemize}

		\State Initialize $A(N_1) \gets A_0$

		\For{each $N_i$ in $\{N_1, N_2, \ldots, N_{\text{max}}\}$}
		\State Train model $\pi(N_i)$ on $S$ for the given number of epochs
		\State Evaluate $\pi(N_i)$ on $V$ to discover paths connecting $V$ to $s_0$ using $A(N_i)$

		\State $V(N_i) \gets \{ v \in V \mid v$ can be connected to $s_0$ using $A(N_i)$, but not by any $\pi(N_j)$ with $j < i\}$
		\State $W(N_i) \gets \{ v \in V(N_i) \mid$ the longest path connecting $v$ to $s_0$ using $A_0 \}$

		\If{$i \geq n$}
		\State Compare $W(N_{i-n+1})$ to $W(N_i)$
		\State Adjust $A(N_{i+1})$ based on the comparison
		\Else
		\State $A(N_{i+1}) \gets A(N_i)$
		\EndIf
		\EndFor
	\end{algorithmic}
\end{algorithm}

The reason to include the length of the path in the definition of hardness is well illustrated by the AC conjecture. In this case, increasing the number of environment interactions, $N$, leads to a larger number of non-trivial presentations from the Miller--Schupp series being solved (i.e. AC-trivialized) by our RL agent, see \cref{fig:scaling_env}. Moreover, the length of the AC trivialization path also grows for some of the solutions (but not all). Therefore, in order to implement the program outlined above in the specific context of the AC conjecture, we can focus on the longest AC trivialization paths that the model is able to find at each value of $N$.

\begin{figure}[h]
	\centering
	\begin{subfigure}{0.45\textwidth}
		\centering
		\includegraphics[width=\textwidth]{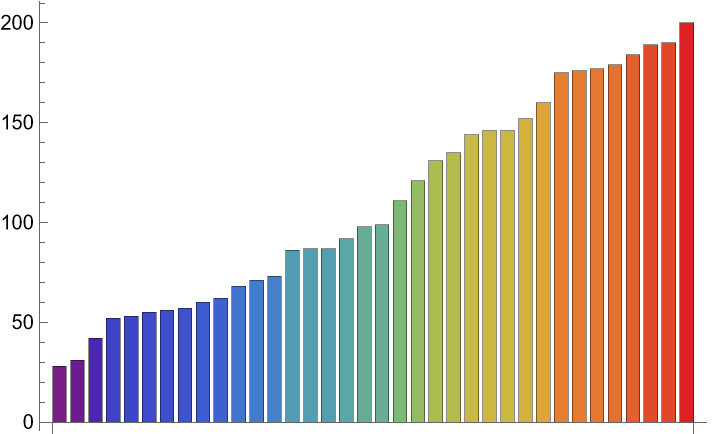}
		\caption{all}
		\label{fig:all_path_length_80M}
	\end{subfigure}
	\hfill
	\begin{subfigure}{0.45\textwidth}
		\centering
		\includegraphics[width=\textwidth]{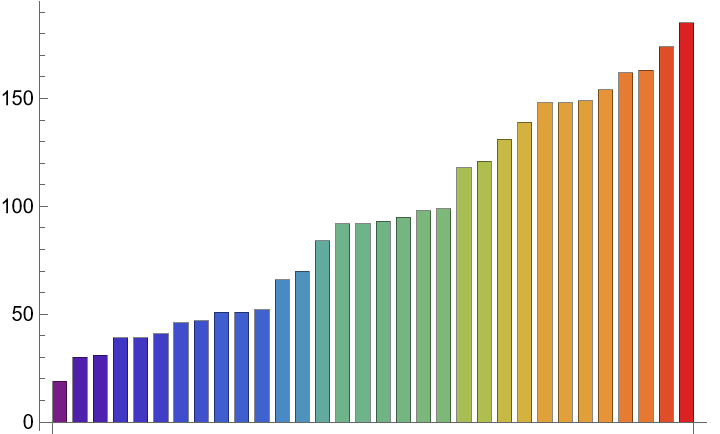}
		\caption{GS-solved}
		\label{fig:solved_path_length_80M}
	\end{subfigure}
	\caption{Path length distributions for AC presentations solved by the RL agent at $N=8 \times 10^7$ from \textit{all} and \textit{GS-solved} datasets are nearly identical. In both cases, hard instances are shown in red.}
	\label{fig:path_length_80M}
\end{figure}

Collecting this information in the process of training can be used to introduce (and remove) supermoves to the action space in an adaptive fashion. There can be many different implementations of this idea that we plan to explore more fully elsewhere. For example, one can consider selecting all or subset of the longest AC trivialization paths that the model finds at each $N$. Out of those, in each case, one can consider selecting the entire trivialization path or a subset, randomly or non-randomly. Alternatively, one can compare the longest trivialization paths at several (consecutive) values of $N$ and choose subsequences of moves that are shared by several long trivialization paths at different $N$.

Below we briefly examine the structure of the longest (i.e. hard) AC trivializations found by our RL agent with several different seed values. First, let us illustrate the effects of stochasticity and consider $N=8 \times 10^7$. The agents with five different seed values were able to solve 354, 337, 330, 328, and 323 presentations, respectively. And their average, $334.4$, is shown in \cref{fig:scaling_env}. Many of these AC presentations can be solved at the earlier stage, with $N=4 \times 10^7$ or less. If in the definition of \textit{hard instances} we require that they are solved by \textit{all} five agents, there are only 5 presentations total. On the other hand, if we require that they are solved by \textit{any} of the 5 agents, the number goes up to 36.

\begin{figure}[h]
	\centering
	\includegraphics[scale=0.6]{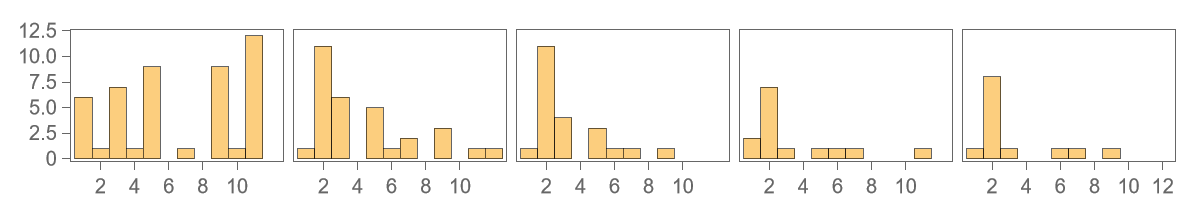}
	\caption{Types of AC-moves that appear in trivialization paths of 5 presentations solved by all 5 agents at $N=8 \times 10^7$. The move $\# 2$ occurs disproportionately more frequently. There are 12 different types of basic AC-moves described in \cref{sec:search}.}
	\label{fig:anatomy_all}
\end{figure}

\begin{figure}[h]
	\centering
	\includegraphics[scale=0.6]{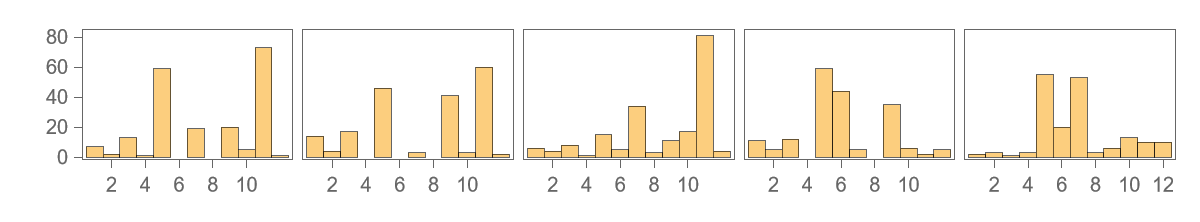}
	\caption{Types of AC-moves in the 5 longest trivialization paths (of length 200, 190, 189, 184, and 179) found by at least one agent at $N=8 \times 10^7$. The most frequent moves are $\# 5$, $\# 7$, $\# 9$, and $\# 11$. There are 12 different types of basic AC-moves described in \cref{sec:search}.}
	\label{fig:anatomy_some}
\end{figure}

Moreover, not surprisingly, the 5 presentations solved by all 5 agents have considerably shorter path lengths (47, 31, 22, 14, and 13) compared to path lengths of the 36 presentations illustrated on the left panel of \cref{fig:path_length_80M} that go up to $200$. Both 5 presentations solved by all agents and 36 presentations solved by at least one of the agents provide viable options for defining hard instances and, in turn, selecting supermoves. However, they lead to qualitatively different results. For example, all 5 presentations solved by all 5 agents are solved at a smaller value of $N$ when required to be solved by only one of the agents. More importantly, they have very different anatomy, illustrated in \cref{fig:anatomy_all} and in \cref{fig:anatomy_some}, respectively.
By examining the longest trivialization paths of the 36 presentations solved by at least one agent at $N=8 \times 10^7$, we often see long strings of moves $\# 5$ and $\# 11$, interlaced with moves $\# 3$, $\# 7$, and $\# 9$. These are our top candidates for the supermoves to be added at $N=8 \times 10^7$, if $N$ was allowed to gradually increase in an adaptive system, as in \cref{fig:dynamicactions}. Note that moves $\# 4$ and $\# 8$ are least common in the examples presented in both \cref{fig:anatomy_all} and \cref{fig:anatomy_some}.

\begin{figure}[h]
	\centering
	\includegraphics[scale=0.5]{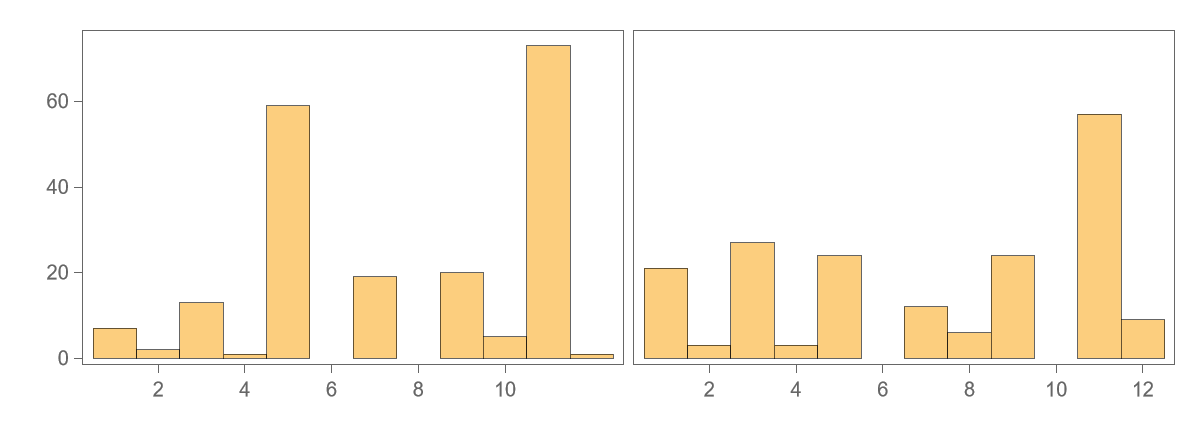}
	\caption{An example illustrating anatomy of successful AC paths found by an RL agent in different runs working on different datasets, both of which contain the same hard instance. Even though the total number of moves can vary, e.g. in this example we had to rescale one of the distributions by a factor of 3, curiously, the proportion of moves and combinations used are quite similar. In both cases, RL agents used the same opening move.}
	\label{fig:anatomy200}
\end{figure}

As in other parts of this paper, we performed the analysis on two datasets of sizes 1190 and 533 that, respectively, contain all members of the Miller--Schupp family with $n \leq 7 \, \land \, \length(w) \leq 7$ and only those solved by the greedy search. The results are qualitatively similar, as we already saw in \cref{fig:path_length_80M} that illustrates length distributions of the successful AC paths in the two cases. Similarly, a closer look at the anatomy of the successful paths ---successful for the RL agent--- reveals no qualitative differences between the two datasets and, importantly, consistency of our notion of \textit{hardness} based on the path length. The largest level of stochasticity that one may expect perhaps can be illustrated by an example of the presentation
\[
\angles{x, y \mid x^{-1} y^2 x y^{-3} = 1 \,, \; x^{-2} y^{-1} x y^{-4} = 1}
\]
that an RL agent was able to solve at $N = 4 \times 10^7$ with 62 moves in one case and at $N = 8 \times 10^7$ with 200 moves in the other case. Despite considerable variance, in both cases successful AC paths are dominated by the move $\# 11$, interlaced with moves $\# 3$, $\# 5$, $\# 7$, and $\# 9$ according to patterns described above (cf. \cref{fig:anatomy200}). This can serve as evidence for robustness of the supermove selection process proposed here, and provides further support to \cref{alg:adaptive_ai_model} and its variations.

In the following sections we explore the anatomy of successful AC paths further using the tools of unsupervised learning and topological data analysis. These tools offer a complementary perspective on hard-to-find AC paths and ways to find them.

	% !TEX root = ../ac_paper.tex

\section{Language modeling} \label{sec:lm}

One of the main messages in this paper is that dynamically learning the notion of hardness is of paramount importance to the success of RL agents in problems with extremely long horizons and sparse rewards. One way of approaching this goal is to build a larger AI system in which RL model is accompanied by another component that is good at recognizing hard vs easy instances, AC presentations in our case study. Therefore, as a prerequisite, it is natural to explore data analysis tools that can do such identification. In this section, we explore how well Language Models can do this job, and in the next section we explore alternative data analysis tools based on the structure of the AC graph, whose vertices represent balanced presentations and edges represent basic AC moves.

The approach via language modeling appears very natural because balanced presentations are described by ``words'' in group generators, and this is something that language models learn remarkably well.
Specifically, each presentation with two relators is a sequence made of six letters, also known as ``tokens" in the nomenclature of Natural Language Processing, i.e. $x$, $y$, $x^{-1}$, and $y^{-1}$, and two ``stop tokens": one that separates two relators of a presentation and another that marks the end of a presentation.
Given this vocabulary $V$ of six tokens, we can ask what is the probability $p(t_1, \dots, t_N)$ for $t_i \in V$ of the occurrence of a specific presentation in the space of all balanced presentations.
Using the chain rule of probability theory,
\[
p(t_1 \cdots t_{N}) = \prod \limits_{i=1}^{N} p (t_{i} \mid t_{1} \cdots t_{i-1})
\]
Here $p (t_{N} \mid t_{1} \cdots t_{N-1})$, often called the $N$-gram probability distribution, is the probability of a token $t_N$ following a sequence of tokens $t_{1} \cdots t_{N-1}$.
To model the language of balanced presentations, we can alternatively estimate the $N$-gram probability distributions for all $N$.

Over the last few years, Transformer models have shown great success in modeling human-understandable languages to the extent that these models can create text almost indistinguishable from that of a human expert.
Specifically, the architecture used for modeling language is the auto-regressive ``decoder-only" Transformer, which we review in detail in \cref{sec:transformer_review}.
In \cref{sec:transformer_datasets}, we discuss the method with which we generate the dataset required for training the model.
Finally, in \cref{sec:transformer_results}, we share details of some insights we learned from this process.

\subsection{Transformers: a review\label{sec:transformer_review}}

Here, we give a short review of the architecture of a decoder-only transformer.
For more details, see \cite{vaswani2023attention, elhage2021mathematical, douglas2023large}.

Given an input sequence $t_1, t_2, \dots, t_{N}$, a decoder-only transformer predicts the probability distribution $p(t \mid t_1, t_2, \dots, t_{N})$ over the set $V$ of tokens of size $n_{\text{vocab}}$.
The probability is computed by applying the softmax function to the logits $T(t)$, which are estimated by applying the following sequence of operations.\footnote{
The softmax function, $\softmax \colon \R^n \to (0, 1)^{n}$, is defined as $\softmax(x)_i = e^{x_i} / \sum\limits_{j=1}^n e^{x_j}$.}
First, assign to each token in the vocabulary a distinct label in the range $1, 2, \dots, n_{\text{vocab}}$; re-writing the original sequence as a sequence of integers.
We will label these integers also as $t_i$.
Next, write the sequence in terms of ``one-hot encoded vectors", i.e. a matrix $t \in \R^{N \times n_{\text{vocab}}}$ such that
\[
t_{ij} = \delta_{i t_i}
\]
and embed the sequence in a $\dm$-dimensional vector space,\footnote{
Here, $t$ and all $x_j$ are two-dimensional tensors.
Hence, it is appropriate to apply tensors of linear transformations to them.
Often in a transformer architecture, these operations are of the form $\mathbb{1} \otimes \cdots$; in these cases, we drop the identity transformation and simply write the operation as $\cdots$.
For example, $\mathbb{1} \otimes W_U$, $\mathbb{1} \otimes W^m_I$, $\mathbb{1} \otimes W^m_O$, etc.
In this case, we will sometimes write $W_U$, $W^m_I$, $W^m_O$ respectively, assuming it is clear from the context and the dimensionality of these matrices that they are tensored with identity transformations.}
\[
x_0 = (W_P \otimes \mathbb{1} + \mathbb{1} \otimes W_E) t
\]
Here, $W_P \in \R^{\dm \times N}$ and $W_E \in \R^{\dm \times n_{\text{vocab}} }$ are matrices of learnable parameters, known as the ``positional embedding" and ``token embedding" matrices.

An $L$-layer transformer alternates between applying a ``multi-head attention layer" ($\sum\limits_{h \in H} h$) and an ``MLP-layer" ($m$) $L$ times.
For $i=0, \dots, L-1$,
\[
\begin{aligned}
	x_{2i+1} &= x_{2i} + \sum_{h \in H} h(\LN(x_{2i})), \\
	x_{2i + 2} &= x_{2i + 1} + m(\LN(x_{2i + 1})).
\end{aligned}
\]
Each $x_j$ is an element of $\mathbb{R^{N \times \dm}}$, with the interpretation that its $i$-th row is the embedding of the sequence $t_1, \dots, t_i$ in the embedding space $\R^{\dm}$ as learned by the preceeding $j+1$ operations.
Finally, one applies an ``unembedding layer", $W_U \in \R^{n_{\text{vocab}} \times \dm}$, to convert the output of the final layer to an $n_{\text{vocab}}$-dimensional vector of logits that estimate the sought-after probability distribution.
\[
\begin{aligned}
	T(t) &= W_U x_{2L-1}, \\
	p(t) &=\softmax (T(t)).
\end{aligned}
\]

The functions $\LN$, $m$ and $h$ are defined as follows.
$\LN$ is the LayerNorm operation that normalizes the input of each layer to make the optimization process more stable~(\cite{ba2016layer}):
\[
\LN(x) = \left(\mathbb{1} \otimes \text{diag}(\gamma) \right) \frac{(x-\overline{x})}{\sqrt{\text{var}(x)}} + \mathbb{1} \otimes \beta\,.
\]
Here, $\overline{x}$ and $\text{var}(x)$ are mean and variance of each row of $x$, and $\gamma, \beta \in \R^{\dm}$ are learnable parameters.
The MLP-layer $m$ is a non-linear operation,
\[
m(x) =W ^m_O \ \max(W_I^m x, 0)
\]
with learnable parameters $W^m_I \in \R^{d_{\text{MLP}} \times \dm}$, $W^m_O \in \R^{\dm \times d_{\text{MLP}}}$.
It is standard to set $d_{\text{MLP}} = 4 \dm$.

Finally, the multi-headed attention-layer $\sum_{h \in H} h$ is a sum of $n_{\text{heads}}$ ``attention-head" operations $h$, where
\[
h(x) = (A^h(x) \otimes W^h_O W^h_V) x.
\]
Here, $W^h_V \in \R^{d_{\text{head}} \times \dm}$,
$W^h_O \in \R^{\dm \times d_{\text{head}}}$
are matrices of learnable parameters;
$d_{\text{head}}$ is the ``attention-head dimension" that satisfies $d_{\text{head}} \times n_{\text{head}} = d_{\text{model}}$; and the attention matrix $A^h$ is computed with the help of learnable matrices
$W^h_Q, W^h_K \in \R^{d_{\text{head}} \times \dm}$,
\[
A^h(x) = \softmax^\star \left(\frac{x^T (W^h_Q)^T W^h_K x}{\sqrt{d_{\text{head}}}}\right).
\]
The attention-head is an $N \times N$ matrix, with the interpretation that $A^h(x)_{ij}$
is the ``attention" paid to the token
$t_j$ in estimating
$p(t_{i+1} \mid t_1, \dots, t_i)$.
$\softmax^\star$ is a variant of the $\softmax$ function suitable for auto-regressive tasks: it sets the upper triangular part of its input to zeros before applying the
$\softmax$ operation.
That is, future tokens, $t_k$ for $k > i$, play no role in the prediction of $p(t_{i+1} \mid t_1, \dots, t_i)$.

We train the transformer model by minimizing the cross-entropy loss between the distributions of predicted and correct labels for the next tokens in a sequence.
The parallelism offered by the processing of all tokens in a sequence at once is extremely beneficial for efficient training of the model for the language modeling task.

In practice, the embedding matrix $W_E$ and the unembedding matrix $W_U$ are often ``tied" together, i.e. $W_E = W_U^T$ \cite{press2017using, inan2017tying}.
The rows of $W_E = W_U^T$ are interpreted as the embeddings of words/sentences, to which one may apply the usual operations of a vector space \cite{Bengio:2003, mikolov-etal-2013-linguistic}.
For example, the cosine of the angle between two embedding vectors, also known as the ``cosine similarity", is often used to measure the similarity between two texts.
Two semantically similar texts have higher cosine similarity between them, while semantically different texts correspond to (almost) orthogonal vectors in the embedding space.

\subsection{Training and evaluation datasets\label{sec:transformer_datasets}}

We now discuss the training and validation datasets used to train and evaluate our Transformer model.
As our main interest in this paper has been in the presentations of the Miller--Schupp series, we generated a dataset of balanced presentations that are AC-equivalent to the Miller--Schupp presentations.
Specifically, we apply sequences of AC-moves to the 1190 presentations with $n \leq 7$ and $\length(w) \leq 7$ discussed in \cref{sec:AC}, creating a dataset of about 1.8 million presentations.
Approximately 1 million of these presentations are AC-equivalent to the presentations that remained unsolved by greedy search (c.f. \cref{sec:search}).
Only a small amount (roughly 15 percent) of the original Miller--Schupp presentations were part of this dataset.

\begin{figure}
	\centering
	\includegraphics[scale=0.15]{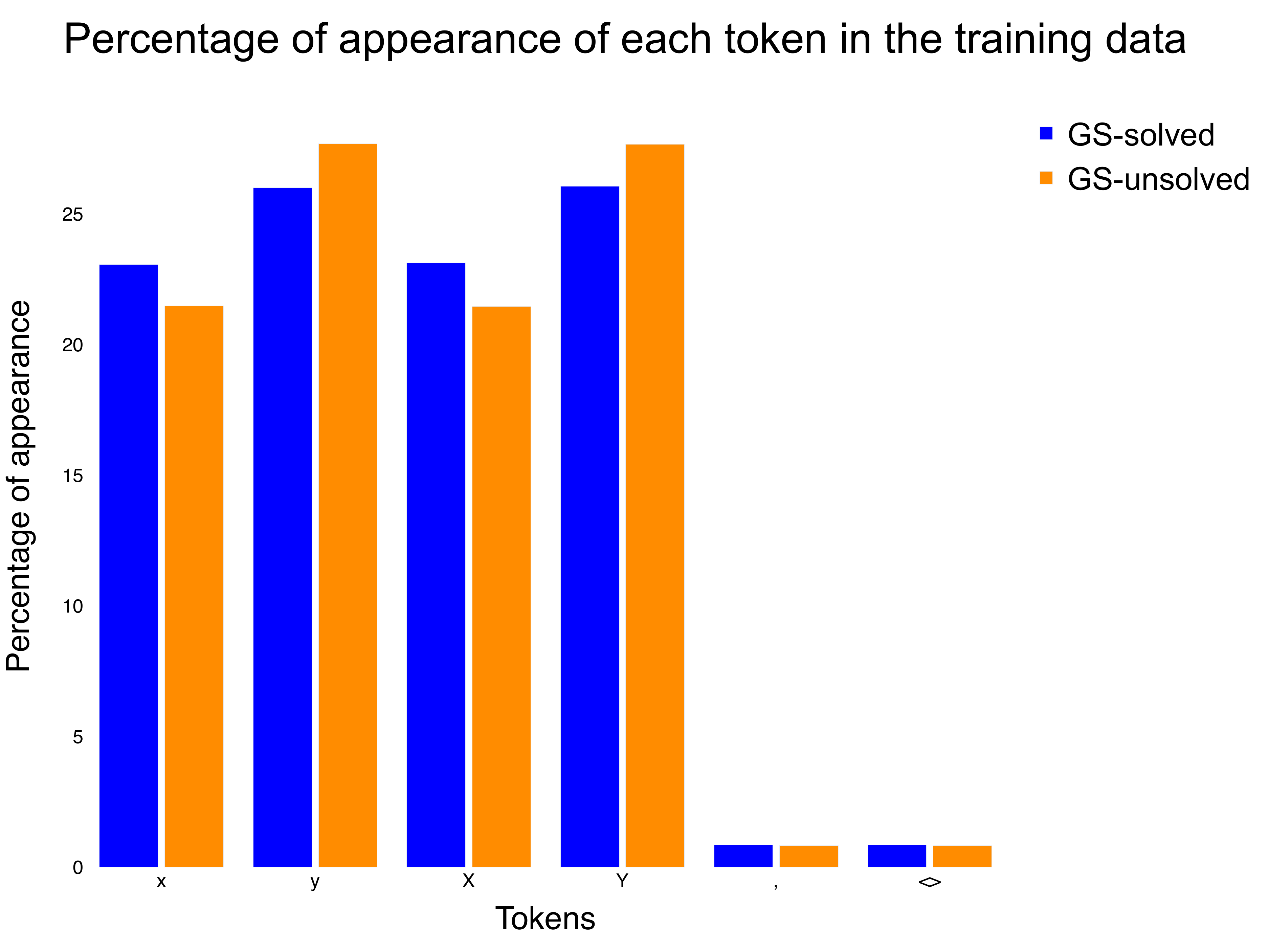}
	\caption{Percentage of appearance of each token in the two subsets of the training dataset that are equivalent to GS-solved and GS-unsolved presentations.
	To be clear, we computed the percentages separately for each subset of the training data, i.e. the heights of all blue (and orange) bars adds separately to 100.}
	\label{fig:tokens_hist}
\end{figure}

The dataset is tokenized using six tokens: two stop tokens and one token each for the two generators and their inverses.
The tokenized dataset had about $2.17 \times 10^8$ tokens.
As our goal is to get insights into properties that distinguish GS-solved and GS-unsolved presentations, we performed an exploratory data analysis of the two subsets of data associated to these presentations.
We plot the percentage of appearance of each token for these subsets in \cref{fig:tokens_hist}.
The ratio of frequency of $y^{\pm 1}$ to the frequency of $x^{\pm 1}$ is higher in the GS-unsolved dataset.
This is likely because the GS-unsolved presentations have larger $n$, and larger $n$ corresponds to a higher number of occurrence of $y^{\pm 1}$ in the Miller--Schupp presentation.
Interestingly, this effect remains in the dataset even after applying thousands of AC-moves to the original presentations.

\begin{figure}
	\centering
	\includegraphics[scale=0.15]{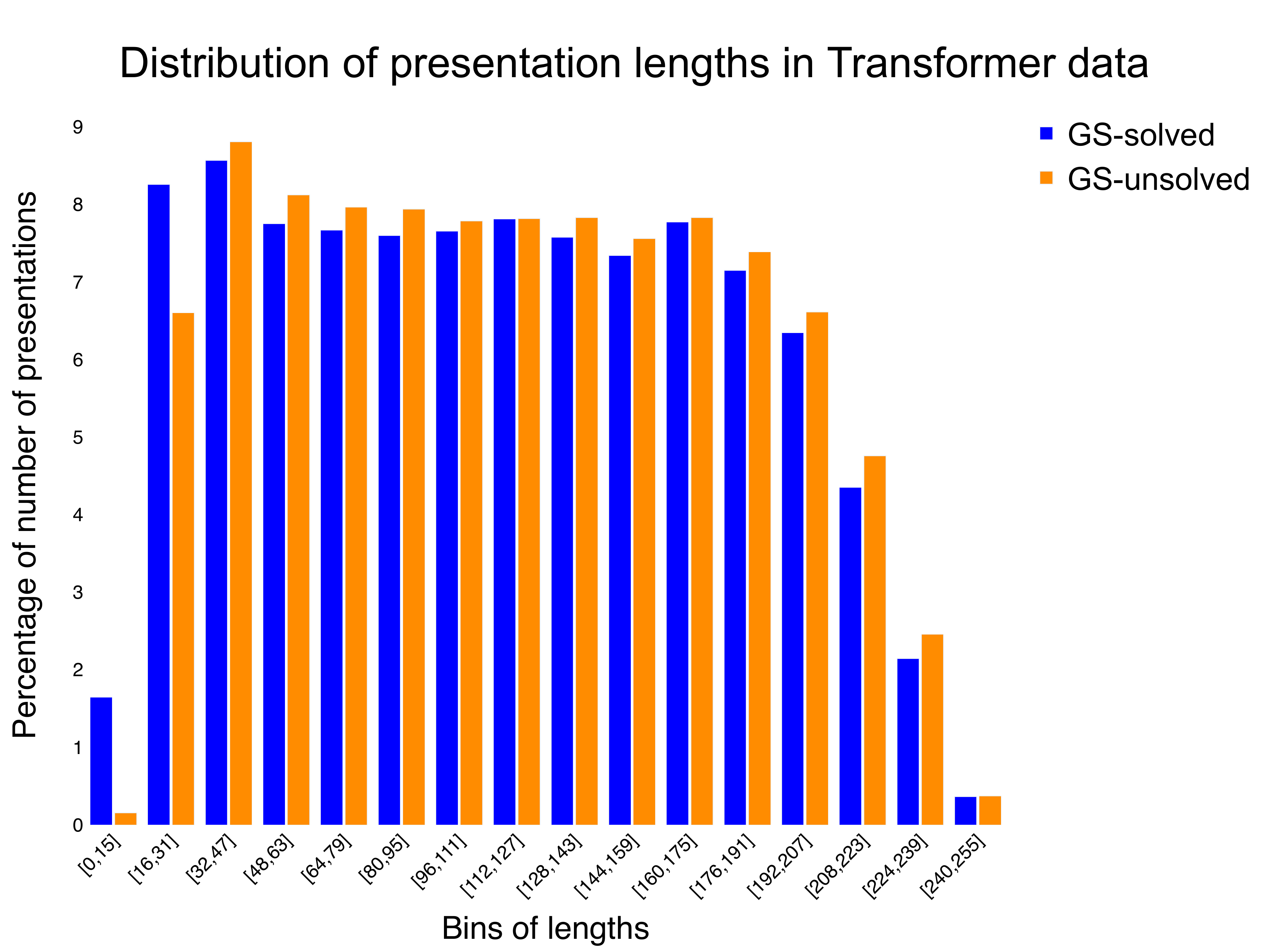}
	\caption{Percentage of presentations in various ranges of lengths.
	Percentages were computed independently for the two subsets of the dataset, corresponding to presentations that are AC-equivalent to GS-solved and GS-unsolved presentations.
	We used \cref{alg:apply_ac_moves} from \cref{app:algorithm} to ensure the almost-uniform distribution depicted here.}
	\label{fig:gpt_data}
\end{figure}

We paid special attention to ensure that our dataset contains presentations of a wide range of lengths so as not to bias our model towards learning trends specific to any fixed length.
To this end, we devised an algorithm (\cref{alg:apply_ac_moves} in \cref{app:algorithm}) that creates an almost uniform distribution over the lengths of the presentations.
(See \cref{fig:gpt_data}.) We set aside $10\%$ of our entire data for validation.

\begin{figure}
	\centering
	\includegraphics[scale=0.16]{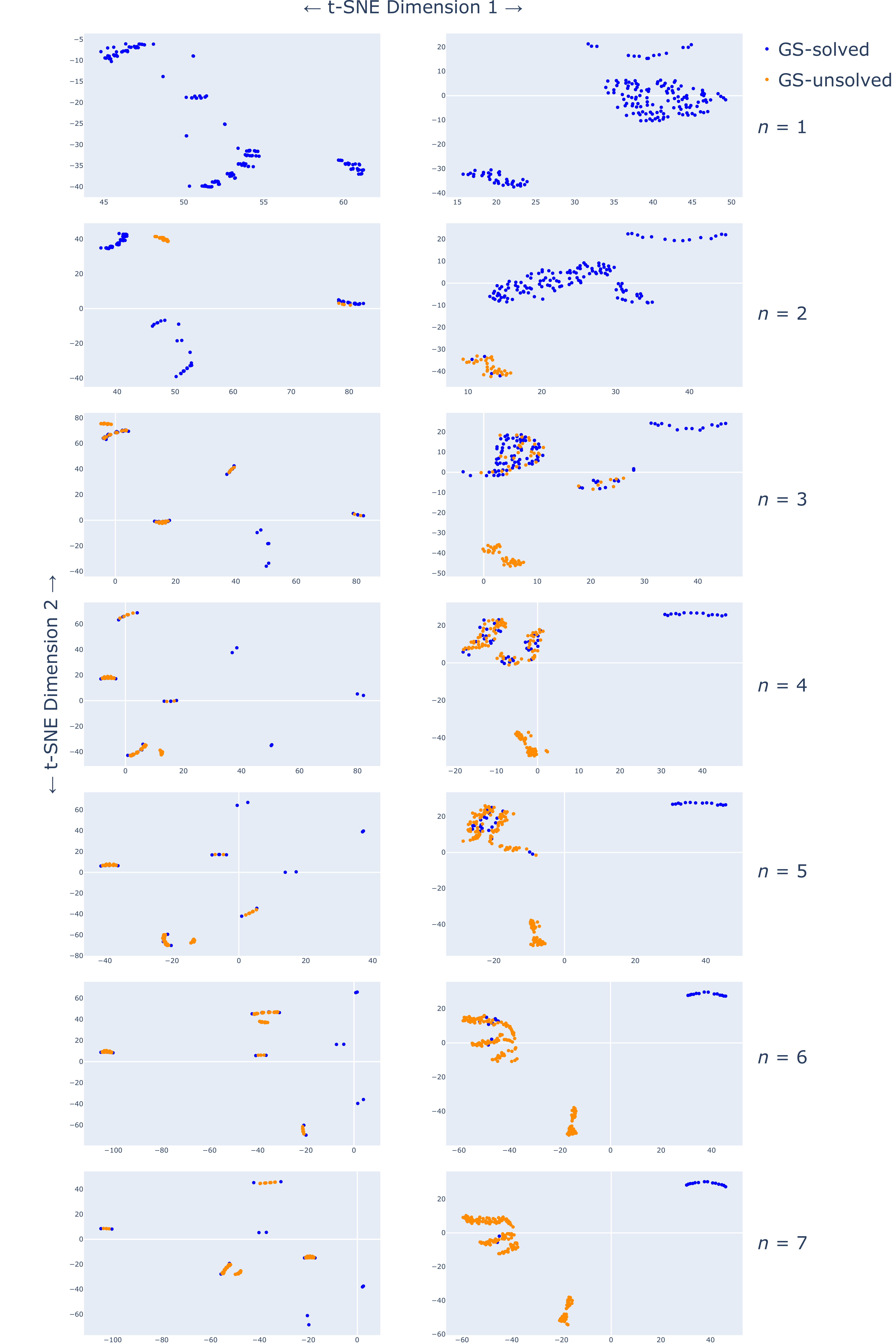}
	\captionsetup{width=1.1\textwidth}
	\caption{t-SNE plots depicting embeddings of the Miller--Schupp presentations $MS(n, w)$.
	The left (right) column shows embeddings learned by an untrained (trained) transformer model.
	Each row corresponds to a value of $n$.
	Trained models cluster together GS-solved and GS-unsolved presentations indicating the possibility of a difference in the linguistic structure of the two sets of presentations.}
	\label{fig:tsne_embeddings}
\end{figure}

\subsection{Results}\label{sec:transformer_results}

A randomly initialized model with the initialization scheme given in \cite{Radford2019LanguageMA} has a cross entropy loss of
%$- \ln \left(\frac{1}{n_{\text{vocab}}} \right) \approx - 1.7918$,
$-\ln (1/n_{\text{vocab}}) \approx 1.7918$.
With training, we could achieve a validation loss of $0.7337$.\footnote{
We tuned the hyperparameters a little but it is quite likely that one can achieve a better performing model with more hyperparameter tuning.
Similarly, more training data will necessarily help with the performance.
We trained a Transformer model with hyperparameters given in \cref{app:hyperparameters}.}
We used the untrained and the trained model to get the embeddings of all $1190$ presentations of the Miller--Schupp series with $n \leq 7$ and $\length(w) \leq 7$.
We used t-SNE to project these embedding vectors to a plane \cite{JMLR:v9:vandermaaten08a}.
The plots are shown in grid in \cref{fig:tsne_embeddings}.

Each row of \cref{fig:tsne_embeddings} corresponds to a fixed value of $n$.
The left (resp. right) column depicts t-SNE projections of embeddings obtained by an untrained (resp. trained) model.
t-SNE dependence on a distance measure: it learns to map vectors that are closer together in the higher-dimensional space, with respect to this distance measure, close together in the plane \cite{JMLR:v9:vandermaaten08a}.
We used cosine simiarity between embedding vectors as the distance measure for our plots.
We note that the GS-solved and GS-unsolved presentations seem to cluster much more in the plots in the right column.
This indicates that a trained Transformer model is able to distinguish between GS-solved and GS-unsolved presentations to a good extent, albeit not perfectly.\footnote{
Note also that t-SNE admits a hyperparameter known as ``perplexity", and the projections learned by t-SNE depend on the hyperparameter \cite{wattenberg2016how}.
Thus, in general, t-SNE plots must be interpreted with care.
The plots shown in \cref{fig:tsne_embeddings} were all made with the perplexity value of $30$.
We checked however that the clusters of GS-solved and GS-unsolved presentations continue to exists at a broad range of perplexity values.}

Note that the training dataset contained no information about the ease of solvability of a presentation.
It also did not contain many presentations of the Miller--Schupp series itself.
Instead, it contained presentations that are AC-equivalent to the Miller--Schupp series presentations.
Our observation that a Transformer model trained on this dataset can distinguish between the GS-solved and GS-unsolved presentations indicates that:
\begin{enumerate}[label=\alph*)]
	\item There likely exists an invariant at the level of the ``language" of the balanced presentations that distinguishes GS-solved vs GS-unsolved presentations.
	\item This invariant survives application of thousands of AC-moves we used to generate the training examples in our dataset.
\end{enumerate}

    % !TEX root = ../ac_paper.tex

\section{Global hardness measure}\label{sec:global}

In \cref{ss:greedy_search}, we explored a greedy approach to finding AC-trivializations of a presentation \(\pi\).
Specifically, the goal was to construct a sequence of presentations \((\pi_0, \dots, \pi)\), where \(\pi_0\) is the trivial presentation, such that each consecutive pair in the sequence is related by an AC-move.
Furthermore, at each step \(k\), the presentation \(\pi_k\) was chosen to have the shortest possible length among all presentations connected to \(\pi_{k+1}\) via an AC-move.
In general, the length of a presentation in an AC-trivialization tends to exceed the length of the original presentation being trivialized.
The minimum increase in length across all possible AC-trivializations serves as an invariant of the presentation, its \textit{$\ell$-increase}, which is correlated with how hard it is to trivialize, as depicted in \cref{fig:performance} (B).

In this section, we provide a principled summarization of this invariant across all presentations.
This invariant serves as a measure of the hardness of the AC trivialization problem and provides a basis for comparison with other similar \textit{path-to-the-base} search problems.

\subsection{Intuition}\label{s:global_hardness_intuition}

Defining a global $\ell$-increase hardness measure by aggregating the $\ell$-increase hardness of every presentation suffers from overcounting.
This is because paths with $\ell$-increase 0 exist, and for such paths, trivializing one endpoint automatically trivializes the other at no additional cost.

Let $\Gamma_k$ be the induced subgraph containing all presentations with length at most $k$.
Let us compare the components of $\Gamma_k$ and $\Gamma_{k+1}$.\footnote{Recall that the \textit{components} of a graph refer to the equivalence classes of its nodes under the relation that identifies two if there is a path between them.}

Consider a node $v$ of length $k+1$.
If $v$ cannot be joined to a component of $\Gamma_k$ in $\Gamma_{k+1}$, then $v$ belongs to a new component that is said to be \textit{born} at $k+1$.
If $v$ can be joined to a component of $\Gamma_k$, then its hardness need not be considered since the path is of $\ell$-increase 0 and trivializing any presentation in the component with smaller length is more costly.
If $v$ connects two distinct components, say born at $k_1 \leq k_2$, then the component born at $k_2$ is said to \textit{die} at $k+1$.
By a similar reasoning, this component contributes $(k+1) - k_2$ to the overall hardness of the problem, the difference between its death and birth values.
If a component persists indefinitely, we assign it a death value of \(\infty\).

We propose that the multiset consisting of all such pairs $(\text{birth}, \text{death})$, excepting the component defined by the trivial presentation, serves as a principled hardness measure for the AC trivialization problem and any other \emph{path-to-the-base} search problem on a graph equipped with weights on its nodes.

\subsection{Formalization}

For any such problem, this multiset coincides with a key invariant from topological data analysis: the barcode of the persistent reduced homology in degree 0 of the associated filtered based graph.
Let us now explain this construction.

\subsubsection{Homology}

First, consider a graph \(G = (V, E)\), which we omit from the notation when convenient.

Let us work over the field with two elements and define \(C_0\) and \(C_1\) as the vector spaces generated by \(V\) and \(E\), respectively.
Explicitly, the elements in \(C_0\) and \(C_1\) are finite sums of vertices and edges, respectively.

Define the boundary map \(\partial\colon C_1 \to C_0\) on basis elements by:
\[
\partial(v, w) = v + w.
\]

The \textit{homology in degree $0$} of \(G\), denoted \(H_0\), is the quotient of the vector space \(C_0\) by \(\operatorname{img}(\partial)\):
\[
H_0 = \frac{C_0}{\operatorname{img}(\partial)}.
\]

The dimension of this vector space corresponds to the number of connected components of \(G\).

If \(G\) is a subgraph of \(G'\), then the inclusion of basis elements induces a well-defined linear map:
\[
H_0(G) \to H_0(G').
\]

When the graph is equipped with a preferred node, i.e., when it is based, it is common to quotient $C_0$ by the subspace generated by the base node.
The resulting invariant is called the \textit{reduced homology in degree $0$} of the based graph.

\subsubsection{Persistent homology}

Now, consider a filtered graph:
\[
\Gamma_0 \leq \Gamma_1 \leq \Gamma_2 \leq \dotsb,
\]
i.e., a collection of graphs with $\Gamma_i$ a subgraph of $\Gamma_{i+1}$.
Applying the homology construction provides a collection of vector spaces and linear maps, termed the persistent homology in degree 0 of the filtered graph:
\[
H_0(\Gamma_0) \to H_0(\Gamma_1) \to H_0(\Gamma_2) \to \dotsb.
\]

The \textit{barcode} of this diagram is a multiset of half-open intervals, called \textit{bars}, denoted by \(\operatorname{bar}(H_0) = \{[b, d)\}\).
The barcode is characterized the following property: for any pair of non-negative integers \(i \leq j\), the rank of the map \(H_0(\Gamma_i) \to H_0(\Gamma_j)\) equals the number of bars (counted with multiplicity) containing the interval \([i, j]\).
In particular, the dimension of \(H_0(\Gamma_i)\) is equal to the number of bars containing \(i\).

If all graphs in the filtered graph are based at the same node, then we can perform the reduced homology construction and obtain the barcode of the \textit{persistent reduced homology in degree $0$}.

\subsubsection{Filtration of the AC graph}

If $\Gamma_k$ is the induced subgraph of the AC graph containing all presentations with length at most $k$.
We obtain a filtration
\[
\Gamma_0 \leq \Gamma_1 \leq \Gamma_2 \subseteq \dotsb
\]
whose persistent reduced homology in degree 0 gives a barcode agreeing with the hardness multiset defined in \cref{s:global_hardness_intuition}.
Although beyond the scope of this paper, we mention that the absence of infinite bars in this barcode is equivalent to the AC conjecture.

\subsection{Experimental results}

To compute an approximation of the barcode for both AC- and AC$'$-trivialization problems,
we begin by defining a finite approximation of their respective filtered graphs.
For $L \in \{11, \dots, 16\}$ we take the induced subgraph containing all nodes admitting a trivialization path containing only presentation of length at most $L$.
The upper bound was imposed by our computational resources, whereas the lower bound was chosen to highlight non-trivial behavior.

For each such approximation we compute its hardness multiset (the barcode of persistent reduced homology in degree 0) using \texttt{giotto-TDA} \cite{tauzin2021giotto}\footnote{Specifically, its binding of the \texttt{SimplexTree} data structure introduced in \cite{boissonnat2014simplex} and implemented in \texttt{GUDHI} \cite{maria2014gudhi}.}.
We then count the number $h_i$ of pairs $(b, d)$ with $d-b = i$.

For the AC moves we have:

\medskip\hfil
\begin{tabular}{|c|c|c|c|c|c|}
	\hline
	$L$ & $nodes$ & $edges$ & $h_1$ & $h_2$ & $h_3$ \\ \hline
	11 & 350356 & 1002439 & 4 & 0 & 0 \\ \hline
	12 & 791140 & 2251375 & 16 & 0 & 0 \\ \hline
	13 & 3238052 & 9321629 & 72 & 4 & 0 \\ \hline
	14 & 7199908 & 20573343 & 144 & 4 & 0 \\ \hline
	15 & 29243812 & 84391763 & 508 & 52 & 8 \\ \hline
	16 & 64623652 & 185162236 & 1034 & 88 & 20 \\ \hline
\end{tabular}

\medskip
Whereas for AC$'$ moves we have:

\medskip\hfil
\begin{tabular}{|c|c|c|c|c|c|}
	\hline
	$L$ & $nodes$ & $edges$ & $h_1$ & $h_2$ & $h_3$ \\ \hline
	11 & 350356 & 655928 & 19 & 0 & 0 \\ \hline
	12 & 791140 & 1467080 & 67 & 0 & 0 \\ \hline
	13 & 3238052 & 6107112 & 243 & 16 & 0 \\ \hline
	14 & 7199908 & 13414744 & 483 & 16 & 0 \\ \hline
	15 & 29243812 & 55306744 & 1819 & 136 & 32 \\ \hline
	16 & 64623652 & 120824232 & 3923 & 208 & 80 \\ \hline
\end{tabular}

\medskip
Taking the quotient of the values of both tables, we observe that, as expected, the number of nodes remains unchanged.
Perhaps unsurprisingly, the number of edges scales consistently by approximately 1.5.
More unexpectedly, we find that the scaling factor remains fairly constant across the other columns.
This suggests that the overall complexity of the problem is preserved under the reformulation, albeit consistently scaled.

\hfil
\begin{tabular}{|c|c|c|c|c|c|}
	\hline
	$L$ & $nodes$ & $edges$ & $h_1$ & $h_2$ & $h_3$ \\ \hline
	11 & 1 & 1.5283 & 0.2105 & - & - \\ \hline
	12 & 1 & 1.5346 & 0.2388 & - & - \\ \hline
	13 & 1 & 1.5264 & 0.2963 & 0.2500 & - \\ \hline
	14 & 1 & 1.5336 & 0.2981 & 0.2500 & - \\ \hline
	15 & 1 & 1.5259 & 0.2793 & 0.3824 & 0.25 \\ \hline
	16 & 1 & 1.5325 & 0.2636 & 0.4231 & 0.25 \\ \hline
\end{tabular}
	% !TEX root = ../ac_paper.tex

\section{Local analysis}\label{sec:local}

% Throughout this work, we have focused on how hard presentations are to solve, with special attention to difficulty for different algorithmic methods.
Hardness is a global property, as the typical paths encountered in this work are significantly longer--often by orders of magnitude--than the length of the trivialized presentation.
In this section, we study how much of a presentation's hardness can be predicted by its local structure, with special attention given to topological features.  

The simplest feature of a presentation that we consider, beyond its length, is the number of presentations that can be connected to it via a small number of AC moves.

In \cref{ss:nbhd_sizes}, we analyze the distribution of these neighborhood sizes for the 1,190 presentations in the Miller--Schupp series considered in \cref{sec:ppo}, along with their PPO-solved and PPO-unsolved labels obtained from the experiments described therein.  
We find that the PPO-unsolved presentations are concentrated within a few specific neighborhood sizes, whereas the PPO-solved presentations exhibit a more uniform distribution across a broader range of sizes.  

\cref{ss:comparison_local_features} employs a supervised learning approach using random forests to evaluate the predictive power of various neighborhood features in the MS series.
These features are analyzed for their ability to predict the hardness of presentations, specifically the PPO-solved/unsolved binary label in the MS series.
Our results reveal that certain topological features of the graph improve on the predictive power of neighborhood size, as studied in the first subsection below.

\subsection{Neighborhood sizes}\label{ss:nbhd_sizes}

Let us return to our previously considered data set of 1190 presentations in the Miller--Schupp series for \(n \leq 7\) and \(\length(w) \leq 7\).
Using the methods described in \cref{sec:ppo}, we trained a PPO agent that successfully solved 417 of these presentations.
We will refer to the set of these 417 presentations as PPO-solved and the remaining 773 presentations as PPO-unsolved.
Our goal is to analyze the relationship between these labels and the sizes of their respective AC neighborhoods.
A presentation is considered to be in the \(k\)-step neighborhood of another if they can be connected by applying at most \(k\) AC-moves.

\subsubsection{Neighborhood size distribution}

\begin{figure}
	\includegraphics[scale=.4]{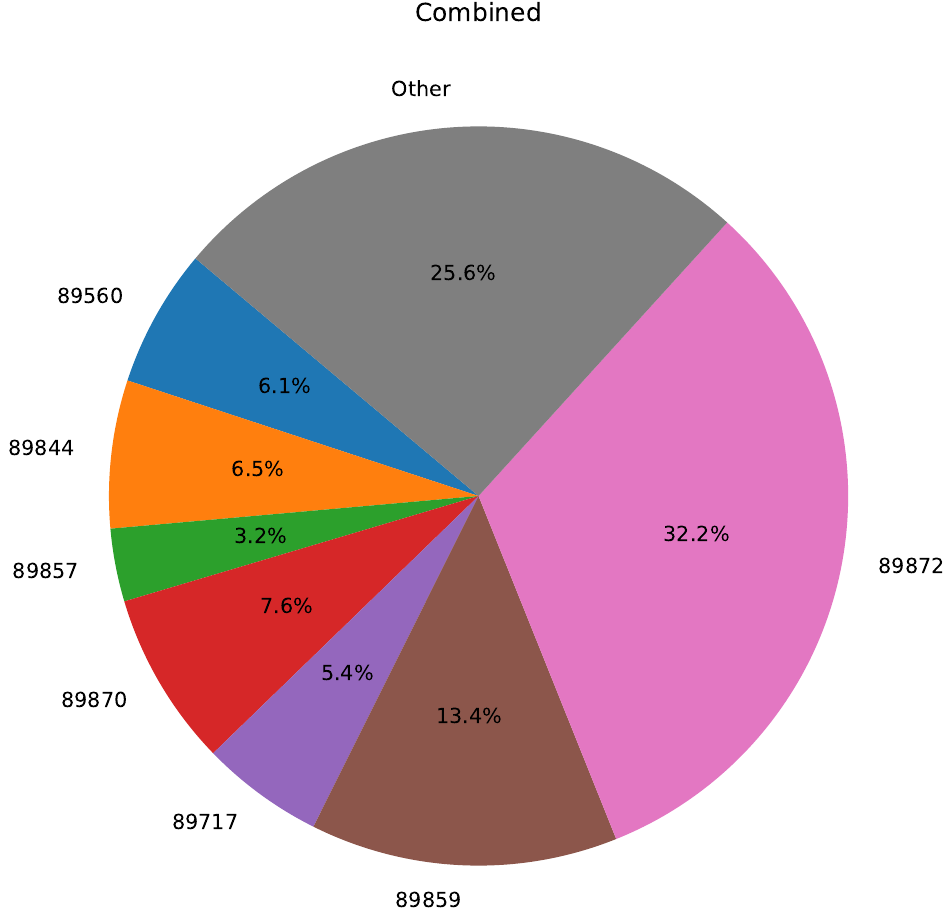}
	\caption{Sizes of the 5-step neighborhood of all considered presentations in the Miller--Schupp series. We group neighborhood sizes whose representation is below 2.5\%.}
	\label{fig:prime_combined_pie}
\end{figure}

\begin{figure}
	\centering
	\includegraphics[scale=.35]{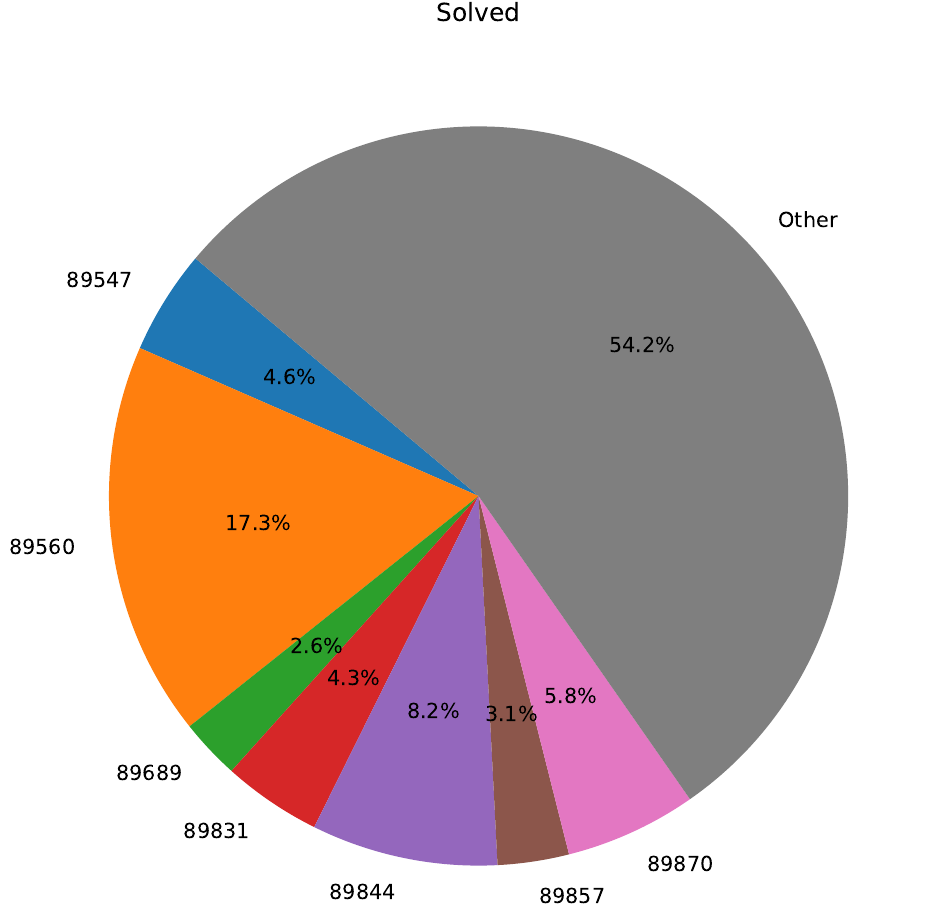}
	\
	\includegraphics[scale=.35]{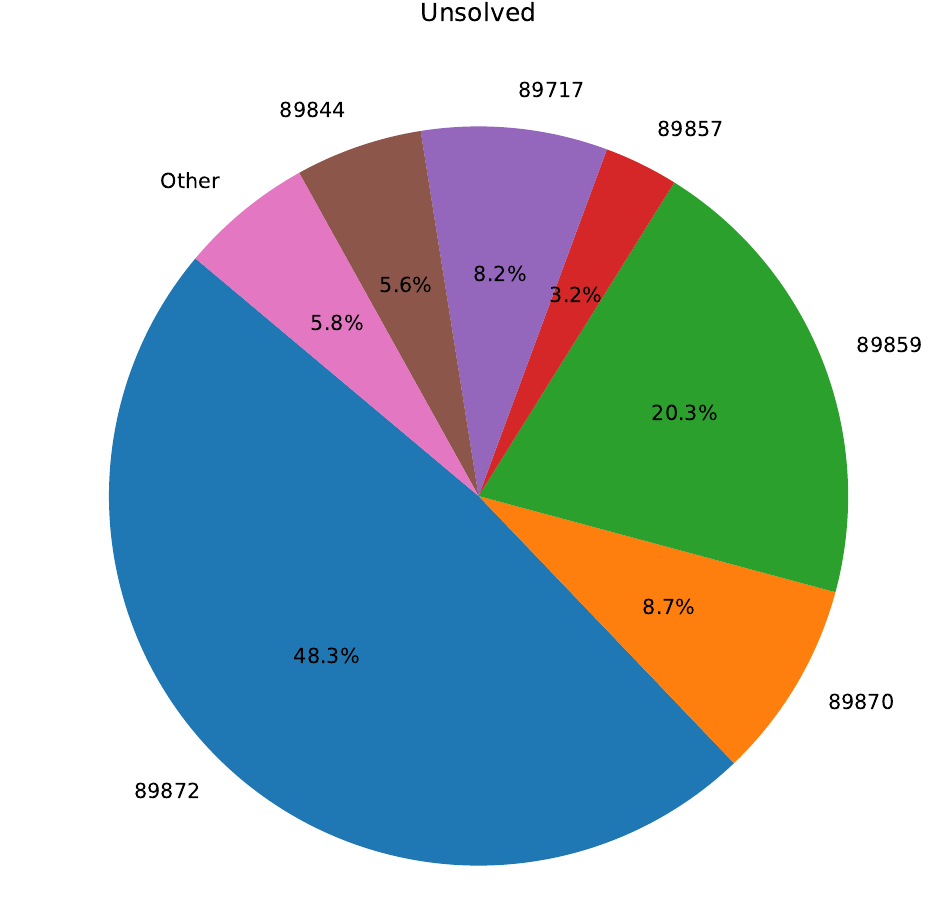}
	\caption{Pie charts for the neighborhood size of PPO-solved and PPO-unsolved presentations. We grouped sizes with representation below 2.5\%.}
	\label{fig:prime_pies}
\end{figure}

We will focus on 5-step neighborhoods of these presentations, referring to the number of presentations in them as the size of the neighborhood.
There are 131 distinct neighborhood sizes in our dataset.
Their basic statistics are
\[
\begin{tabular}{cccccc}
	\toprule
	\textbf{Min} & \textbf{Max} & \textbf{Mean} & \textbf{Median} \\
	\midrule
	72,964 & 89,872 & 89,532 & 89,859 \\
	\bottomrule
\end{tabular}
\]
This distribution of neighborhood sizes exhibits a number of curious feature.
For example, it is not merely a discretization of a typical continuous distribution a la Gaussian or uniform distribution.
The entire distribution fits in a relatively narrow band of sizes between 70k and 90k, heavily skewed toward the upper end.
In a discretization of a typical continuous distribution, each individual value would have a very small representation. 
However, in the case at hand, individual size values constitute large fractions of the entire pie chart (cf. \cref{fig:prime_combined_pie} and \cref{fig:prime_pies}).

Specifically, the largest neighborhood size accounts for nearly a third of all considered presentations.
However, it represents only 2.4\% of PPO-solved presentations, while constituting almost half (48.3\%) of the PPO-unsolved presentations.
For more details, please refer to \cref{fig:prime_pies}.

In contrast, using BFS, these proportions are 7.1\% and 52.5\%, respectively.

Another notable feature visible in \cref{fig:prime_pies} is that just three neighborhood sizes account for over three-quarters of all PPO-unsolved presentations.
When considering six neighborhood sizes, this proportion rises to 96.9\%.
In fact, only twelve neighborhood sizes are present among PPO-unsolved presentations, whereas all 131 sizes appear among PPO-solved presentations.
The most common neighborhood size for PPO-solved presentations is 89,560, representing only 17.3\% of them.
Moreover, 54.2\% of all PPO-solved presentations have a neighborhood size shared by less than 2.5\% of other PPO-solved presentations.

\begin{figure}
	\centering
	\includegraphics[scale=.34]{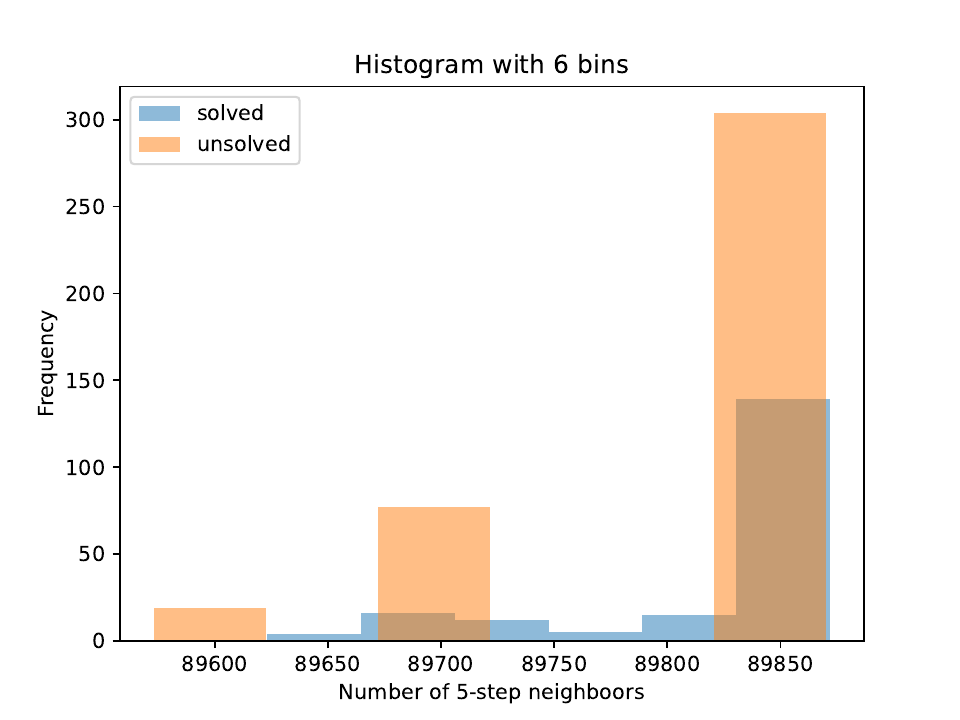}
	\includegraphics[scale=.34]{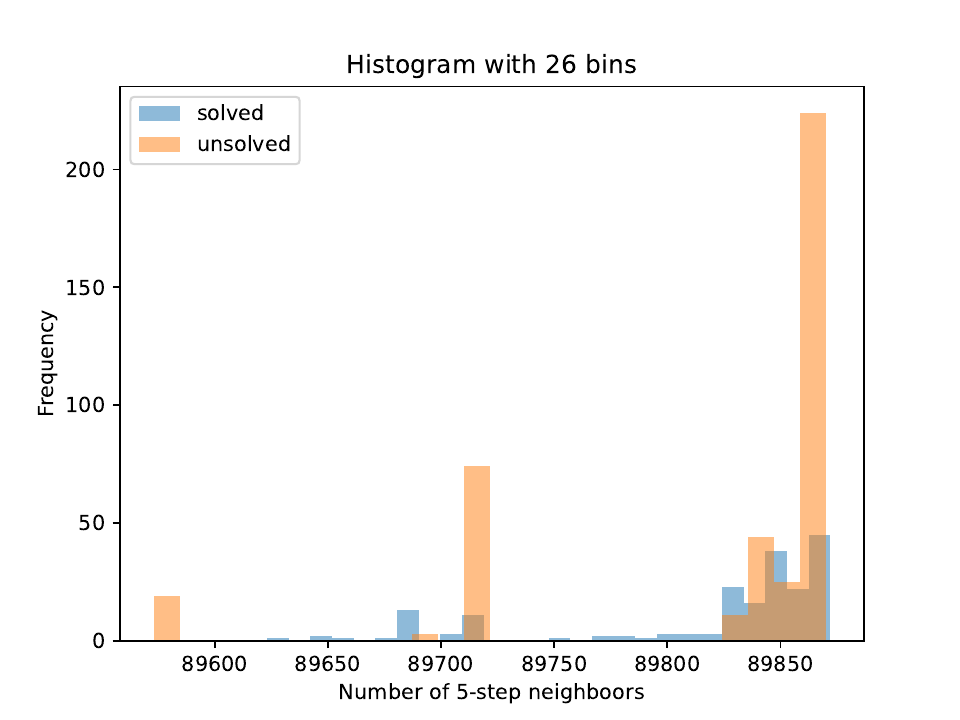}
	\caption{Histograms with 6 and 26 bins respectively of the neighborhood sizes of the 417 PPO-solved and 773 PPO-unsolved presentations.}
	\label{fig:prime_histogram}
\end{figure}

As we observed, having a maximal neighborhood size provides significant insight into whether a presentation is labeled as PPO-solved or PPO-unsolved.
Additionally, the minimum neighborhood size among PPO-unsolved presentations--89,573--is also quite telling, as 54\% of PPO-solved presentations have neighborhood sizes smaller than this value.
This percentage can be further improved by considering that the neighborhood sizes of PPO-unsolved presentations are concentrated within three specific bands.
Please refer to \cref{fig:prime_histogram} for more details.
We find that 64.3\% of PPO-solved presentations fall outside the three bands \([89,575, 89,575]\), \([89,715, 89,831]\), and \([89,844, 89,872]\), which together contain over 99\% of PPO-unsolved presentations.
By removing the first band, which included only one size, and narrowing the last band to \([89,859, 89,872]\), the resulting two bands now encompass the neighborhood sizes of over 88\% of PPO-unsolved presentations, while their complement includes 77.7\% of PPO-solved presentations.

\subsection{A comparison of local features}\label{ss:comparison_local_features}

In this subsection we use a supervised learning technique based on random trees to predict the label PPO-solved and PPO-unsolved, using a battery of different features of the 5-step neighborhoods in the MS series: number of vertices, number of edges, barcode vector, maximal length of a bar, sum of the barcode vector entries, persistence entropy, as well as basic features of the presentation: lengths of each relator, length of the presentations.

The data set of 1190 presentations (417 PPO-solved and 773 PPO-unsolved) was randomly split into training subset and test subset in proportion 4:1.
We used XGBoost Classifier on the training set and compared F1 scores on the test set.
All code is available at our GitHub repository.

The base line model is provided by considering the length of the presentation as the only feature.
The F1 score of the resulting classifier is 0.885. %0.838.
Unsurprisingly, the neighborhood size outperforms this feature and provides a new baseline with an F1 score of 0.930. %0.879.
It is interesting to mention that the correlation between length and neighborhood size is low (0.217) compared to, for example, that with the number of edges (0.999).

\begin{figure}
	\includegraphics[scale=.35]{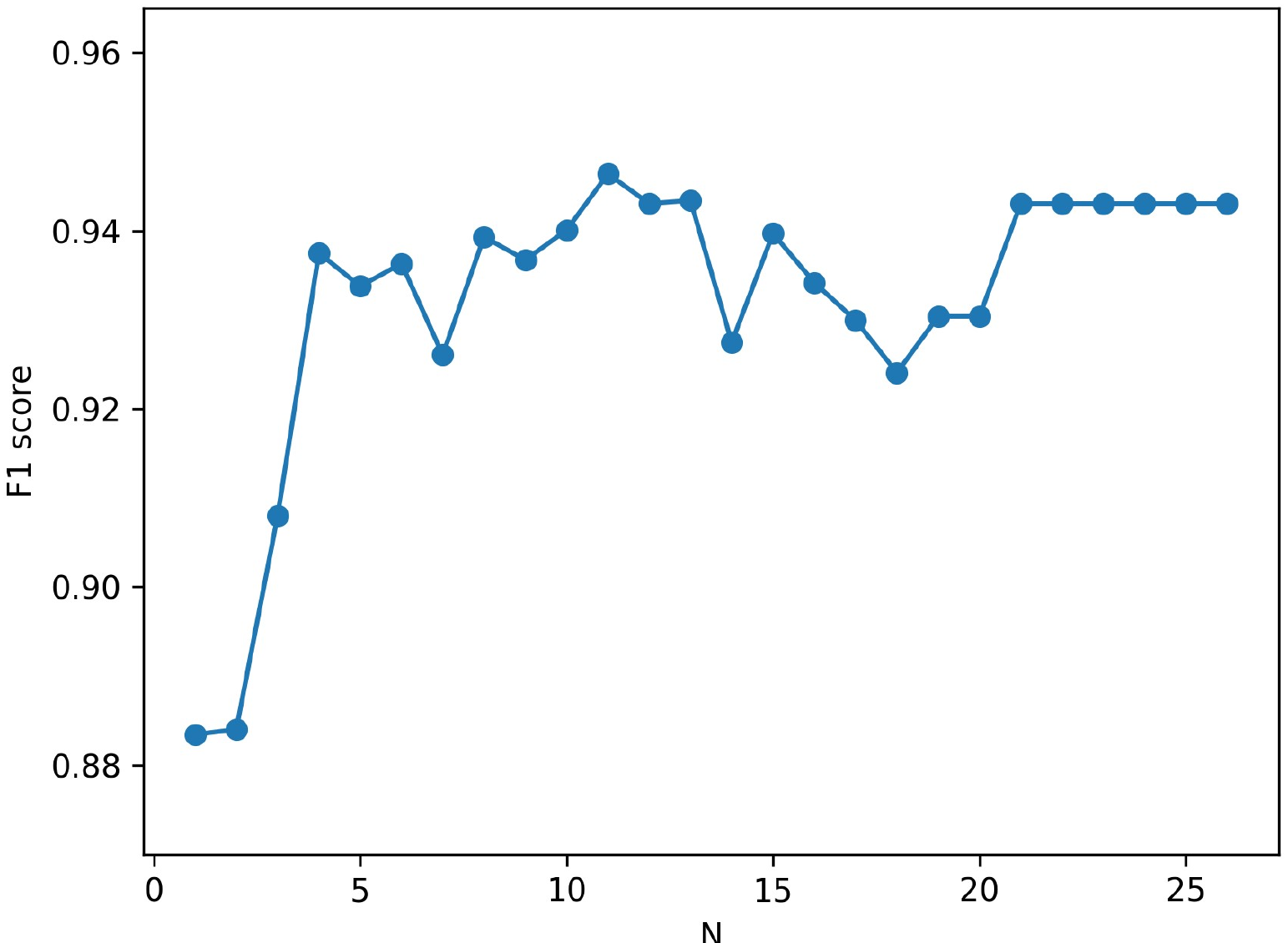}
	\caption{F1 score obtained taking the first \(N\) entries of the barcode vector}
	\label{fig:barcode_f1_scores}
\end{figure}

Another notable result is that purely topological features can outperform neighborhood size in predictive tasks. 
By computing the barcode, as defined in the previous section, for the filtered graph induced by the 5-step neighborhood and counting the number of bars \([b, d)\) for each value of \(d - b\), we construct a \textit{barcode vector}. 
Using this barcode vector as a single feature achieves an F1 score of 0.943. 
For an analysis of the performance of specific subsets of the barcode vector entries, refer to \cref{fig:barcode_f1_scores}.

An exhaustive search over all feature subsets enabled the selection of the optimal set of features based on the F1 score of the corresponding classifier. 
The selected subset includes the size of the neighborhood, its barcode vector, and the sum of the barcode entries. 
This classifier achieves an F1 score of 0.962.

% The selected subset turned out to be purely topological, specifically the barcode vector and its cumulative features: sum, max, and persistence entropy.
% This subset achieved an F1 score of 0.949.

\begin{figure}
	\includegraphics[scale=.6]{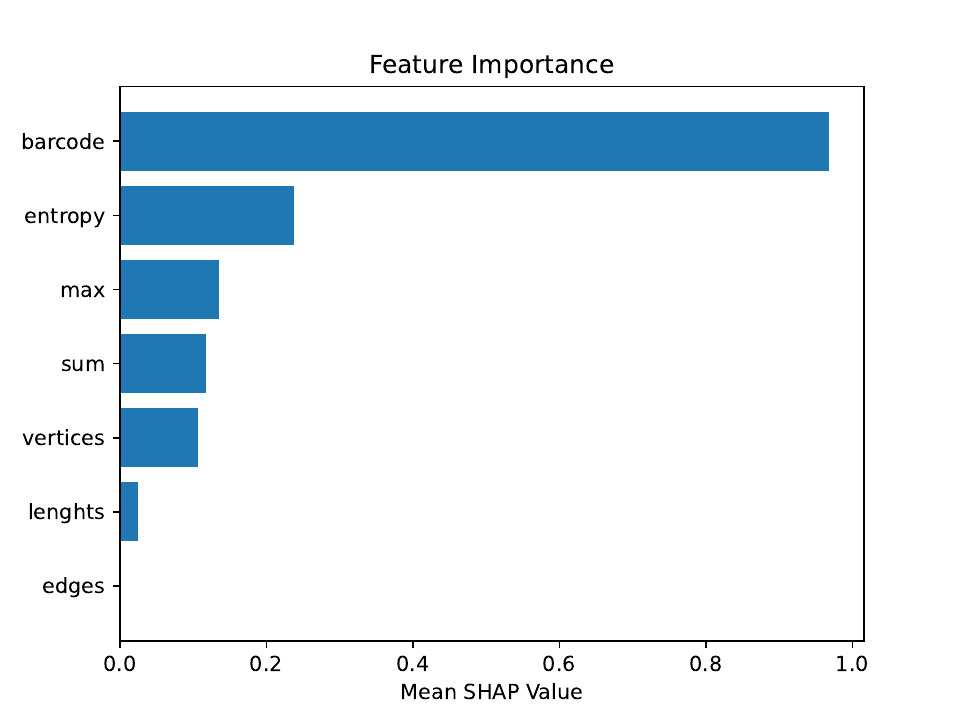}
	\caption{Feature importance measured by SHAP values for each feature, calculated in general setting, for the test set.}
	\label{fig:SHAP_values}
\end{figure}

We used the tree explainer technique and SHAP values to analyze the importance of each of considered features when building a modeled with all of them (number of vertices, number of edges, barcode vector, maximal length of a bar, sum of the barcode vector entries, persistence entropy, length of each relator, and length of the presentation).
The following features stood out as the most influential: the lower entries of the barcode vector, number of vertices, entropy, and the sum and the maximum of the barcode vector entries.
Aggregated SHAP values are presented in \cref{fig:SHAP_values}.
The feature `lengths' includes accumulated values of three features: length of each relator and length of the presentation.

In addition to the above features derived from persistent homology, we also explored two other types of features.  
Surprisingly, these features only marginally improved the accuracy of PPO-solved/unsolved predictions.  
The first type was based on node centrality, while the second focused on the spectral properties of the neighborhood graphs.  
The latter is particularly intriguing due to its connection to Markov processes and the well-known relationship between random walks on graphs and the Laplacian.  
More analysis is needed to fully understand the relationship between these local graph features and hardness.
	
% !TEX root = ../ac_paper.tex

\section{Stable Andrews--Curtis conjecture and unknot diagrams}\label{sec:stable}

Our focus in this work has so far been on the standard Andrews--Curtis conjecture. In this section, we turn our attention to a closely related variant, known as the “stable” or “weak” Andrews--Curtis conjecture, which is also of great interest, not least because of its connections to algebraic topology. (See e.g. \cite{Meier2016,Bagherifard2021} for recent work on this connection.)

\subsection{Stable Andrews--Curtis conjecture} \label{subsec:stable}

The stable Andrews--Curtis conjecture, also called the weak Andrews--Curtis conjecture, allows for two further moves in addition to the standard AC-moves (AC1)--(AC3), for the purpose of trivializing a presentation. Namely,
\begin{enumerate}[label=(AC\arabic*)]
	\setcounter{enumi}{3}
	\item Include a new generator and a trivial relator, i.e. replace $\angles{x_1, \dots, x_n \mid r_1, \dots, r_n}$ by $\angles{x_1, \dots, x_n, x_{n+1} \mid r_1, \dots, r_n, x_{n+1}}$.
	\item Remove a trivial relator and the corresponding generator, i.e. the inverse of (AC4).
\end{enumerate}

Two balanced presentations that can be related by a sequence of transformations (AC1)--(AC5) are said to be \textit{stably AC-equivalent}. A presentation that is stably AC-equivalent to a trivial presentation is referred to as \emph{stably AC-trivial}. The stable Andrews--Curtis conjecture states that every balanced presentation is stably AC-trivial.

As the conjecture allows for addition and removal of new generators and trivial relators, the following lemma is quite useful.

\begin{lemma}[Substitution and Removal]\label{lem:substitution}
	Let $P=\langle x_1,\ldots, x_n, y \mid r_1,\ldots, r_n, y^{-1}w\rangle$ be a presentation of the trivial group, where $w$ is a word in $x_1,\ldots,x_n$. Then $P'=\langle x_1,\ldots, x_n \mid r_1',\ldots, r_n'\rangle$ is stably AC-equivalent to $P$, where $r_i'$ is $r_i$ with all occurrences of $y$ replaced by $w$.
\end{lemma}

\begin{proof}
	First, we substitute $w$ for $y$ in $r_1,\ldots,r_n$, giving $\widetilde{P}=\langle x_1,\ldots,x_n,y\mid r_1',\ldots,r_n',y^{-1}w\rangle$ (cf. \cref{def:ac-sub}). Since $r_1',\ldots,r_n'$ do not contain $y$, we can define the surjective homomorphism $\phi:\widetilde{P}\longrightarrow P'$ by $\phi(x_i)=x_i$ and $\phi(y)=w$. Since $\widetilde{P}$ is trivial, $P'$ is trivial. Therefore, the normal closure of $r_1',\ldots, r_n'$ gives the free group generated by $x_1,\ldots, x_n$, which means $w$ can be written as a product of conjugates of the $r'_i$, namely, $w=w_1 (r_{i_1}')^{\pm1}w_1^{-1}w_2 (r_{i_2}')^{\pm 1}w_2^{-1}\cdots w_m (r_{i_m}')^{\pm 1}w_m^{-1}$, for words $w_i$ in the $x_i$ and $i_k \in \{1,\ldots,n\}$. Note that $m$ may be much larger than $n$. Now using (AC1)--(AC3) we can transform  $y^{-1}w$ to $w_1 (r_{i_1}')^{\pm1}w_1^{-1}\cdots w_m (r_{i_m}')^{\pm 1}w_m^{-1}w^{-1}y$ which equals $y$, and then we can remove $y$ by (AC5).
\end{proof}

We note that, much like the substitution move in the standard Andrews--Curtis conjecture, the \emph{substitution and removal} move can be interpreted as a supermove in the case of stable Andrews--Curtis conjecture. For substitution, the number of AC moves constituting the supermove is a linear function of the lengths of the relators (see \cref{sec:AC}). In contrast, our proof of \cref{lem:substitution} does not yield a similar result for the \emph{substitution and removal} move due to the absence of a bound on $m$. Finding this bound, or discovering an alternative proof of the lemma that establishes such a bound on the number of AC moves packaged into this supermove, would be very useful.

The families of presentations studied in this paper---namely, the Akbulut--Kirby series and the Miller--Schupp series---also serve as potential counterexamples to the stable Andrews--Curtis conjecture.\footnote{Specifically, $\AK(3)$---the shortest potential counterexample for the standard Andrews--Curtis conjecture is also a potential counterexample for the stable Andrews--Curtis conjecture.}
This naturally leads to a broader question about the relationship between the stable Andrews--Curtis conjecture and its standard counterpart: do there exist any stably AC-trivial presentations of the trivial group that could serve as potential counterexamples to the standard Andrews--Curtis conjecture?
A construction of such candidate presentations from Wirtinger presentations of knot diagrams of the unknot was given in \cite{MMS}. We will review their construction in \cref{subsec:ac-unknot}. In \cref{sec:proofs}, we will prove that a substantial subclass of these presentations are, in fact, AC-trivial.
Furthermore, we will conjecture that any presentation obtained in this way is also AC-trivial.

\subsection{Stably AC-trivial presentations from knot diagrams} \label{subsec:ac-unknot}

\subsubsection{Stably AC-Trivial Presentations From Wirtinger Presentations}

We recall briefly the construction of the Wirtinger presentations of a knot group. Given any oriented knot diagram with $n$ crossings, we assign generators $\{x_i\}_{i=1}^n$ to each of the arcs, and each crossing gives us a relator $\{r_i\}_{i=1}^n$. In our notation, the left crossing in \cref{fig:wp} corresponds to the relator $y=x^{-1}zx$ and the right one gives $y=xzx^{-1}$. These generators and relators form a balanced presentation of the fundamental group of the knot complement. Moreover, there exists an ordering of the relators and a choice of $\pm 1$ as exponents such that the relators satisfy the equation
$$\prod_{i=1}^nr_i^{\pm1}=1.$$
Thus any relator $r_k$ can be eliminated, giving $\langle x_1,\ldots,x_n\mid r_1,\ldots,r_{k-1},r_{k+1},\ldots,r_n\rangle$, without changing the underlying group.

\begin{figure}
	\centering
	\begin{tikzpicture}
		\draw[thick,->] (0,0)--(1,1);
		\draw[thick,->] (1,0)--(0.6,0.4);
		\draw[thick,->] (0.4,0.6)--(0,1);
		\node at (1.2,1.2){$x$};
		\node at (-0.2,1.2){$y$};
		\node at (1.2,0){$z$};
		\node at (0.5,-1){$y=x^{-1}zx$};

		\draw[thick,->] (4,1)--(3,0);
		\draw[thick,->] (4,0)--(3.6,0.4);
		\draw[thick,->] (3.4,0.6)--(3,1);
		\node at (4.2,1.2){$x$};
		\node at (2.8,1.2){$y$};
		\node at (4.2,0){$z$};
		\node at (3.5,-1){$y=xzx^{-1}$};

	\end{tikzpicture}
	\caption{Types of Crossings}
	\label{fig:wp}
\end{figure}

When the knot is an unknot, the fundamental group of the knot complement is $\mathbb{Z}$, which can be generated by any generator $x_i$. If we add a relator $w$, which is any word in $x_1,\ldots,x_n$ with exponent sum $\pm1$, we get a balanced presentation $\langle x_1,\ldots,x_n\mid r_1,\ldots,r_{k-1},r_{k+1},\ldots,r_n,w\rangle$ of the trivial group. The result below establishes that every such balanced presentation is stably AC-trivial.

\begin{proposition}[Myasnikov, Myasnikov, and Shpilrain, \cite{MMS}] \label{prop:unknot-stable}
	Let $\angles{ x_1,\ldots, x_n\, \mid \, r_1, \ldots, r_n }$ be the Wirtinger presentation of an unknot diagram. Then for any $k=1,\ldots,n$ and word $w$ in the $x_i$ with exponent sum $\pm 1$, the balanced presentation of the trivial group $\langle x_1,\ldots, x_n\,|\, r_1,\ldots, r_{k-1}, r_{k+1},\ldots, r_n, w\rangle$ is stably Andrews--Curtis-trivial.
\end{proposition}

The proof of this proposition relies on the realization of Reidemeister moves for knot diagrams as stable AC moves for the associated balanced presentation of the trivial group. The details of the proof are given in \cref{app:reid}.

\subsubsection{Balanced presentations with fewer generators.}
\label{sec:unknot-gives-potential-counterexamples-ac}

Starting with a balanced presentation coming from \cref{prop:unknot-stable}, one may recursively apply stable AC moves to reduce the number of generators and relators.
The resultant presentations are all stably AC-trivial by construction, but they could, a priori, serve as potential counterexamples to the standard Andrews--Curtis conjecture \cite{MMS}.\footnote{We will show in the next section, however, that a large subset of these are, in fact, AC-trivial.}

\begin{figure}
	\centering
	\begin{tikzpicture}
		\draw[thick] (0,0)--(1.8,0);
		%x3
		\draw[thick] (2.2,0)--(5.8,0);
		%x2
		\draw[thick,rounded corners] (6.2,0)--(7,0)--(7,1.5)--(2,1.5)--(2,-7)--(7,-7)--(7,-5)--(6.2,-5);
		%x1
		\draw[thick,rounded corners] (0,0)--(-2,0)--(-2,-0.8);
		%x3
		\draw[thick,rounded corners] (-2,-1.2)--(-2,-6)--(0,-6)--(0,-4.2);
		%x4
		\draw[thick,rounded corners] (0,-3.8)--(0,-2.2);
		%x5
		\draw[thick,rounded corners] (0,-1.8)--(0,-1)--(-3,-1)--(-3,-2)--(-2.2,-2);
		%x6
		\draw[thick,rounded corners] (-1.8,-2)--(1.8,-2);
		%x7
		\draw[thick,rounded corners] (1.8,-4)--(-1,-4)--(-1,-5)--(-0.2,-5);
		%x12
		\draw[thick,rounded corners] (0.2,-5)--(1.8,-5);
		%x13
		\draw[thick,rounded corners] (2.2,-4)--(4,-4)--(4,-0.2);
		%x11
		\draw[thick,rounded corners] (4,0.2)--(4,0.75)--(6,0.75)--(6,-6)--(5,-6)--(5,-5.2);
		%x10
		\draw[thick,rounded corners] (2.2,-5)--(5.8,-5);
		%x14
		\draw[thick,rounded corners] (5,-4.8)--(5,-2)--(4.2,-2);
		%x9
		\draw[thick,rounded corners] (2.2,-2)--(3.8,-2);
		%x8

		\node at (4.5,1.5){$>$};
		\node at (3,0){$<$};
		\node at (0,0){$<$};
		\node at (-2,-3){\rotatebox{90}{$<$}};
		\node at (0,-3){\rotatebox{90}{$>$}};
		\node at (-1,-1){$<$};
		\node at (1,-2){$>$};
		\node at (3,-2){$>$};
		\node at (4.5,-2){$>$};
		\node at (6,-2.5){\rotatebox{90}{$>$}};
		\node at (3,-4){$<$};
		\node at (-0.5,-4){$<$};
		\node at (1,-5){$>$};
		\node at (3,-5){$>$};

		\node at (4.5,2){$x_1$};
		\node at (3,0.5){$x_2$};
		\node at (0,0.5){$x_3$};
		\node at (-2.5,-3){$x_4$};
		\node at (0.5,-3){$x_5$};
		\node at (-1,-0.5){$x_6$};
		\node at (1,-1.5){$x_7$};
		\node at (3,-1.5){$x_8$};
		\node at (4.5,-1.5){$x_9$};
		\node at (6.5,-2.5){$x_{10}$};
		\node at (3,-3.5){$x_{11}$};
		\node at (-0.5,-3.5){$x_{12}$};
		\node at (1,-5.5){$x_{13}$};
		\node at (3,-5.5){$x_{14}$};
	\end{tikzpicture}
	\caption{Unknot diagram}
	\label{fig:unknot}
\end{figure}

As an example of this procedure, consider the unknot diagram shown in \cref{fig:unknot}. The associated Wirtinger presentation is as follows,
\[
\begin{aligned}
	W=\left\langle x_1, x_2, \dots, x_{13}, x_{14} \right. \mid\, &x_1=x_{10} x_{14} x_{10}^{-1}, x_2=x_{10}^{-1} x_1 x_{10},\\
	&x_3=x_1^{-1} x_2 x_1, x_4=x_6^{-1} x_3 x_6,\\
	&x_5=x_{12} x_4 x_{12}^{-1},x_6=x_7^{-1} x_5 x_7, \\
	&x_7=x_4^{-1} x_6 x_4, x_8= x_1 x_7 x_1^{-1}, \\
	&x_9=x_{11}^{-1} x_8 x_{11}, x_{10}=x_{14} x_9 x_{14}^{-1} \\
	&x_{11}=x_2^{-1} x_{10} x_2, x_{12}=x_1^{-1} x_{11} x_1 \\
	& x_{13}=x_4 x_{12} x_4^{-1}, x_{14}=x_1 x_{13} x_1^{-1}\left. \right\rangle .
\end{aligned}
\]

Using stable AC moves, we can reduce the number of generators in $W$ to obtain several 3-generator families of presentations. Consider, for example, deleting the relator $r_6$, adding a word $w$, and eliminating generators in the following order through \cref{lem:substitution}: $x_1$, $x_2$, $x_3$, $x_4$, $x_5$, $x_7$, $x_8$, $x_9$, $x_{11}$, $x_{12}$, and $x_{13}$. We are left with a presentation of three generators $x_6$, $x_{10}$, and $x_{14}$. Renaming $x = x_{10}$, $y = x_{14}$, and $z=x_6$, the resultant presentation equals
\[
\begin{aligned}
	\angles{x, y, z \mid &
		% y^{-1} x^{-1} y x
		[y^{-1}, x^{-1}]
		y x^{-1} z^{-1} x y^{-1} x^{-1} [y^{-1}, x]
		%y^{-1} x y x^{-1}
		z x [y^{-1}, x^{-1}]
		% y^{-1} x^{-1} y x
		y x^{-1} z x y^{-1} x^{-1}, \\
		& % y^{-1} x y x^{-1}
		[y^{-1}, x] z^{-1} x
		%y^{-1} x^{-1} y x
		[y^{-1}, x^{-1}]
		y x^{-1} z x y^{-1}
		%x^{-1} y^{-1} x y
		[x^{-1}, y^{-1}]
		x y x^{-1} z^{-1} x y^{-1}
		%x^{-1} y^{-1} x y
		[x^{-1}, y^{-1}]
		x^{-1} z x y^{-1} x^{-1}, \\
		& w}.
\end{aligned}
\]
After choosing an appropriate $w$, we may eliminate one more generator, thus obtaining two-generator presentations through this method.

\begin{remark}
	Note, however, that if we were to obtain a 2-generator family without picking a word $w$ (and merely through recursive use of stable AC-moves), we would obtain an infinite family of the form $\angles{ x,y \mid r_1,w }$. Members of this family are all AC-trivial through the following argument. If $\angles{ x,y \mid r_1,w }$ is a presentation coming from an unknot diagram, then $\angles{ x,y \mid r_1 }$ is a presentation of $\mathbb{Z}$ with $r_1 = xy^{\pm 1}$, and any presentation of the form $\angles{ x,y \mid xy^{\pm 1},w }$ is easily AC-trivializable.
\end{remark}

\begin{remark}
	The unknot diagram of \cref{fig:unknot} was originally presented in \cite{MMS}. Their manuscript contained an unfortunate misprint in the 13th relator, where it is written as $x_{13}=x_5 x_{12} x_5^{-1}$.  The resultant presentation is not a Wirtinger presentation of any knot diagram, affecting the validity of one of their main theorems (\cite[Theorem 1.4]{MMS}). For more details, see \cref{app:mms}.
\end{remark}

\subsection{Knot diagrams give AC-trivial presentations} \label{sec:proofs}
In this section, we will prove that every presentation of the form
$\angles{ x_1,\ldots, x_n \mid r_1,\ldots, r_{k-1}, r_{k+1},\ldots, r_n, w}$ as in \cref{prop:unknot-stable}, is in fact AC-trivial. Moreover, we will show that presentations with fewer generators that come from recursively applying \cref{lem:substitution} are also AC-trivial.

The proofs of AC-triviality rely on the following lemma, which establishes that presentations with different choices of $w$ are all AC-equivalent.

\begin{lemma}
	\label{lem:all_ac_equiv}
	Let $\langle x_1,\ldots, x_n \mid r_1,\ldots,r_{n-1}\rangle$ be a presentation of $\mathbb{Z}$. Then there exists one word $x$ in the $x_i$, such that for any word $w$ that makes the group trivial, the balanced presentation of the trivial group $\langle x_1,\ldots, x_n \mid r_1,\ldots,r_{n-1}, w\rangle$ is AC-equivalent to $\langle x_1,\ldots, x_n \mid r_1,\ldots,r_{n-1}, x\rangle$.
\end{lemma}

\begin{proof}[Proof]
	Since $\langle x_1,\ldots, x_n \mid r_1,\ldots,r_{n-1}\rangle$ is a presentation of $\mathbb{Z}$, there exists one word $x$ that generates $\mathbb{Z}$. Then any $x_i$ can be given by the product of $x^{n_i}$ with a product of conjugates of $r_1,...,r_{n-1}$ for some $n_i\in\mathbb{Z}$.

	Suppose $w=x_{i_1}^{\epsilon_ 1}x_{i_2}^{\epsilon_2}\cdots x_{i_m}^{\epsilon_m}$ for $i_k\in \{1,\ldots,n\}$ and $\epsilon_k\in\mathbb{Z}$. For each $x_i$ in $w$, multiply $w$ by the corresponding product of conjugates of $r_1,...,r_{n-1}$ to replace $x_i$ by $x^{n_i}$. Then $w$ is changed to $x^p$ for some $p$. Since $x$ generates $\mathbb{Z}$, $p=\pm 1$.
\end{proof}

Note that the lemma holds for \emph{any} presentation $\langle x_1,\ldots, x_n \mid r_1,\ldots,r_{n-1}\rangle$ of $\mathbb{Z}$, and not just the ones coming from knot diagrams of the unknot. We will now use this lemma to prove our result for presentations of \cref{prop:unknot-stable}.

\begin{theorem} \label{thm:unknot}
	Let $\angles{ x_1,\ldots, x_n \mid r_1, \ldots, r_n }$ be the Wirtinger presentation of an unknot diagram. Then for any $k=1,\ldots,n$ and word $w$ in the $x_i$ with exponent sum $\pm 1$, the balanced presentation of the trivial group $\angles{ x_1,\ldots, x_n \mid r_1,\ldots, r_{k-1}, r_{k+1},\ldots, r_n, w}$ is Andrews--Curtis-trivial.
\end{theorem}

\begin{proof}
	The relation  $$\prod\limits_{i=1}^nr_i^{\pm1}=1$$ for a Wirtinger presentation of a knot diagram allows us to write any one relator $r_i$ as a product of other relators and their inverses. Hence, the presentations
	$$\angles{x_1, \ldots, x_n \mid r_1, \ldots, r_{k-1}, r_{k+1}, \ldots, r_n, w}$$ with different choices of $k \in \{1, \ldots, n\}$ and fixed $w$ are all AC-equivalent. Without loss of generality, we set $k=n$. As the presentation comes from a Wirtinger presentation, we can write it as
	\[
	\angles{ x_1,\ldots,x_n \mid x_1=z_2x_2z_2^{-1}, x_2=z_3x_3z_3^{-1},\ldots, x_{n-1}=z_{n}x_nz_{n}^{-1}, w },
	\]
	where $z_i\in \{x_1^{\pm 1},\ldots, x_n^{\pm 1}\}$.

	Using \cref{lem:all_ac_equiv}, we choose $w = x_n$.
	With the second-last relator \( x_{n-1} = z_{n-1} x_n z_{n-1}^{-1} \), substituting \( x_n = 1 \) simplifies it to \( x_{n-1} = 1 \). Substituting \( x_{n-1} = 1 \) into the preceding relator, we proceed iteratively, simplifying each relation until the first relation reads \( x_1 = 1 \).
	Thus, the presentation reduces to  $\angles{x_1, \ldots, x_n \mid x_1, \ldots, x_n}$,
	rendering it AC-trivial.
\end{proof}

\begin{remark}
	\label{rem:sub+rem}
	Starting with a balanced presentation of \cref{thm:unknot} and reducing the number of generators only through recursive application of substitution and removal (\cref{lem:substitution}), we obtain a large class of AC-trivial presentations with fewer generators. Indeed, each relator in any such presentation is still a simple conjugation relation ($x_i=zx_jz^{-1}$ where $z$ is some word in the generators). Thus we can use the same argument as in the proof of \cref{thm:unknot} to trivialize the relators iteratively.

	We also note that for any AC-trivial presentation, we can freely apply any moves (AC1)–(AC4), as these do not change the AC-triviality of the presentation.
\end{remark}

For example, consider the 3-generator family presented in the previous subsection, which was obtained through repeated application of \cref{lem:substitution}. The first relator is $x$ equals $z$ conjugated by  $yxyx^{-1}z^{-1}xy^{-1}x^{-1}y^{-1}xyx^{-1}$, and the second relator is $y$ equals $x$ conjugated by $xyx^{-1}z^{-1}xy^{-1}x^{-1}yxyx^{-1}zxy^{-1}x^{-1}y^{-1}$. Substituting $w=z$ simplifies the first relator to $x$, which when substituted into the second relator, simplifies it to $y$.

\medskip
We note, however, that the method described in \cref{rem:sub+rem} is only one way to reduce the number of generators. It is also possible to consider more general sequences of stable AC moves, which may produce presentations where relators are not all in simple conjugation form. It would be interesting to prove or disprove that any such presentation is also AC-trivial.

\begin{conjecture}
	\label{conj:unknot}
	Let $\angles{ x_1,\ldots, x_n \mid r_1,\ldots, r_{k-1},r_{k+1},\ldots, r_n, w}$ be the AC-trivial presentation coming from an unknot diagram as in \cref{thm:unknot}. Then any stably AC-equivalent presentation $\angles{ x_1,\ldots, x_m \mid s_1,\ldots, s_{m-1}, w}$ is also AC-trivial.
\end{conjecture}

Moreover, Wirtinger presentations of the unknot diagram are specific cases of presentations of $\mathbb{Z}$ where each generator generates the group. It is also interesting to consider whether or not any such presentation of $\mathbb{Z}$ leads to AC-trivial presentations.

\begin{conjecture}
	\label{conj:general_Z}
	Let $\langle x_1,\ldots x_n | r_1,\ldots, r_{n-1}\rangle$ be any presentation of $\mathbb{Z}$ where each $x_i$ generates $\mathbb{Z}$, i.e., $x_i=x_j^{\pm 1}$ for any $i,j$. Then for any word $w$ in the $x_i$ with exponent sum $\pm 1$, the presentation $\langle x_1,\ldots x_n | r_1,\ldots, r_{n-1}, w\rangle$ is AC-trivial.
\end{conjecture}
	\appendix
	% !TEX root = ../ac_paper.tex

\section{Hyperparameters}\label{app:hyperparameters}

Here we discuss the hyperparameters used to train the Proximal Policy Optimization (PPO) and Transformer models of \cref{sec:rl} and \cref{sec:lm} respectively.

\subsubsection*{PPO Hyperparameters} 
The hyperparameters of PPO are given in \cref{tab:ppo_hyperparameters}. These hyperparameters were defined in the main text in \cref{sec:ppo}. Each of the two networks ---the actor and the critic--- was a 2-layer feed forward neural network with 512 neurons and $tanh$ non-linearlities. We used Adam optimizer for training.

The performance of PPO is known to be highly sensitive to various implementation details in addition to the choice of hyperparameters \cite{shengyi2022the37implementation, engstrom2020implementation}. We used the single-file implementation of PPO in CleanRL \cite{huang2022cleanrl}, which has been well-benchmarked against the results of the original paper \cite{schulman2017proximal}. Following \cite{engstrom2020implementation}, we used advantage normalization and clipped value loss. If the KL divergence between the old and the updated policy exceeded the target KL divergence in a mini-batch, we skipped the remaining mini-batches in the optimization phase to avoid a large jump in policy.
We did not ablate all of the choices of hyperparameters to investigate the importance of each choice.

\begin{table}[ht]
	\centering
	\begin{tabular}{lc}
		\hline
		Hyperparameter & Value \\
		\hline
		Horizon $(T)$ & 200 \\
		Rollout Length $(T')$ & 200 \\
		Number of parallel actors & 28 \\
		Total Number of Rollouts & $\sim 2 \times 10^5$ \\
		Maximum Learning Rate & $1.0 \times 10^{-4}$ \\
		Minimum Learning Rate & 0 \\
		Learning Rate Schedule & Linear Decay \\
		Number of epochs & 1 \\
		Number of mini-batches & 4 \\
		Optimization mini-batch size & 1400 \\
		Discount ($\gamma$) & 0.999 \\
		GAE parameter ($\lambda$) & 0.95 \\
		Clipping parameter $\epsilon$ & 0.2 \\
		Value Loss coefficient, $c_1$ & 0.5 \\
		Entropy Loss coefficient, $c_2$ & 0.01 \\
		Adam epsilon parameter & $10^{-5}$ \\
		Target KL divergence & 0.01 \\
		\hline
	\end{tabular}
	\caption{Table of Hyperparameters}
	\label{tab:ppo_hyperparameters}
\end{table}

\subsubsection*{Transformer Hyperparameters}
The Transformer model studied in \cref{sec:lm} is an 8-layer transformer model with the embedding space dimension of $512$ and $4$ attention heads. The context window of the Transformer had length $1024$. We used a batch size of $12$ and constant learning rate of $6 \times 10^{-5}$. We trained for a total of $25000$ iterations. We used AdamW optimizer for training with hyperparameters $\beta_1 = 0.9$ and $\beta_2 = 0.99$. We did not use any dropout during training.
   	 % !TEX root = ../ac_paper.tex

\section{Andrews--Curtis paths} \label{app:paths}
In \cref{sec:variable_horizon} of the paper, we provided two presentations which we solved with the help of a reinforcement learning agent. We will provide the AC path that solves the second of those presentations,
\[
\angles{x, y \mid x^{-1} y^4 x y^{-5} , \ x^{-1} y x y^{-1} x^{-1} y^{-3}},
\]
in this appendix. We give this path in terms of (AC$'$1) and (AC$'$2) moves introduced in \cref{sec:AC}.
\[
\begin{aligned}
h_1 &= \ r_2 \rightarrow r_2 r_1, & \quad h_5 &= \ r_2 \rightarrow x^{-1} r_2 x, & \quad h_9 &= \ r_2 \rightarrow x r_2 x^{-1}, \\
h_2 &= \ r_1 \rightarrow r_1 r_2^{-1}, & \quad h_6 &= \ r_1 \rightarrow y^{-1} r_1 y, & \quad h_{10} &= \ r_1 \rightarrow y r_1 y^{-1}, \\
h_3 &= \ r_2 \rightarrow r_2 r_1^{-1}, & \quad h_7 &= \ r_2 \rightarrow y^{-1} r_2 y, & \quad h_{11} &= \ r_2 \rightarrow y r_2 y^{-1}, \\
h_4 &= \ r_1 \rightarrow r_1 r_2, & \quad h_8 &= \ r_1 \rightarrow x r_1 x^{-1}, & \quad h_{12} &= \ r_1 \rightarrow x^{-1} r_1 x, \\
\end{aligned}
\]
The AC path has length $381$ and is given below.
\[
\begin{aligned}
& h_{2} \cdot h_{8} \cdot h_{6} \cdot h_{4} \cdot h_{10} \cdot h_{2} \cdot h_{8} \cdot h_{6} \cdot h_{6} \cdot h_{8} \cdot h_{10} \cdot h_{10} \cdot h_{10} \cdot h_{7} \cdot h_{0} \cdot h_{4} \cdot h_{6} \cdot h_{6} \\ &
h_{6} \cdot h_{6} \cdot h_{11} \cdot h_{8} \cdot h_{10} \cdot h_{10} \cdot h_{2} \cdot h_{8} \cdot h_{6} \cdot h_{4} \cdot h_{10} \cdot h_{2} \cdot h_{8} \cdot h_{6} \cdot h_{6} \cdot h_{6} \cdot h_{8} \cdot h_{10} \\ &
h_{10} \cdot h_{10} \cdot h_{7} \cdot h_{3} \cdot h_{5} \cdot h_{5} \cdot h_{5} \cdot h_{0} \cdot h_{4} \cdot h_{6} \cdot h_{11} \cdot h_{5} \cdot h_{9} \cdot h_{7} \cdot h_{11} \cdot h_{7} \cdot h_{11} \cdot h_{5} \\ &
h_{4} \cdot h_{9} \cdot h_{2} \cdot h_{8} \cdot h_{6} \cdot h_{6} \cdot h_{6} \cdot h_{8} \cdot h_{10} \cdot h_{8} \cdot h_{10} \cdot h_{10} \cdot h_{4} \cdot h_{7} \cdot h_{0} \cdot h_{4} \cdot h_{6} \cdot h_{11} \\ &
h_{2} \cdot h_{8} \cdot h_{6} \cdot h_{6} \cdot h_{8} \cdot h_{10} \cdot h_{8} \cdot h_{10} \cdot h_{2} \cdot h_{8} \cdot h_{4} \cdot h_{10} \cdot h_{7} \cdot h_{0} \cdot h_{4} \cdot h_{6} \cdot h_{4} \cdot h_{11} \\ &
h_{10} \cdot h_{2} \cdot h_{8} \cdot h_{6} \cdot h_{4} \cdot h_{10} \cdot h_{8} \cdot h_{7} \cdot h_{0} \cdot h_{4} \cdot h_{6} \cdot h_{11} \cdot h_{4} \cdot h_{7} \cdot h_{11} \cdot h_{2} \cdot h_{8} \cdot h_{6} \\ &
h_{2} \cdot h_{8} \cdot h_{6} \cdot h_{8} \cdot h_{8} \cdot h_{10} \cdot h_{7} \cdot h_{0} \cdot h_{4} \cdot h_{0} \cdot h_{4} \cdot h_{0} \cdot h_{4} \cdot h_{0} \cdot h_{4} \cdot h_{6} \cdot h_{11} \cdot h_{2} \\ &
h_{8} \cdot h_{6} \cdot h_{4} \cdot h_{6} \cdot h_{8} \cdot h_{8} \cdot h_{10} \cdot h_{10} \cdot h_{10} \cdot h_{10} \cdot h_{2} \cdot h_{8} \cdot h_{6} \cdot h_{4} \cdot h_{10} \cdot h_{8} \cdot h_{7} \cdot h_{0} \\ &
h_{4} \cdot h_{6} \cdot h_{11} \cdot h_{4} \cdot h_{2} \cdot h_{8} \cdot h_{6} \cdot h_{6} \cdot h_{8} \cdot h_{8} \cdot h_{10} \cdot h_{10} \cdot h_{7} \cdot h_{0} \cdot h_{4} \cdot h_{0} \cdot h_{4} \cdot h_{6} \\ &
h_{11} \cdot h_{8} \cdot h_{7} \cdot h_{11} \cdot h_{5} \cdot h_{4} \cdot h_{8} \cdot h_{11} \cdot h_{7} \cdot h_{4} \cdot h_{8} \cdot h_{11} \cdot h_{7} \cdot h_{9} \cdot h_{6} \cdot h_{8} \cdot h_{8} \cdot h_{10} \\ &
h_{4} \cdot h_{10} \cdot h_{10} \cdot h_{8} \cdot h_{2} \cdot h_{8} \cdot h_{6} \cdot h_{7} \cdot h_{8} \cdot h_{11} \cdot h_{8} \cdot h_{10} \cdot h_{4} \cdot h_{10} \cdot h_{10} \cdot h_{2} \cdot h_{8} \cdot h_{6} \\ &
h_{4} \cdot h_{6} \cdot h_{8} \cdot h_{8} \cdot h_{8} \cdot h_{10} \cdot h_{7} \cdot h_{11} \cdot h_{7} \cdot h_{4} \cdot h_{11} \cdot h_{10} \cdot h_{7} \cdot h_{0} \cdot h_{4} \cdot h_{6} \cdot h_{11} \cdot h_{4} \\ &
h_{2} \cdot h_{8} \cdot h_{6} \cdot h_{8} \cdot h_{8} \cdot h_{8} \cdot h_{10} \cdot h_{7} \cdot h_{0} \cdot h_{4} \cdot h_{0} \cdot h_{4} \cdot h_{0} \cdot h_{4} \cdot h_{6} \cdot h_{8} \cdot h_{11} \cdot h_{2} \\ &
h_{8} \cdot h_{8} \cdot h_{10} \cdot h_{10} \cdot h_{10} \cdot h_{2} \cdot h_{8} \cdot h_{2} \cdot h_{8} \cdot h_{6} \cdot h_{4} \cdot h_{6} \cdot h_{8} \cdot h_{10} \cdot h_{8} \cdot h_{10} \cdot h_{10} \cdot h_{7} \\ &
h_{0} \cdot h_{4} \cdot h_{6} \cdot h_{11} \cdot h_{4} \cdot h_{2} \cdot h_{8} \cdot h_{6} \cdot h_{4} \cdot h_{2} \cdot h_{8} \cdot h_{6} \cdot h_{8} \cdot h_{10} \cdot h_{7} \cdot h_{11} \cdot h_{7} \cdot h_{5} \\ &
h_{11} \cdot h_{8} \cdot h_{10} \cdot h_{4} \cdot h_{8} \cdot h_{9} \cdot h_{7} \cdot h_{4} \cdot h_{8} \cdot h_{9} \cdot h_{7} \cdot h_{0} \cdot h_{4} \cdot h_{0} \cdot h_{4} \cdot h_{6} \cdot h_{11} \cdot h_{5} \\ &
h_{9} \cdot h_{4} \cdot h_{2} \cdot h_{8} \cdot h_{6} \cdot h_{4} \cdot h_{6} \cdot h_{8} \cdot h_{8} \cdot h_{10} \cdot h_{10} \cdot h_{7} \cdot h_{0} \cdot h_{4} \cdot h_{6} \cdot h_{11} \cdot h_{6} \cdot h_{8} \\ &
h_{8} \cdot h_{10} \cdot h_{4} \cdot h_{10} \cdot h_{2} \cdot h_{8} \cdot h_{6} \cdot h_{8} \cdot h_{8} \cdot h_{10} \cdot h_{7} \cdot h_{11} \cdot h_{7} \cdot h_{5} \cdot h_{11} \cdot h_{9} \cdot h_{7} \cdot h_{9} \\ &
h_{7} \cdot h_{4} \cdot h_{0} \cdot h_{4} \cdot h_{6} \cdot h_{11} \cdot h_{8} \cdot h_{8} \cdot h_{8} \cdot h_{10} \cdot h_{7} \cdot h_{0} \cdot h_{4} \cdot h_{6} \cdot h_{4} \cdot h_{11} \cdot h_{10} \cdot h_{2} \\ &
h_{8} \cdot h_{8} \cdot h_{8} \cdot h_{10} \cdot h_{7} \cdot h_{0} \cdot h_{4} \cdot h_{6} \cdot h_{11} \cdot h_{4} \cdot h_{2} \cdot h_{8} \cdot h_{8} \cdot h_{2} \cdot h_{8} \cdot h_{10} \cdot h_{2} \cdot h_{8} \\ &
h_{6} \cdot h_{4} \cdot h_{6} \cdot h_{8} \cdot h_{10} \cdot h_{7} \cdot h_{0} \cdot h_{4} \cdot h_{6} \cdot h_{11} \cdot h_{8} \cdot h_{8} \cdot h_{8} \cdot h_{8} \cdot h_{8} \cdot h_{8} \cdot h_{8} \cdot h_{1} \\ &
h_{7} \cdot h_{5} \cdot h_{11}
\end{aligned}
\]
The sequence should be read from left to right: we first apply $h_2$, then $h_8$, and so forth.

Besides the well known Akbulut–Kirby and Miller–Schupp families, we tried other types of potential counterexamples to the Andrews-Curtis conjecture. For example, we were able to trivialize the following presentation
\[
P (5,2; \tfrac{4}{5} ) = \angles{ x,y \mid x^5 = y^2, y x^{-2} (y x^{-2})^2 (y^{-1} x^2)^2 }
\]
communicated to us by Jeffrey Meier and Alexander Zupan, based on their earlier work \cite{MeierZupan}. This presentation belongs to a 3-parameter family which has a nice geometric origin, where the presentations being trivial is implied by the corresponding R-links having Generalized Property R \cite{MR2740649}.

\begin{theorem}
The presentation $P (5,2; \frac{4}{5} )$ is AC-trivial and can be trivialized in 49 steps.
\end{theorem}

\begin{proof}
The statement is certified by the following explicit AC path of length $49$:
\[
\begin{aligned}
& h_{11} \cdot h_{11} \cdot h_{1} \cdot h_{11} \cdot h_{11} \cdot h_{11} \cdot h_{6} \cdot h_{8} \cdot h_{8} \cdot h_{6} \cdot h_{0} \cdot h_{8} \cdot h_{8} \cdot h_{9} \cdot h_{11} \cdot h_{1} \cdot h_{9} \cdot h_{11} \\ &
h_{5} \cdot h_{7} \cdot h_{1} \cdot h_{9} \cdot h_{11} \cdot h_{8} \cdot h_{3} \cdot h_{5} \cdot h_{4} \cdot h_{4} \cdot h_{2} \cdot h_{8} \cdot h_{10} \cdot h_{4} \cdot h_{9} \cdot h_{0} \cdot h_{5} \cdot h_{4} \\ &
h_{4} \cdot h_{1} \cdot h_{9} \cdot h_{10} \cdot h_{4} \cdot h_{3} \cdot h_{7} \cdot h_{5} \cdot h_{2} \cdot h_{2} \cdot h_{1} \cdot h_{5} \cdot h_{1}
\end{aligned}
\]
As usual, the sequence should be read from left to right, e.g. we first apply $h_{11}$ twice, then $h_{1}$, and so on.
\end{proof}
        %\input{app/RLpath381}
	% !TEX root = ../ac_paper.tex

\section{Neighborhood constructions}\label{s:neighborhoods}

\subsection{Neighborhoods for MS series}

We define the $n$-neighborhood of a balanced presentation $\pi$ as the set of all balanced presentations that can be obtained by applying at most $n$ AC-moves to $\pi$.
We used \cref{alg:bfs_neigh}, a variation of BFS, to generate $5$-neighborhoods of presentations in the Miller--Schupp series. 
As before, we disregard the order of the relators.

\begin{algorithm}
	\caption{Breadth-First Search Algorithm Bounded by Number of Steps}\label{alg:bfs_neigh}
	\begin{algorithmic}[1]
		\State \textbf{Input:} A balanced presentation $\pi$, positive integer $n$
		\State \textbf{Output:} $n$-neighborhood of $\pi$
		\State Initialize a queue $Q$, set of visited nodes $visited$, and numerical map $dist$ that represents the minimal number of AC-moves needed to transform $\pi$ into a given presentation
		\State Mark $\pi$ as visited, put it into queue $Q$, and set its distance to $0$
		\While{$Q$ is not empty}
		\State $u \gets $ top of $Q$ \Comment{Remove the front node of $Q$}
		\For{every AC-move $m$}
		\State $child \gets m(u)$
		\If{$dist[u] < n$ and $child$ is not in $visited$}
		\State Put $child$ in $Q$ and mark it as visited
		\State Set $dist[child] = dist[u] + 1$
		\EndIf
		\EndFor
		\EndWhile
		\Return set $visited$
	\end{algorithmic}
\end{algorithm}

\subsection{Neighborhoods of the identity}

For any $\ell \in \{3, \dots, 16\}$, we constructed a neighborhood of the identity using an algorithm based on BFS search (\cref{alg:bfs_1}).
This neighborhood contains all presentations that can be connected to the identity via a path of AC-moves, where each presentation in the path has a length less than or equal to $\ell$, that is, the full based subgraph containing vertices with connectivity value less than or equal to $\ell$.
We consider the relators of a presentations as a set (meaning that the order of relators is not important; implemented as a tuple of relators in lexicographic order) 

\begin{algorithm}
	\caption{Breadth-First Search Algorithm Bounded by Size}
	\label{alg:bfs_1}
	\begin{algorithmic}[1]
		\State \textbf{Input:} A balanced presentation $\pi$, maximal size of presentation $n$
		\State \textbf{Output:} Set of enumerated presentations connected to the starting presentation that are achievable without exceeding the size limit, and set of edges with filtrations
		\State Initialize a queue $Q$, set of visited nodes $visited$, and numerical map $name$ that will enumerate presentations
		\State Mark $\pi$ as visited, put it into queue $Q$, and assign it the number $0$
		\While{$Q$ is not empty}
		\State $u \gets $ top of $Q$ \Comment{Remove the front node of $Q$}
		\For{every AC-move $m$}
		\State $child \gets m(u)$
		\If{$child$'s size $\leq n$ and $child$ is not visited}
		\State Put $child$ in $Q$ and mark it as visited
		\State Assign $child$ the next available number
		\EndIf
		\If{$child$'s size $\leq n$ and $u$'s number is smaller than $child$'s number}
		\State Return edge $(u, child)$ with proper filtration
		\EndIf
		\EndFor
		\EndWhile
	\end{algorithmic}
\end{algorithm}
	% \input{app/profiles}
	% !TEX root = ../ac_paper.tex

\section{Language modeling dataset generation \label{app:algorithm}}

This appendix describes the method, \cref{alg:apply_ac_moves}, used to generate the training and evaluation datasets for the Transformer model, as referenced in \cref{sec:lm}. Our aim was to create datasets featuring presentations of varying lengths. We began with a presentation \(P_0\) from the Miller--Schupp series, where $n \leq 7$ and $\length(w) \leq 7$, setting a maximum relator length \(l_{\text{max}} = 128\). Presentations were generated in \(n=128\) phases, each phase allowing a maximum relator length \(l_i \sim \mathcal{U}(l + i \cdot l_{\text{inc}}, l + (i+1) \cdot l_{\text{inc}})\). Here, \(l\) represents the longest relator length in \(P_0\) and \(l_{\text{inc}} = (l_{\text{max}} - l)/n\) is the incremental increase per phase. In each phase, we selected a presentation \(P\) from the previous phase and applied \(N=1000\) AC$'$ moves. Any AC$'$ move that exceeded the length \(l_i\) resulted in no change.

We repeated this for all 1190 presentations in the Miller--Schupp series, ultimately producing approximately 1.8 million balanced presentations. The length distribution of these presentations is detailed in \cref{fig:gpt_data}.

\begin{algorithm}
	\caption{Transformer Dataset Generation}
	\label{alg:apply_ac_moves}
	\begin{algorithmic}[1]
		\State \textbf{Input:}
		\begin{itemize}
			\item[] $P_0$ -- an initial presentation with $l$ as the length of the longest relator
			\item[] $n$ -- number of phases
			\item[] $m$ -- number of presentations in each phase
			\item[] $N$ -- number of AC$'$ moves to apply in each phase
			\item[] $l_{\text{max}}$ -- upper bound on presentation lengths in the dataset
		\end{itemize}
		\State\textbf{Output:}
		\begin{itemize}
			\item[] Dataset (the final collection of presentations)
		\end{itemize}
		\State Dataset $\gets \emptyset$ \Comment{Initialize the dataset of all presentations}
		\State $l_{\text{inc}} \gets (l_{\text{max}} - l) / n$ \Comment{Increment for the maximum relator length per phase}
		\For{$i = 0$ \textbf{to} $n-1$} \Comment{Loop over each phase}
			\For{$j = 1$ \textbf{to} $m$} \Comment{Generate $m$ presentations for each phase}
			\State $l_i \sim \mathcal{U}(l + i \cdot l_{\text{inc}}, l + (i+1) \cdot l_{\text{inc}})$
			\Comment{Sample maximum relator length}
			\State $P \gets (i = 1) \ ? \ P_0 \ : \ \text{Dataset}[(i-1) \cdot m + j - 1]$
			\For{$k = 1$ \textbf{to} $N$} \Comment{Apply $N$ AC$'$ moves with relator length $l_i$}
				\State $A \sim \text{AC}' \text{ Moves}$
				\State $P \gets A \cdot P$
			\EndFor
			\State Dataset $\gets$ Dataset $\cup \{P\}$
			\Comment{Add the presentation $P$ to the Dataset}
			\EndFor
		\EndFor
	\end{algorithmic}
\end{algorithm}
    \section{Reidemeister moves realized by stable AC moves}\label{app:reid}

In this appendix, we prove \cref{prop:unknot-stable}, which we rewrite below.

\begin{proposition}[Myasnikov, Myasnikov, and Shpilrain, \cite{MMS}]
    Let $\angles{ x_1,\ldots, x_n\, \mid \, r_1, \ldots, r_n }$ be the Wirtinger presentation of an unknot diagram. Then for any $k=1,\ldots,n$ and word $w$ in the $x_i$ with exponent sum $\pm 1$, the balanced presentation of the trivial group $\langle x_1,\ldots, x_n\,|\, r_1,\ldots, r_{k-1}, r_{k+1},\ldots, r_n, w\rangle$ is stably Andrews-Curtis-trivial.
\end{proposition}

As reviewed in \cref{subsec:ac-unknot}, given any oriented knot diagram, we assign generators $\{x_i\}_{i=1}^n$ to each of the arcs and relators $\{ r_i\}_{i=1}^n$ to each of the crossings. These generators and relators form a balanced presentation of the fundamental group of the knot complement. There exists an ordering of the relators and a choice of $\pm 1$ as exponents such that the relators satisfy the equation, 
\[
\prod\limits_{i=1}^n r_i^{\pm 1} = 1
\]
Thus any relator can be eliminated, giving $\angles{x_1, \dots , x_n \mid r_1, \dots, r_{k-1}, r_{k+1}, \dots, r_n }$ without changing the underlying group. When the knot is an unknot, adding a relator $w$, which is a word in $x_1, \dots , x_n$ with exponent sum $\pm 1$, gives a balanced presentation of the trivial group $\angles{x_1, \dots , x_n \mid r_1, \dots, r_{k-1}, r_{k+1}, \dots, r_n , w}$. 

Here, we will show that the application of Reidemeister moves to a knot diagram of the unknot corresponds to application of stable AC moves (AC1)--(AC5) to the associated balanced presentation of the trivial group. This relationship was first studied in \cite{WADA1994241}.
As a knot diagram of the unknot is reduced to the trivial diagram, the associated balanced presentation is shown to be stably AC-trivial.

\begin{figure}
    \centering
\begin{tikzpicture}
\draw[thick,rounded corners] (0,0)--(2,0)--(2,-1)--(1,-1)--(1,-0.2);
\draw[thick,rounded corners] (1,0.2)--(1,1);

\draw[<->] (2.5,0)--(3.5,0);

\draw[thick,rounded corners] (4,0)--(5,0)--(5,1);

\node at (4.5,-0.5){$x$};
\node at (0.5,-0.5){$x$};
\node at (1.5,0.5){$y$};

\node at (0.5,0){$>$};
\node at (1,0.5){\rotatebox{90}{$>$}};
\node at (4.5,0){$>$};

\node at (-0.5,0){\textbf{R1}};

\node at (7.5,-3){\textbf{R2b}};

\draw[thick] (9,-2)--(9,-2.3);
\draw[thick] (9,-2.7)--(9,-3.3);
\draw[thick] (9,-3.7)--(9,-4);

\draw[thick,rounded corners] (10,-2)--(10,-2.5)--(8.5,-2.5)--(8.5,-3.5)--(10,-3.5)--(10,-4);

\node at (9,-1.7){$x$};
\node at (10,-1.7){$y$};
\node at (9.2,-3){$z$};
\node at (9,-4.3){$u$};

\node at (9,-2){\rotatebox{90}{$>$}};
\node at (10,-2){\rotatebox{90}{$>$}};
\node at (9,-3){\rotatebox{90}{$>$}};
\node at (9,-3.7){\rotatebox{90}{$>$}};

\draw[<->] (10.5,-3)--(11.5,-3);

\draw[thick] (12.5,-2)--(12.5,-4);
\draw[thick] (13.5,-2)--(13.5,-4);

\node at (12.5,-3){\rotatebox{90}{$>$}};
\node at (13.5,-3){\rotatebox{90}{$>$}};

\node at (12.5,-1.7){$x$};
\node at (13.5,-1.7){$y$};

\draw[thick] (1,-2)--(1,-2.3);
\draw[thick] (1,-2.7)--(1,-3.3);
\draw[thick] (1,-3.7)--(1,-4);

\draw[thick,rounded corners] (2,-2)--(2,-2.5)--(0.5,-2.5)--(0.5,-3.5)--(2,-3.5)--(2,-4);

\node at (1,-2){\rotatebox{90}{$>$}};
\node at (2,-2.3){\rotatebox{90}{$<$}};
\node at (1,-3){\rotatebox{90}{$>$}};
\node at (1,-3.7){\rotatebox{90}{$>$}};

\draw[<->] (2.5,-3)--(3.5,-3);

\draw[thick] (4.5,-2)--(4.5,-4);
\draw[thick] (5.5,-2)--(5.5,-4);

\node at (4.5,-3){\rotatebox{90}{$>$}};
\node at (5.5,-3){\rotatebox{90}{$<$}};

\node at (4.5,-1.7){$x$};
\node at (5.5,-1.7){$y$};

\node at (1,-1.7){$x$};
\node at (2,-1.7){$y$};
\node at (1.2,-3){$z$};
\node at (1,-4.3){$u$};

\node at (-0.5,-3){\textbf{R2a}};

\draw[thick] (1,-5)--(1,-8);
\draw[thick] (0.5,-7.5)--(0.8,-7.5);
\draw[thick,rounded corners] (1.2,-7.5)--(2,-7.5)--(2,-5.5)--(3.5,-5.5);
\draw[thick] (0.5,-6.5)--(0.8,-6.5);
\draw[thick] (1.2,-6.5)--(1.8,-6.5);
\draw[thick] (2.2,-6.5)--(3.5,-6.5);

\draw[<->] (4,-6.5)--(5,-6.5);

\draw[thick] (5.5,-6.5)--(6.8,-6.5);
\draw[thick,rounded corners] (5.5,-7.5)--(7,-7.5)--(7,-5.5)--(7.8,-5.5);
\draw[thick,rounded corners] (7.2,-6.5)--(7.8,-6.5);
\draw[thick] (8,-5)--(8,-8);
\draw[thick] (8.2,-5.5)--(8.5,-5.5);
\draw[thick] (8.2,-6.5)--(8.5,-6.5);

\node at (0.5,-5.5){$x$};
\node at (0.5,-6.2){$y$};
\node at (0.5,-7.2){$z$};
\node at (3,-5.2){$r$};
\node at (3,-6.2){$u$};
\node at (1.5,-6.2){$v$};

\node at (1,-5.5){\rotatebox{90}{$>$}};
\node at (0.7,-6.5){$>$};
\node at (0.7,-7.5){$>$};
\node at (1.5,-6.5){$>$};
\node at (3,-5.5){$>$};
\node at (3,-6.5){$>$};

\node at (8,-4.5){$x$};
\node at (6,-6.2){$y$};
\node at (6,-7.2){$z$};
\node at (9,-5.5){$r$};
\node at (9,-6.5){$u$};
\node at (7.5,-6.2){$t$};

\node at (8,-5){\rotatebox{90}{$>$}};
\node at (6,-6.5){$>$};
\node at (6,-7.5){$>$};
\node at (8.5,-6.5){$>$};
\node at (8.5,-5.5){$>$};
\node at (7.5,-6.5){$>$};

\node at (-0.5,-6){\textbf{R3}};

\end{tikzpicture}
    \caption{Reidemeister moves}
    \label{fig:rm}
\end{figure}

Two diagrams of the same knot can be related by a finite sequence of the three
Reidemeister moves shown in \cref{fig:rm}. Here we distinguish \textbf{R2a} and \textbf{R2b} just because they give slightly different computation about the presentations. Since we want to consider the behavior of balanced presentations $\angles{ x_1,\ldots, x_n \mid  r_1,\ldots, r_{k-1}, r_{k+1},\ldots, r_n, w}$ of the trivial group under Reidemeister moves, we can assume that the relator eliminated, $r_k$, is not one given by the crossings in \textbf{R1}, \textbf{R2a}, \textbf{R2b}, \textbf{R3}. We note, however, that since any relator in a Wirtinger presentation can be written as the product of the other relators, the relator we eliminated can be recovered using (AC1)--(AC3).
\\
\\
\textbf{R1}:
\begin{align*}
\langle x_1,\ldots,x_n,x,y\mid r_1,\ldots,r_n,yx^{-1},w\rangle\longleftrightarrow\langle x_1,\ldots,x_n,x\mid r_1',\ldots,r_n',w'\rangle
\end{align*}
Here $r_i'$ and $w'$ are obtained from replacing $y$ in $r_i$ and $w$ by $x$. The equivalence comes from the substitution in \cref{lem:substitution}.
\\
\\
\textbf{R2a}:
\begin{align*}
\,\,&\langle x_1,\ldots,x_n,x,y,z,u\mid r_1,\ldots,r_{n+1},xyz^{-1}y^{-1},zy^{-1}u^{-1}y,w\rangle
\\
&\longleftrightarrow\langle x_1,\ldots,x_n,x,y,u\mid r_1',\ldots,r_{n+1}',ux^{-1},w'\rangle
\\
&=\langle x_1,\ldots,x_n,x,y,u\mid r_1,\ldots,r_{n+1},ux^{-1},w'\rangle
\\
&\longleftrightarrow\langle x_1,\ldots,x_n,x,y\mid \tilde{r}_1,\ldots,\tilde{r}_{n+1},\tilde{w}\rangle
\end{align*}
Here $r_i',w'$ are obtained from replacing $z$ in $r_i,w$ by $y^{-1}xy$; $\tilde{r_i}$ from replacing $u$ in $r_i$ by $x$; and $\tilde{w}$ from replacing $u$ in $w'$ by $x$. The second equation comes from the fact that the $r_i$ do not contain $z$.
\\
\textbf{R2b} is similar to \textbf{R2a}.
\\
\\
\textbf{R3}:
\begin{align*}
&\langle x_1,\ldots,x_n,x,y,z,u,v,r\mid r_1,\ldots,r_{n+2},vxy^{-1}x^{-1},urv^{-1}r^{-1},rxz^{-1}x^{-1},w\rangle
\\
&\longleftrightarrow\langle x_1,\ldots,x_n,x,y,z,u,r\mid r_1,\ldots,r_{n+2},urxy^{-1}x^{-1}r^{-1},rxz^{-1}x^{-1},w'\rangle
\\
&\longleftrightarrow\langle x_1,\ldots,x_n,x,y,z,u,r\mid r_1,\ldots,r_{n+2},uxzy^{-1}z^{-1}x^{-1},rxz^{-1}x^{-1},w'\rangle
\\
&\longleftrightarrow\langle x_1,\ldots,x_n,x,y,z,u,r,t\mid r_1,\ldots,r_{n+2},uxzy^{-1}z^{-1}x^{-1},rxz^{-1}x^{-1},tzy^{-1}z^{-1},w'\rangle
\\
&\longleftrightarrow\langle x_1,\ldots,x_n,x,y,z,u,r,t\mid r_1,\ldots,r_{n+2},rxz^{-1}x^{-1},tzy^{-1}z^{-1},uxt^{-1}x^{-1},w'\rangle
\end{align*}
Here $w'$ is obtained from replacing $v$ in $w$ by $xyx^{-1}$.

After reducing our diagram to the trivial diagram of the unknot, we are left with $\langle x \mid \bar w\rangle$, where $\bar w$ is $w$ after applying all the Reidemeister moves. This is because the trivial diagram of the unknot gives the presentation $\langle x\mid \,\,\rangle$ of the infinite cyclic group, so $\bar w$ is the only relator remaining.  Moreover, $\bar w = x^{\pm1}$ since at each stage, $w$ has exponent sum $\pm 1$, so $\langle x \mid \bar w\rangle$ can be reduced to the trivial presentation $\langle \,\, \mid \,\, \rangle$. Thus \cref{prop:unknot-stable} is proven.
    % !TEX root = ../ac_paper.tex

\section{A misprint in the Wirtinger presentation of \cite{MMS}.}  \label{app:mms}
The unknot diagram of \cref{fig:unknot} was originally presented in \cite{MMS}. Their manuscript contained an unfortunate misprint in the 13th relator, where it is written as $x_{13}=x_5 x_{12} x_5^{-1}$.  The resultant presentation—denoted $W’$—is not a Wirtinger presentation of any knot diagram. In a Wirtinger presentation, any single relator is redundant, meaning that removing any relator still yields a valid presentation of the knot group. In the case of an unknot, this group is the infinite cyclic group $\mathbb{Z}$. However, removing different relators from $W’$ produces different groups. For instance, if the relator $x_{7} = x_4^{-1} x_{6} x_4$ is removed, the resulting group is the three-strand braid group, whereas removing $x_{14} = x_1 x_{13} x_1^{-1}$ yields $\mathbb{Z}$.

In \cite[Theorem 1.4]{MMS}, the following 3-generator family of presentations,
\[
\angles{ x,y,z \mid  x=z\cdot [[y^{-1},x^{-1}],z], y=x\cdot [[y^{-1},x^{-1}],z^{-1}]\cdot [z^{-1},x], w},
\] 
was obtained by deleting the 12th relator from $W'$, adding $w$, and then recursively eliminating other generators. The theorem statement claimed that every element of this 3-generator family presents the trivial group. However, due to the misprint, $W'$ with the 12th relator eliminated is not a presentation of $\mathbb{Z}$.

Despite this error, it is possible to get presentations of the trivial group after choosing appropriate $w$. For example, when $w = x^{-1} y z$, the 3-generator presentation is stably AC-equivalent to a length $25$ presentation,
\[
\langle x, y \mid
	x^{-1}y^{-1}xy^{-1}x^{-1}yxy^{-2}xyx^{-1}y, \
	y^{-1}x^{-1}y^2x^{-1}y^{-1}xyxy^{-2}x 
\rangle,
\]
which is AC-equivalent to $\AK(3)$. Note, however, that unlike any presentation AC-equivalent to a correct Wirtinger presentation, these presentations are not necessarily stably AC-trivial. For completeness, we provide the AC path that connects the length $25$ presentation to $\AK(3)$,
\[
\begin{aligned}
& h_9 \cdot h_7 \cdot h_4 \cdot h_8 \cdot h_{11} \cdot h_5 \cdot h_{11} \cdot h_9 \cdot h_3 \cdot h_{10} \cdot h_{12} \cdot h_7 \cdot h_7 \cdot h_9 \cdot h_{11} \cdot h_5 \cdot h_3 \cdot h_5 \cdot \\
& h_4 \cdot h_3 \cdot h_{12} \cdot h_5 \cdot h_7 \cdot h_7 \cdot h_1 \cdot h_9 \cdot h_{11} \cdot h_8 \cdot h_3 \cdot h_5 \cdot h_{10} \cdot h_2 \cdot h_6 \cdot h_{12} \cdot h_9 \cdot h_7 \cdot \\
& h_5 \cdot h_{11} \cdot h_{10} \cdot h_3 \cdot h_8 \cdot h_{11} \cdot h_9 \cdot h_2 \cdot h_{10} \cdot h_{12} \cdot h_5 \cdot h_7 \cdot h_9 \cdot h_{11} \cdot h_1 \cdot h_9 \cdot h_8.
\end{aligned}
\]
Here, $h_i$ are AC$'$ moves as defined in \cref{app:paths}.

        %\input{app/families}
        %\input{app/stable_ak3}
	% !TEX root = ../ac_paper.tex

\subsection*{Funding}

The work of A.S. is supported by the US Department of Energy grant DE-SC0010008 to Rutgers University. The authors acknowledge the contributions of Office of Advanced Research Computing (OARC) at Rutgers University for providing access to the Amarel cluster and other computing resources.
A.M.'s work is supported by NSERC grants RES000678 and R7444A03. A.M. also gratefully acknowledges the excellent working conditions provided by the Max Planck Institute for Mathematics in Bonn.
The work of P.K. and B.L. is supported by the SONATA grant no. 2022/47/D/ST2/02058 funded by the Polish National Science Centre. This research was carried out with the support of the Interdisciplinary Centre for Mathematical and Computational Modelling at the University of Warsaw (ICM UW).
The work of S.G. is supported in part by a Simons Collaboration Grant on New Structures in Low-Dimensional Topology, by the NSF grant DMS-2245099, and by the U.S. Department of Energy, Office of Science, Office of High Energy Physics, under Award No. DE-SC0011632.
Z.W. is partially supported by ARO MURI contract W911NF-20-1-0082. Y.Q. thanks Nankai Zhide Foundation for support.
	\sloppy
	\printbibliography
    %\input{sec/legacy}
%    \todos
\end{document}